\documentclass{article}

\usepackage{iclr2017_arXiv,times}
\usepackage{hyperref}
\usepackage{url}

\usepackage{tabularx}
\usepackage{multirow}

\usepackage{algorithm2e}

\usepackage{graphicx} 

\usepackage{bm}
\usepackage{amsmath}

\DeclareMathOperator*{\argmax}{arg\,max}

\usepackage{xcolor}

\SetCommentSty{mycommfont}

\usepackage{hyperref}
\hypersetup{
    colorlinks,
    citecolor=black,
    filecolor=black,
    linkcolor=black,
    urlcolor=black
}

\usepackage{multirow}

\usepackage{amssymb} 
\usepackage{rotating} 

\usepackage{lipsum}

\title{Deep Reinforcement Learning}

\author{
  Yuxi Li (yuxili@gmail.com)\\
  }

\begin{document}

\maketitle

\begin{abstract} 

\noindent
We discuss deep reinforcement learning in an overview style. 
We draw a big picture, filled with details.
We discuss six core elements, six important mechanisms, and twelve applications,
focusing on contemporary work, and in historical contexts.
We start with background of artificial intelligence, machine learning, deep learning, and reinforcement learning (RL),
with resources.
Next we discuss RL core elements, including
value function, 
policy,
reward,
model,
exploration vs. exploitation,
and representation.
Then we discuss important mechanisms for RL, including
attention and memory,
unsupervised learning,
hierarchical RL, 
multi-agent RL,
relational RL,
and learning to learn.
After that, we discuss RL applications, 
including games, 
robotics,
natural language processing (NLP), 
computer vision,
finance,
business management,
healthcare, 
education,
energy,
transportation,
computer systems,
and, science, engineering, and art.
Finally we summarize briefly,
discuss challenges and opportunities,
and close with  an epilogue.
\footnote{Work in progress. Under review for Morgan \& Claypool: Synthesis Lectures in Artificial Intelligence and Machine Learning.
This manuscript is based on our previous deep RL overview~\citep{Li2017DeepRL}.
It benefits from discussions with and comments from many people. Acknowledgements will appear in a later version. 
}

\vspace{5mm}


\textbf{}

\end{abstract}

\vfill

\clearpage

\newpage

\tableofcontents

\newpage

\section{Introduction}
\label{introduction}

Reinforcement learning (RL) is about an agent interacting with the environment, learning an optimal policy, by trial and error, for sequential decision making problems, in a wide range of fields in natural sciences, social sciences, and engineering~\citep{Sutton98, Sutton2018, Bertsekas96, Bertsekas12, Szepesvari2010, Powell11}. 

The integration of reinforcement learning and neural networks has a long history~\citep{Sutton2018,Bertsekas96, Schmidhuber2015-DL}. With recent exciting achievements of deep learning~\citep{LeCun2015, Goodfellow2016}, benefiting from big data, powerful computation, new algorithmic techniques, mature software packages and architectures, and strong financial support, we have been witnessing the renaissance of reinforcement learning~\citep{Krakovsky2016}, especially, the combination of deep neural networks and reinforcement learning, i.e., deep reinforcement learning (deep RL).\footnote{We choose to abbreviate deep reinforcement learning as "deep RL", since it is a branch of reinforcement learning, in particular, using deep neural networks in reinforcement learning, and "RL" is a well established abbreviation for reinforcement learning.} 

Deep learning, or deep neural networks, has been prevailing in reinforcement learning in the last several years, in  games, robotics, natural language processing, etc. We have been witnessing breakthroughs, 
like deep Q-network (DQN)~\citep{Mnih-DQN-2015}, AlphaGo~\citep{Silver-AlphaGo-2016, Silver-AlphaGo-2017}, 
and DeepStack~\citep{Moravcik2017}; each of them represents a big family of problems and large number of applications.
DQN~\citep{Mnih-DQN-2015} is for single player games, and single agent control in general.
DQN has implications for most algorithms and applications in RL.
DQN ignites this round of popularity of deep reinforcement learning.
AlphaGo~\citep{Silver-AlphaGo-2016, Silver-AlphaGo-2017} is for two player perfect information zero-sum games.
AlphaGo makes a phenomenal achievement on a very hard problem, and sets a landmark in AI.
The success of AlphaGo influences similar games directly, 
and Alpha Zero~\citep{Silver-AlphaZero-2017} has already achieved significant successes on chess and Shogi.
The techniques underlying AlphaGo~\citep{Silver-AlphaGo-2016} and AlphaGo Zero~\citep{Silver-AlphaGo-2017},
namely, deep learning, reinforcement learning, Monte Carlo tree search (MCTS), and self-play, 
will have wider and further implications and applications.
As recommended by AlphaGo authors in their papers~\citep{Silver-AlphaGo-2016, Silver-AlphaGo-2017},
the following applications are worth further investigation:
general game-playing (in particular, video games),
classical planning,
partially observed planning,
scheduling,
constraint satisfaction,
robotics,
industrial control,
online recommendation systems,
protein folding,
reducing energy consumption, and
searching for revolutionary new materials. 
DeepStack~\citep{Moravcik2017} is for two player imperfect information zero-sum games,
a family of problems with inherent hardness to solve.
DeepStack, similar to AlphaGo, also makes an extraordinary achievement on a hard problem, and sets a milestone in AI.
It will have rich implications and wide applications, e.g., in defending strategic resources and robust decision making for medical treatment recommendations~\citep{Moravcik2017}. 

We also see novel algorithms, architectures, and applications, like
asynchronous methods~\citep{Mnih-A3C-2016}, 
trust region methods~\citep{Schulman2015, Schulman2017PPO, Nachum2018TrustPCL},
deterministic policy gradient~\citep{Silver-DPG-2014,Lillicrap2016},
combining policy gradient with off-policy RL~\citep{Nachum2017Gap, Nachum2018TrustPCL, Haarnoja2018},
interpolated policy gradient~\citep{Gu2017Interpolated},
unsupervised reinforcement and auxiliary learning~\citep{Jaderberg2017, Mirowski2017}, 
hindsight experience replay~\citep{Andrychowicz2017},
differentiable neural computer~\citep{Grave-DNC-2016}, 
neural architecture design~\citep{Zoph2017}, 
guided policy search~\citep{Levine2016}, 
generative adversarial imitation learning~\citep{Ho2016}, 
multi-agent games~\citep{Jaderberg2018Quake},
hard exploration Atari games~\citep{Aytar2018}, 
StarCraft II~\citet{SunPeng2018StarCraft, Pang2018StarCraft},
chemical syntheses planning~\citep{Segler2018},
character animation~\citep{Peng2018},
dexterous robots~\citep{OpenAI2018},
OpenAI Five for Dota 2,
etc. 

Creativity would push the frontiers of deep RL further w.r.t. core elements, important mechanisms, and applications,
seemingly without a boundary.
RL probably helps, if a problem can be regarded as or transformed into a sequential decision making problem.

Why has deep learning been helping reinforcement learning make so many and so enormous achievements? Representation learning with deep learning enables automatic feature engineering and end-to-end learning through gradient descent, so that reliance on domain knowledge is significantly reduced or even removed for some problems. 
Feature engineering used to be done manually and is usually time-consuming,  over-specified, and incomplete.  
Deep distributed representations exploit the hierarchical composition of factors in data to combat the exponential challenges of the curse of dimensionality. 
 Generality, expressiveness and flexibility of deep neural networks  make some tasks easier or possible, e.g., in the breakthroughs, and novel architectures, algorithms, and applications discussed above. 
 
Deep learning, as a specific class of machine learning, is not without limitations, e.g., as a black-box lacking interpretability,  as an "alchemy" without clear and sufficient scientific principles to work with, 
with difficulties in tuning hyperparameters,
and without human intelligence so not being able to compete with a baby in some tasks. 
Deep reinforcement learning exacerbates these issues, and even reproducibility is a problem~\citep{Henderson2018}. 
However, we see a bright future, since there are lots of work to improve deep learning, machine learning, reinforcement learning, deep reinforcement learning, and AI in general.

Deep learning  and reinforcement learning, being selected as one of the  MIT Technology Review 10 Breakthrough Technologies in 2013 and 2017 respectively,  will play their crucial roles in achieving artificial general intelligence. David Silver, the major contributor of AlphaGo~\citep{Silver-AlphaGo-2016, Silver-AlphaGo-2017}, proposes a conjecture:  artificial intelligence = reinforcement learning + deep learning~\citep{Silver2016Tutorial}. We will further discuss this conjecture in Chapter~\ref{discussion}.

There are several frequently asked questions about deep reinforcement learning as below. 
We briefly discuss them in the above, and will further elucidate them in the coming chapters.
\begin{itemize}
 \item Why deep?
 \item What is the state of the art?
 \item What are the issues, and potential solutions?
 \end{itemize}


The book outline follows. First we discuss background of artificial intelligence, machine learning, deep learning, and reinforcement learning, as well as benchmarks and resources, in Chapter~\ref{background}.
Next we discuss RL core elements, including
value function in Chapter~\ref{value}, 
policy in Chapter~\ref{policy}, 
reward in Chapter~\ref{reward},
model in Chapter~\ref{model},
exploration vs exploitation in Chapter~\ref{exploration},
and representation in Chapter~\ref{representation}.
Then we discuss important mechanisms for RL, including
attention and memory  in Chapter~\ref{attention}, 
unsupervised learning in Chapter~\ref{unsupervised},
hierarchical RL in Chapter~\ref{hierarchical}, 
multi-agent RL in Chapter~\ref{MARL}, 
relational RL in Chapter~\ref{relational},
and, 
learning to learn in Chapter~\ref{meta}.
After that, we discuss various RL applications, including
games in Chapter~\ref{games}, 
robotics in Chapter~\ref{robotics},
natural language processing (NLP) in Chapter~\ref{NLP},
computer vision in Chapter~\ref{CV},
finance and business management in Section~\ref{fin},
and more applications in Chapter~\ref{more-apps}, including, 
healthcare in Section~\ref{healthcare},
education in Section~\ref{education},
energy  in Section~\ref{energy},
transportation  in Section~\ref{transportation},
computer systems in Section~\ref{systems},
and, science, engineering and arts in Section~\ref{science}.
We close in Chapter~\ref{discussion},
with a brief summary, discussions about challenges and opportunities, and an epilogue.

\begin{figure}
\includegraphics[width=1.0\linewidth]{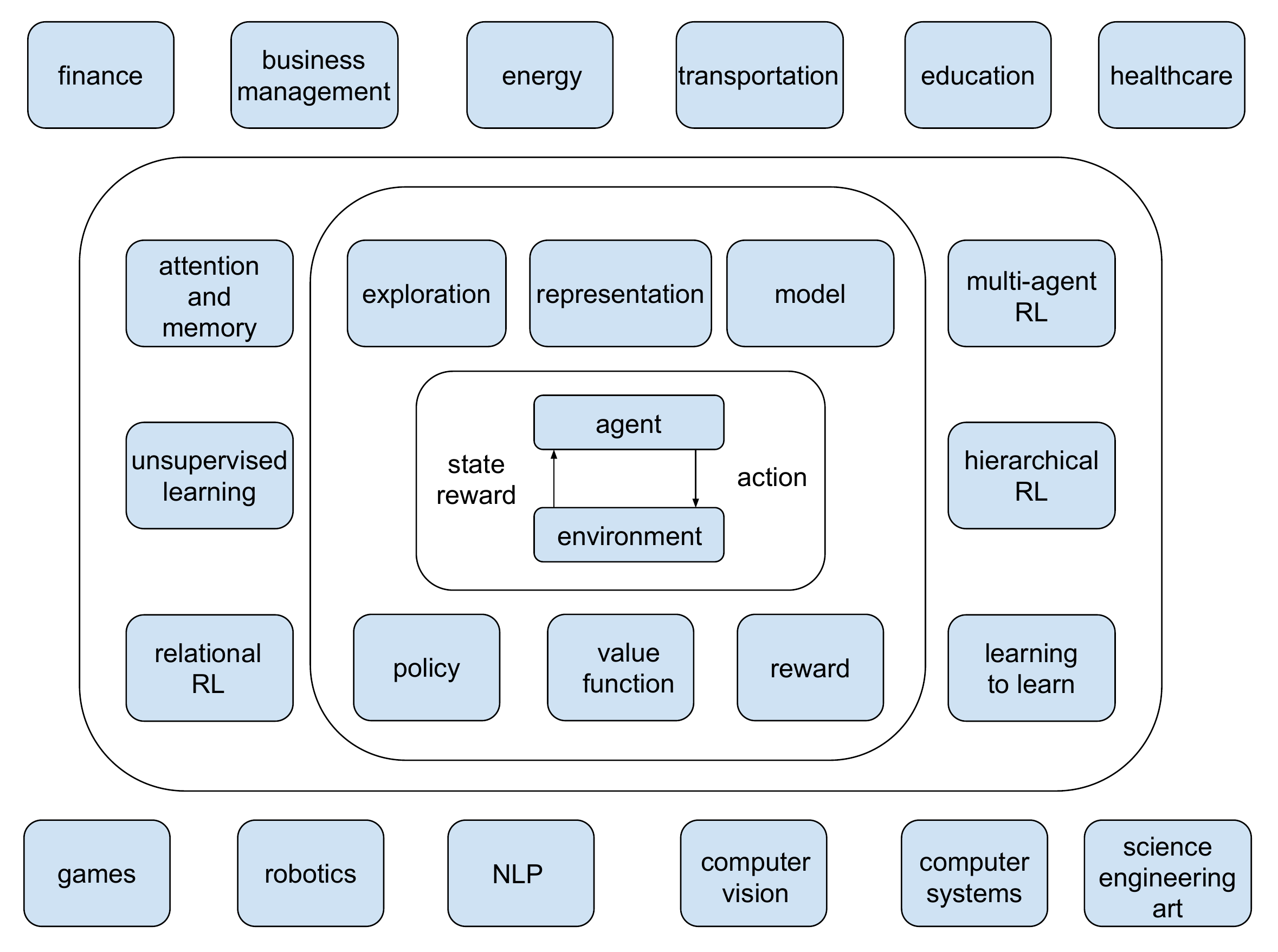}
\caption{Outline}
\label{organization}
\end{figure}

Figure~\ref{organization} illustrates the manuscript outline. The agent-environment interaction sits in the center, around which are core elements,
next important mechanisms,
then various applications.

Main readers of this manuscript are those who want to get more familiar with deep reinforcement learning, in particular, novices to (deep) reinforcement learning.
For reinforcement learning experts, as well as new comers, this book are helpful as a reference. 
This manuscript is helpful for deep reinforcement learning courses, with selected topics and papers.

We endeavour to discuss recent, representative work, and provide as much relevant information as possible, 
in an intuitive, high level, conceptual approach.
We attempt to make this manuscript complementary to~\citet{Sutton2018}, 
the RL classic focusing mostly on fundamentals in RL. 
However, deep RL is by nature an advanced topic; 
and most part of this manuscript is about introducing papers, mostly without covering detailed background. 
(Otherwise, the manuscript length would explode.) 
Consequently, most parts of this manuscript would be rather technical, somewhat rough, without full details.  
Original papers are usually the best resources for deep understanding.

Deep reinforcement learning is growing very fast.
We post blogs to complement this manuscript, and endeavour to track the development of this field , at \url{https://medium.com/@yuxili}.

This manuscript covers a wide spectrum of topics in deep reinforcement learning. 
Although we have tried our best for excellence, there are inevitably shortcomings or even mistakes. 
Comments and criticisms are welcome.


\clearpage


\section{Background}
\label{background}

In this chapter, we briefly introduce  concepts and fundamentals in artificial intelligence~\citep{Russell2009}, machine learning, deep learning~\citep{Goodfellow2016}, and, reinforcement learning~\citep{Sutton2018}. 

We do not give detailed background introduction for artificial intelligence, machine learning and deep learning; 
these are too broad to discuss in details here. 
Instead, we recommend the following recent Nature/Science survey papers: \citet{Jordan2015} for machine learning, and \citet{LeCun2015} for deep learning. For reinforcement learning, we cover some basics as a mini tutorial,  and recommend the textbook, \citet{Sutton2018}, two courses, RL course by David Silver at UCL~\citep{Silver2015Course} and Deep RL course by Sergey Levin at UC Berkeley~\citep{Levine2018}, 
and a recent Nature survey paper~\citep{Littman2015}. 
We present some resources for deep RL in Section~\ref{resources}.

\begin{figure}[h]
\includegraphics[width=1.0\linewidth]{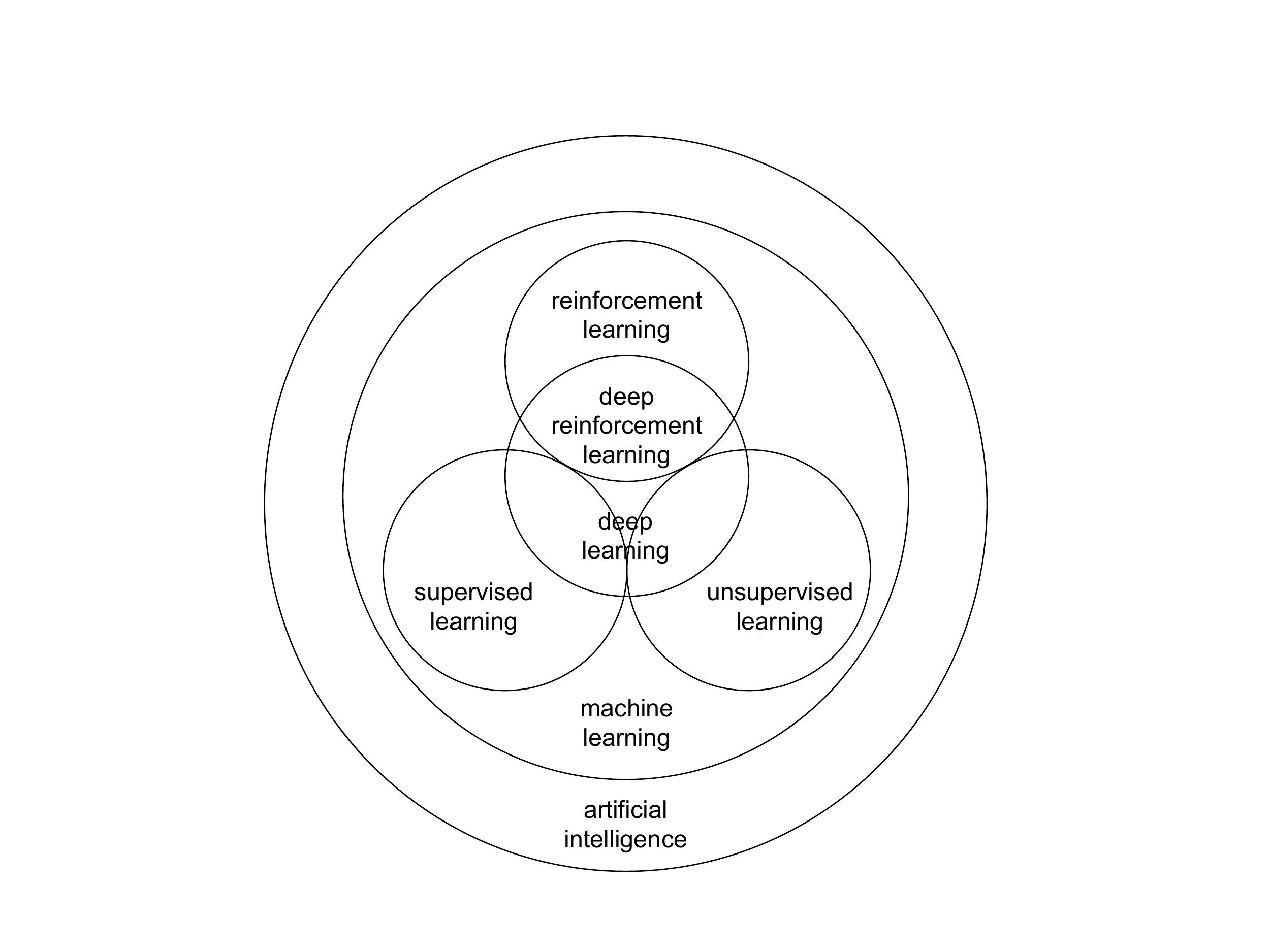}
\caption{Relationship among deep reinforcement learning, deep learning, reinforcement learning, supervised learning,  unsupervised learning,  machine learning, and, artificial intelligence. Deep learning and deep reinforcement learning are addressing many classical AI problems.}
\label{relation}
\end{figure}

In Figure~\ref{relation}, we illustrate relationship among several concepts in AI and machine learning.
Deep reinforcement learning, as the name indicates, is at the intersection of deep learning and reinforcement learning.
We usually categorize machine learning as supervised learning, unsupervised learning, and reinforcement learning.
Deep learning can work with/as supervised learning, unsupervised learning, reinforcement learning, 
and other machine learning approaches.
\footnote{Machine learning includes many approaches: 
decision tree learning,
association rule learning,
artificial neural networks,
inductive logic programming,
support vector machines,
clustering,
Bayesian networks,
reinforcement learning,
representation learning,
similarity and metric learning,
sparse dictionary learning,
genetic algorithms,
and, 
rule-based machine learning,
according to Wikipedia, \url{https://en.wikipedia.org/wiki/Machine_learning}.
Also check textbooks like \citet{Russell2009}, \citet{Mitchell1997}, and \citet{Zhou2016}.
}
Deep learning is part of machine learning, which is part of AI.
Note that all these fields are evolving,
e.g., deep learning and deep reinforcement learning are addressing classical AI problems, 
like logic, reasoning, and knowledge representation.
Reinforcement learning can be important for all AI problems, as quoted from \citet{Russell2009}, 
"reinforcement learning might be considered to encompass all of AI: an agent is placed in an environment and must learn to behave successfully therein", and, 
"reinforcement learning can be viewed as a microcosm for the entire AI problem".

\subsection{Artificial Intelligence}
\label{AI}

Artificial intelligence (AI) is a very broad area.
Even an authoritative AI textbook like~\citet{Russell2009} does not give a precise definition.
\citet{Russell2009} discuss definitions of AI from four perspectives in two comparisons: 
1) thought process and reasoning vs. behaviour, 
and, 2) success in terms of fidelity to human performance vs. rationality, an ideal performance measure. 
We follow the discussions of \citet{Russell2009}, list four ways to define AI, quoting directly from \citet{Russell2009}.

Definition 1, "acting humanly", follows the Turing Test approach. 
"The art of creating machines that perform functions that require intelligence when performed by people."~\citep{Kurzweil1992}
"The study of how to make computers do things at which, at the moment, people are better."~\citep{Rich1991}

A computer passes a Turing Test, if a human interrogator can not tell if the written responses to some written question are from a computer or a human. The computer needs the following capabilities: 
\emph{natural language processing} (NLP) for successful communication in English, 
\emph{knowledge representation} for storage of what it knows or hears, 
\emph{automated reasoning} for answering questions and drawing new conclusions from the stored information, and, 
\emph{machine learning} for adaptation to new scenarios and detection and extrapolation of patterns. 
In a total Turing Test, video signals are involved, so that a computer needs more capabilities,
\emph{computer vision} for object perception, and, 
\emph{robotics} for object manipulation and motion control.
AI researchers have been devoting most efforts to the underlying principles of intelligence, 
mostly covered by the above six disciplines, and less on passing the Turing Test, 
since, duplicating an exemplar is usually not the goal, 
e.g., an airplane may not simply imitate a bird.

Definition 2, "thinking humanly", follows the cognitive modeling approach.
"The exciting new effort to make computers think ... machines with minds, in the full and literal sense."~\citep{Haugeland1989}
"[The automation of] activities that we associate with human thinking, activities such as decision-making, problem solving, learning ..."~\citep{Bellman1978}

Definition 3, "thinking rationally", follows the "law of thought" approach.
"The study of mental faculties through the use of computational models."~\citep{Charniak1985}
"The study of the computation that make it possible to perceive, reason, and act."~\citep{Winston1992}

Definition 4, "acting rationally", follows the rational agent approach.
"Computational Intelligence is the study of the design of intelligent agents."~\citep{Poole1998}
"AI ... is concerned with intelligent behavior in artifacts."~\citep{Nilsson1998}
The rational agent approach is more amenable to scientific development than those based on human behavior or human thought.
\citet{Russell2009} thus focuse on general principles of rational agents and their building components.

We list the foundations and the questions they attempt to answer as in \citet{Russell2009}:
\emph{philosophy}, for questions like,
"Can formal rules be used to draw valid conclusions?",
"How does the mind arise from a physical brain?",
"Where does knowledge come from?", and,
"How does knowledge lead to action?";
\emph{mathematics}, for questions like,
"What are the formal rules to draw valid conclusions?",
"What can be computed?", and,
"How do we reason with uncertain information?";
\emph{economics}, for questions like,
"How should we make decisions so as to maximize payoff?",
"How should we do this when others may not go along?", and,
"How should we do this when the payoff may be far in the future?"; 
\emph{neuroscience}, for questions like,
"How do brains process information?";
\emph{psychology}, for questions like,
"How do humans and animals think and act?";
\emph{computer engineering}, for questions like,
"How can we build an efficient computer?";
\emph{control theory and cybernetics}, for questions like,
"How can artifacts operate under their own control?";
and, 
\emph{linguistics}, for questions like,
"How does language relate to thought?"

\citet{Russell2009} present the history of AI as,
the gestation of artificial intelligence (1943-1955),
the birth of artificial intelligence (1956),
early enthusiasm, great expectations (1952 - 1969),
a dose of reality (1966 - 1973),
knowledge-based systems: the key to power? (1969 - 1979),
AI becomes an industry (1980 - present),
the return of neural networks (1986 - present),
AI adopts the scientific method (1987 - present), 
the emergence of intelligent agents (1995 - present), and,
the availability of very large data sets (2001 -present).

\subsection{Machine Learning}
\label{machinelearning}

Machine learning is about learning from data and making predictions and/or decisions. 
Quoting from \citet{Mitchell1997},
"A computer program is said to learn from experience $E$ with respect to some class of tasks $T$ and performance measure $P$ if its performance at tasks in $T$, as measured by $P$, improves with experience $E$."

Usually we categorize machine learning as supervised, unsupervised, and reinforcement learning.
In supervised learning, there are labeled data; in unsupervised learning, there are no labeled data.
Classification and regression are two types of supervised learning problems, with categorical and numerical outputs respectively.

Unsupervised learning attempts to extract information from data without labels, e.g., clustering and density estimation. Representation learning is a classical type of unsupervised learning. However, training  deep neural networks with supervised learning is a kind of representation learning.
Representation learning finds a representation to preserve as much information about the original data as possible, 
and, at the same time, to keep the representation simpler or more accessible than the original data, with low-dimensional, sparse, and independent representations. 

Deep learning, or deep neural networks, is a particular machine learning scheme, usually for supervised or unsupervised learning, and can be integrated with reinforcement learning, for state representation and/or function approximator.  
Supervised and unsupervised learning are usually one-shot, myopic, considering instant rewards; while reinforcement learning is sequential, far-sighted, considering long-term accumulative rewards. 

Reinforcement learning is usually about sequential decision making.
In reinforcement learning, in contrast to supervised learning and unsupervised learning, 
there are evaluative feedbacks, but no supervised labels.
Comparing with supervised learning, reinforcement learning has additional challenges like  credit assignment, stability, and, exploration.
Reinforcement learning is kin to optimal control~\citep{Bertsekas12, Sutton1992control}, and operations research and management~\citep{Powell11}, and is also related to psychology and neuroscience~\citep{Sutton2018}. 

Machine learning is the basis for big data, data science~\citep{Blei2017,Provost2013}, predictive modeling~\citep{Kuhn2013}, data mining~\citep{Han2011DM}, information retrieval~\citep{Manning2008}, etc, and becomes a critical ingredient for computer vision, natural language processing, robotics, etc. 
Probability theory and statistics~\citep{Hastie2009,Murphy2012, Vapnik1998} 
and optimization~\citep{Boyd04, Bottou2018} are important for statistical machine learning.
Machine learning is a subset of AI; however, it is evolving to be critical for all fields of AI.



A machine learning algorithm is composed of a dataset, a cost/loss function, an optimization procedure, and a model~\citep{Goodfellow2016}. 
A dataset is divided into non-overlapping training, validation, and testing subsets. 
A cost/loss function measures the model performance, e.g., with respect to accuracy, like mean square error in regression and classification error rate. 

The concept of entropy is important for the definition of loss functions.
For a random variable $X$ with distribution $p$, 
the entropy is a measure of its uncertainty, denoted by $\mathbb{H}(X)$ or  $\mathbb{H}(p)$,
\begin{equation}
\mathbb{H}(X) \triangleq - \sum_{k=1}^K p(X=k) \log_2 p(X=k).
\end{equation}
For binary random variables, $X \in \{0,1\}$, we have $p(X=1) = \theta$ and $p(X=0) = 1-\theta$, 
\begin{equation}
\mathbb{H}(X) = -[\theta \log_2\theta + (1-\theta) \log_2(1-\theta)].
\end{equation}
Kullback-Leibler divergence (KL-divergence) or relative entropy,
is one way to measure the dissimilarity of two probability distributions, $p$ and $q$,
\begin{equation}
\mathbb{KL}(p\|q) \triangleq  \sum_{k=1}^K p_k \log \frac{p_k}{q_k} \triangleq  \sum_{k=1}^K p_k \log p_k  - p_k \log q_k = -\mathbb{H}(p) + \mathbb{H}(p, q).
\end{equation}
Here $\mathbb{H}(p, q)$ is the cross entropy.
We have $\mathbb{KL}(p,q) \triangleq  - \sum_{k=1}^K p_k \log q_k$. See~\citet{Murphy2012}.


Training error measures the error on the training data, minimizing which is an optimization problem.
Generalization error, or test error, measures the error on new input data, which differentiates machine learning from optimization.
A machine learning algorithm tries to make the training error, and the gap between training error and testing error small.
A model is under-fitting if it can not achieve a low training error; a model is over-fitting if the gap between training error and test error is large.

A model's capacity measures the range of functions it can fit.
Ockham's razor\footnote{"Occam's razor" is a popular misspelling~\citep{Russell2009}.} states that, with the same expressiveness, simple models are preferred. 
Training error and generalization error vs. model capacity usually form a U-shape relationship. We find the optimal capacity to achieve low training error and small gap between training error and generalization error.
Bias measures the expected deviation of the estimator from the true value; while variance measures the deviation of the estimator from the expected value, or variance of the estimator. 
As model capacity increases, bias tends to decrease, while variance tends to increase, yielding another U-shape relationship between generalization error vs. model capacity. 
We try to find the optimal capacity point, of which under-fitting occurs on the left and over-fitting occurs on the right. 
Regularization adds a penalty term to the cost function, to reduce the generalization error, but not training error.
No free lunch theorem states that there is no universally best model, or best regularizer. 
An implication is that deep learning may not be the best model for some problems.
There are model parameters, and hyperparameters for model capacity and regularization.
Cross-validation is used to tune hyperparameters, to strike a balance between bias and variance, and to select the optimal model.

Maximum likelihood estimation (MLE) is a common approach to derive good estimation of parameters. 
For issues like numerical underflow, the product in MLE is converted to summation to obtain negative log-likelihood (NLL). 
MLE is equivalent to minimizing KL divergence, the dissimilarity between the empirical distribution defined by the training data and the model distribution.
Minimizing KL divergence between two distributions corresponds to minimizing the cross-entropy between the distributions. 
In short, maximization of likelihood becomes minimization of the negative log-likelihood (NLL), or equivalently, minimization of cross entropy.

Gradient descent is a common way to solve optimization problems. 
Stochastic gradient descent extends gradient descent by working with a single sample each time, and usually with minibatches.

Importance sampling is a technique to estimate properties of a particular distribution, by samples from a different distribution, to lower the variance of the estimation, or when sampling from the distribution of interest is difficult.



There are mathematical analysis frameworks for machine learning algorithms.
Kolmogorov complexity~\citep{Li2008Kolmogorov}, or algorithmic complexity,  
studies the notion of simplicity in Ockham's razor.
In probably approximately correct (PAC) learning~\citep{Valiant1984}, 
a learning algorithm aims to select a generalization function,  to achieve low generalization error with high probability.
VC dimension~\citep{Vapnik1998} measures the capacity of a binary classifier.

\subsection{Deep Learning}
\label{deeplearning}

Deep learning is in contrast to "shallow" learning. For many machine learning algorithms, e.g., linear regression, logistic regression, support vector machines (SVMs), decision trees, and boosting, we have input layer and output layer, and the inputs may be transformed with manual feature engineering before training. 
In deep learning, between input and output layers, we have one or more hidden layers. At each layer except the input layer, we compute the input to each unit, as the weighted sum of units from the previous layer; then we usually use nonlinear transformation, or activation function, such as logistic, tanh, or more popularly recently, rectified linear unit (ReLU), to apply to the input of a unit, to obtain a new representation of the input from the previous layer. 
We have weights on links between units from layer to layer. 
After computations flow forward from input to output, at the output layer and each hidden layer, we can compute error derivatives backward, and backpropagate gradients towards the input layer, so that weights can be updated to optimize some loss function.  

A feedforward deep neural network or multilayer perceptron (MLP) is to map a set of input values to output values
with a mathematical function formed by composing many simpler functions at each layer.  
A convolutional neural network (CNN) is a feedforward deep neural network, with convolutional layers, pooling layers, and, fully connected layers. CNNs are designed to process data with multiple arrays, e.g., colour image, language, audio spectrogram, and video, 
benefiting from the properties of such signals: local connections, shared weights, pooling, and, the use of many layers, and are inspired by simple cells and complex cells in visual neuroscience~\citep{LeCun2015}.  
ResNets~\citep{He2016-ResNets} are designed to ease the training of very deep neural networks by adding shortcut connections to learn residual functions with reference to the layer inputs.
A recurrent neural network (RNN) is often used to process sequential inputs like speech and language, element by element, with hidden units to store history of past elements. A RNN can be seen as a multilayer neural network with all layers sharing the same weights, when being unfolded in time of forward computation. It is hard for RNN to store information for very long time and the gradient may vanish. 
Long short term memory networks (LSTM)~\citep{Hochreiter1997} and gated recurrent unit (GRU)~\citep{Chung2014}  are proposed to address such issues, with gating mechanisms to manipulate information through recurrent cells. Gradient backpropagation or its variants can be used for training all deep neural networks mentioned above.


Dropout~\citep{Srivastava2014} is a regularization strategy to train an ensemble of sub-networks by removing non-output units randomly from the original network. Batch normalization~\citep{Ioffe2015, Santurkar2018NIPS} and layer normalization~\citep{Ba2016Layer} are designed to improve training efficiency. 

Deep neural networks learn representations automatically from raw inputs to recover the compositional hierarchies in many natural signals, i.e., higher-level features are composed of lower-level ones, e.g., in images, the hierarch of objects, parts, motifs, and local combinations of edges. Distributed representation is a central idea in deep learning, which implies that many features may represent each input, and each feature may represent many inputs. The exponential advantages of deep, distributed representations combat the exponential challenges of the curse of dimensionality. The notion of end-to-end training refers to that a learning model uses raw inputs, usually without manual feature engineering, to generate outputs, e.g., 
AlexNet~\citep{Krizhevsky2012} with raw pixels for image classification, 
\citet{Graves2013} with a Fourier transformation of audio data for speech recognition;
Seq2Seq~\citep{Sutskever2014} with raw sentences for machine translation, 
and DQN~\citep{Mnih-DQN-2015} with raw pixels and score to play games. 

There are efforts to design new neural network architectures, like capsules~\citep{Sabour2017, Hinton2018}.
There are also efforts to design deep machine learning architectures without neural networks, like DeepForest~\citep{Zhou2017, Feng2017}.

\subsection{Reinforcement Learning}
\label{RL}

We provide background of reinforcement learning briefly in this section, as a mini tutorial. 
It is essential to have a good understanding of reinforcement learning, 
for a good understanding of deep reinforcement learning. 
Deep RL is a particular type of RL, with deep neural networks for state representation and/or function approximation for value function, policy, transition model, or reward function.
\citet{Silver2015Course} is a clear introductory material for reinforcement learning. 

\citet{Sutton2018} present Bibliographical and Historical Remarks at the end of each chapter.
\citet{Russell2009} present Bibliographical and Historical Notes in Chapter 21 Reinforcement Learning.
See \citet{Barto2018} for a talk about a brief history of reinforcement learning.

We first  explain some terms in RL parlance. 
These terms would become clearer after reading the rest of this chapter.
We put them collectively here to make it convenient for readers to check them.

The prediction problem, or policy evaluation, is to compute the state or action value function for a policy. The control problem is to find the optimal policy. Planning constructs a value function or a policy with a model. 

On-policy methods evaluate or improve the behaviour policy, e.g., SARSA fits the action-value function to the current policy, i.e., SARSA evaluates the policy based on samples from the same policy, then refines the policy greedily with respect to action values. In off-policy methods, an agent learns an optimal value function/policy, maybe following an unrelated behaviour policy. For instance, Q-learning attempts to find action values for the optimal policy directly, not necessarily fitting to the policy generating the data, i.e., the policy Q-learning obtains is usually different from the policy that generates the samples.  The notion of on-policy and off-policy can be understood as same-policy and different-policy.

The exploration vs exploitation dilemma is about the agent needs to exploit the currently best action to maximize rewards greedily, yet it has to explore the environment to find better actions, when the policy is not optimal yet, or the system is non-stationary.  

In model-free methods, the agent learns with trail-and-error from experience directly; 
the model, i.e., for state transition and reward, is not known. RL methods that use models are model-based methods;  the model may be given, e.g. in the game of computer Go, or learned from experience. 

In an online mode, training algorithms are executed on data acquired in sequence. In an offline mode, or a batch mode, models are trained on the entire data set. 

With bootstrapping, an estimate of state or action value is updated from subsequent estimates.

\subsubsection{Problem Setup}


An RL agent interacts with an environment over time. At each time step $t$, the agent receives a state $s_t$ in a state space $\mathcal{S}$, and selects an action $a_t$ from an action space $\mathcal{A}$, following a policy $\pi(a_t|s_t)$, which is the agent's behavior, i.e., a mapping from state $s_t$ to actions $a_t$.
The agent receives a scalar reward $r_t$, and transitions to the next state $s_{t+1}$, according to the environment dynamics, or model, for reward function $\mathcal{R}(s,a)$, and, state transition probability $\mathcal{P}(s_{t+1}|s_t, a_t)$, respectively. In an episodic problem, this process continues until the agent reaches a terminal state and then it restarts. The return is the discounted, accumulated reward with the discount factor $\gamma \in (0,1]$,
\begin{equation}
R_t = \sum_{k=0}^{\infty} \gamma^k r_{t+k}.
\end{equation}
The agent aims to maximize the expectation of such long term return from each state.  The problem is set up in discrete state and action spaces. It is not hard to extend it to continuous spaces.   
In partially observable environments, an agent can not observe states fully, but has observations.

When an RL problem satisfies the Markov property, i.e., the future depends only on the current state and action, but not on the past, it is formulated as a Markov decision process (MDP), defined by the 5-tuple $(\mathcal{S}, \mathcal{A}, \mathcal{P},\mathcal{R}, \gamma)$. When the system model is available, we use dynamic programming methods: policy evaluation to calculate value/action value function for a policy, value iteration and policy iteration for finding an optimal policy. When there is no model, we resort to RL methods. RL methods also work when the model is available. An RL environment can be a multi-armed bandit, an MDP, a partially observable MDP (POMDP), a game, etc.

\subsubsection{Value Function}

A value function is a prediction of the expected, accumulative, discounted, future reward, measuring how good each state, or state-action pair, is. 
The state value, 
\begin{equation}
v_{\pi}(s) = \mathbb{E}[R_t | s_t = s], \mbox{ where, }R_t = \sum_{k=0}^{\infty} \gamma^k r_{t+k},
\end{equation}
is the expected return for following policy $\pi$ from state $s$.
The action value, 
\begin{equation}
q_{\pi}(s, a) = \mathbb{E}[R_t | s_t = s, a_t = a],
\end{equation}
is the expected return for selecting action $a$ in state $s$ and then following policy $\pi$. 
Value function $v_{\pi}(s)$ decomposes into the Bellman equation: 
\begin{equation}
v_{\pi}(s) = \sum_a \pi(a|s) \sum_{s',r}p(s',r|s,a)[r + \gamma v_{\pi}(s')].
\label{eqn:BellmanV}
\end{equation}
An optimal state value, 
\begin{equation}
v_{*}(s) = \max_{\pi} v_{\pi}(s) = \max_{a}q_{\pi ^{*}}(s,a),
\end{equation}
 is the maximum state value achievable by any policy for state $s$,
which decomposes into the Bellman equation: 
\begin{equation}
v_{*}(s) = \max_a \sum_{s',r}p(s',r|s,a)[r + \gamma v_{*}(s')].
\end{equation}
Action value function $q_{\pi}(s, a)$ decomposes into the Bellman equation: 
\begin{equation}
q_{\pi}(s, a) = \sum_{s',r}p(s',r|s,a)[r + \gamma \sum_{a'} \pi(a'|s')q_{\pi}(s', a')].
\end{equation}
An optimal action value function, 
\begin{equation}
q_{*}(s, a) = \max_{\pi} q_{\pi}(s, a),
\end{equation}
 is the maximum action value achievable by any policy for state $s$ and action $a$,
which decomposes into the Bellman equation: 
\begin{equation}
q_{*}(s, a) = \sum_{s',r}p(s',r|s,a)[r + \gamma \max_{a'}q_{*}(s', a')].
\end{equation}
We denote an optimal policy by $\pi^{*}$. 

Consider the shortest path problem as an example.
In graph theory, the single-source shortest path problem  is to find the shortest path between a pair of nodes so that the sum of weights of edges in the path is minimized. 
In RL, the state is the current node.
At each node, following the link to each neighbour is an action.
The transition model indicates that, after choosing a link to follow, the agent goes to a neighbour.
The reward is then the negative of link weight/cost/distance.
The discount factor can be $\gamma = 1$, since it is an episodic task.
The goal is to find a path to maximize the negative of the total cost, i.e., to minimize the total distance.
An optimal policy is to choose the best neighbour to traverse to achieve the shortest path;
and, for each state/node, an optimal value is the shortest distance from that node to the destination.
Dijkstra's algorithm is an efficient algorithm, with the information of the graph, including nodes, edges, and weights. RL can work in a model-free approach, by wandering in the graph according to some policy, without such global graph information. 
RL algorithms are more general than Dijkstra's algorithm, 
although with global graph information, Dijkstra's algorithm is very efficient.

\subsubsection{Exploration vs. Exploitation}

An RL agent needs to trade off between exploration of uncertain policies and exploitation of the current best policy, a fundamental dilemma in RL.
Here we introduce a simple approach, $\epsilon$-greedy, where $\epsilon \in (0,1)$, usually a small value close to 0.
In $\epsilon$-greedy, an agent selects a greedy action $a = \argmax_{a \in \mathcal{A}} Q(s,a)$, 
for the current state $s$, with probability $1-\epsilon$, 
and, selects a random action with probability $\epsilon$.
That is, the agent exploits the current value function estimation with probability $1-\epsilon$, 
and explores with probability $\epsilon$.

We will discuss more about the exploration vs exploitation dilemma in Chapter~\ref{exploration}, 
including several principles:
naive methods such as $\epsilon$-greedy,
optimistic initialisation,
upper confidence bounds,
probability matching,
and, information state search,
which are developed in the settings of multi-armed bandit,
but are applicable to RL problems~\citep{Silver2015Course}.

\subsubsection{Dynamic Programming}

Dynamic programming (DP) is a general method for problems with optimal substructure and overlapping subproblems.
MDPs satisfy these properties, where,
Bellman equation gives recursive decomposition, 
and, value function stores and reuses sub-solutions~\citep{Silver2015Course}.
DP assumes full knowledge of the transition and reward models of the MDP. 
The prediction problem is to evaluate the value function for a given policy,
and the control problem is to find an optimal value function and/or an optimal policy.

Iterative policy evaluation is an approach to evaluate a given policy $\pi$.
It iteratively applies Bellman expectation backup, 
\begin{equation}
v_{k+1}(s) = \sum_{a \in \mathcal{A}} \pi(a|s) [ \mathcal{R}(s,a) + \gamma \sum_{s' \in \mathcal{S}} \mathcal{P}(s'|s, a) v_k(s')],
\end{equation}
so that at each iteration $k + 1$,
for all states $s \in  \mathcal{S}$, update $v_{k+1}(s)$ from value functions of its successor states $v_k(s')$.
The value function will converge to $v_\pi$, the value function of the policy $\pi$.


Policy iteration (PI) alternates between policy evaluation and policy improvement, to generate a sequence of improving policies. 
In policy evaluation, the value function of the current policy is estimated to obtain $v_\pi$. 
In policy improvement, the current value function is used to generate a better policy, e.g., by selecting actions greedily with respect to the value function $v_\pi$.
This policy iteration process of iterative policy evaluation and greedy policy improvement
will converge to an optimal policy and value function.

We may modify the policy iteration step, stopping it before convergence.
A generalized policy iteration (GPI) is composed of any policy evaluation method and any policy improvement method.

Value iteration (VI) finds an optimal policy.
It iteratively applies Bellman optimality backup,
\begin{equation}
v_*(s) = \max_a  \mathcal{R}(s,a) + \gamma \sum_{s' \in \mathcal{S}} \mathcal{P}(s'|s, a) v_k(s').
\end{equation}
At each iteration $k + 1$, it updates $v_{k+1}(s)$ from $v_k(s')$, for all states $s \in \mathcal{S}$.
Such synchronous backup will converge to the value function of an optimal policy.
We may have asynchronous DP, and approximate DP.

We have Bellman equation for value function in Equation (\ref{eqn:BellmanV}). 
Bellman operator is defined as, 
\begin{equation}
(T^{\pi}v)(s) \doteq \sum_a \pi(a|s) \sum_{s',r}p(s',r|s,a)[r + \gamma v_{\pi}(s')].
\end{equation}
TD fix point is then, 
\begin{equation}
v_{\pi} = T^{\pi}v_{\pi}.
\end{equation}
We can define the Bellman expectation backup operator, in matrix forms,
\begin{equation}
T^\pi(v) = \mathcal{R} + \gamma \mathcal{P}^\pi v,
\end{equation} 
and the Bellman optimality backup operator, 
\begin{equation}
T^*(v) = \max_{a\in \mathcal{A}} \mathcal{R}^a + \gamma \mathcal{P}^a v,
\end{equation}
We can show that these operators are contractions with fixed points, respectively,
which help prove the convergence of policy iteration, value iteration, and certain general policy iteration algorithms.
See \citet{Silver2015Course}, \citet{Sutton2018}, and \citet{Bertsekas96} for more details.

\subsubsection{Monte Carlo}

Monte Carlo methods learn from complete episodes of experience, 
not assuming knowledge of transition nor reward models, 
and use sample means for estimation.
Monte Carlo methods are applicable only to episodic tasks.

With Monte Carlo methods for policy evaluation, we use empirical mean return rather than expected return for the evaluation.
By law of large numbers, the estimated value function converges to the value function of the policy. 

On-policy Monte Carlo control follows a generalized policy iteration scheme.
For policy evaluation, it uses Monte Carlo policy evaluation for the action value.
For policy improvement, it uses $\epsilon$-greedy policy improvement.
It can be shown that Greedy in the limit with infinite exploration (GLIE) Monte-Carlo control converges to the optimal action-value function~\citep{Singh2000}.

In off-policy learning, we evaluate a target policy, following a behaviour policy.
With off-policy, we can learn with observations from humans or other agents,
reuse experience from old policies,
learn an optimal policy while following an exploratory policy,
and, learn multiple policies based on experience of one policy.~\citep{Silver2015Course}

We can use importance sampling for off-policy Monte Carlo methods,
by multiply importance sampling correction weights along the whole episode,
to evaluate the target policy with experience generated by the behaviour policy.
This may increase variance dramatically though.

\subsubsection{Temporal Difference Learning}

Temporal difference (TD) learning is central in RL. TD learning usually refers to the learning methods for value function evaluation in~\citet{Sutton1988}. Q-learning~\citep{Watkins1992} and SARSA~\citep{Rummery1994}  are temporal difference control methods. 

TD learning~\citep{Sutton1988} learns value function $V(s)$ directly from experience with TD error, with bootstrapping, in a model-free, online, and fully incremental way.  TD learning is a prediction problem. The update rule is, 
\begin{equation}
V(s) \leftarrow V(s) + \alpha [r + \gamma V(s') - V(s)],
\end{equation}
 where $\alpha$ is a learning rate, and $r + \gamma V(s') - V(s)$ is called the TD error. 
 Algorithm~\ref{TD-algo} presents the pseudo code for tabular TD learning. 
 Precisely, it is tabular TD(0) learning, where "0" indicates it is based on one-step returns.

Bootstrapping estimates state or action value function based on subsequent estimates as in the TD update rule, and is common in RL, like in TD learning, Q-learning, and SARSA.
 Bootstrapping methods are usually faster to learn, and enable learning to be online and continual. 
 Bootstrapping methods are not instances of true gradient decent, 
 since the target depends on the values to be estimated. 
 The concept of semi-gradient descent is then introduced~\citep{Sutton2018}.  

\begin{algorithm}[h]
\SetAlgoNoLine
\hrulefill

\textbf{Input: } the policy $\pi$ to be evaluated\\
\textbf{Output: } value function $V$\\
\hrulefill

initialize $V$ arbitrarily, e.g., to 0 for all states\\
\For{each episode}{
	initialize state $s$\\
	\For{each step of episode, state $s$ is not terminal}{
	        $a \leftarrow $ action given by $\pi$ for $s$\\
	        take action $a$, observe $r$, $s'$\\
	        $V(s) \leftarrow V(s) + \alpha [r + \gamma V(s') - V(s)]$\\
	        $s \leftarrow s'$
	}
}
\hrulefill
\caption{TD learning, adapted from \citet{Sutton2018}}
\label{TD-algo}
\end{algorithm}

\begin{algorithm}[h]
\SetAlgoNoLine

\hrulefill

\textbf{Output: } action value function $Q$

\hrulefill

initialize $Q$ arbitrarily, e.g., to 0 for all states, set action value for terminal states as 0\\
\For{each episode}{
	initialize state $s$\\
	\For{each step of episode, state $s$ is not terminal}{
	        $a \leftarrow $ action for $s$ derived by $Q$, e.g., $\epsilon$-greedy\\
	        take action $a$, observe $r$, $s'$\\
	        $a' \leftarrow $ action for $s'$ derived by $Q$, e.g., $\epsilon$-greedy\\
	        $Q(s, a) \leftarrow Q(s, a) + \alpha [r + \gamma Q(s', a') - Q(s,a)]$\\
	        $s \leftarrow s'$, $a \leftarrow a'$
	}
}

\hrulefill

\caption{SARSA, adapted from \citet{Sutton2018}}
\label{SARSA-algo}
\end{algorithm}

\begin{algorithm}[H]
\SetAlgoNoLine

\hrulefill

\textbf{Output: } action value function $Q$\\

\hrulefill

initialize $Q$ arbitrarily, e.g., to 0 for all states, set action value for terminal states as 0\\
\For{each episode}{
	initialize state $s$\\
	\For{each step of episode, state $s$ is not terminal}{
	        $a \leftarrow $ action for $s$ derived by $Q$, e.g., $\epsilon$-greedy\\
	        take action $a$, observe $r$, $s'$\\
	        $Q(s, a) \leftarrow Q(s, a) + \alpha [r + \gamma  \max_{a'} Q(s', a') - Q(s, a)]$\\
	        $s \leftarrow s'$
	}
}

\hrulefill
\caption{Q-learning, adapted from \citet{Sutton2018}}
\label{Q-algo}
\end{algorithm}

SARSA, representing state, action, reward, (next) state, (next) action, is an on-policy control method to find an optimal policy, with the update rule, 
\begin{equation}
Q(s, a) \leftarrow Q(s, a) + \alpha [r + \gamma Q(s', a') - Q(s,a)].
\end{equation}
Algorithm~\ref{SARSA-algo} presents the pseudo code for tabular SARSA, i.e., tabular SARSA(0).

Q-learning is an off-policy control method to find the optimal policy. Q-learning learns the action value function, with the update rule, 
\begin{equation}
Q(s, a) \leftarrow Q(s, a) + \alpha [r + \gamma  \max_{a'} Q(s', a') - Q(s, a)].
\end{equation}
 Q-learning refines the policy greedily  w.r.t. action values by the max operator. Algorithm~\ref{Q-algo} presents the pseudo code for Q-learning, precisely, tabular Q(0) learning.

TD-learning, Q-learning and SARSA converge under certain conditions. From an optimal action value function, we can derive an optimal policy. 

\subsubsection{Multi-step Bootstrapping}

The above algorithms are referred to as TD(0), Q(0) and SARSA(0), learning with one-step return. We have variants with multi-step return for them in the forward view.  
In $n$-step update, $V(s_t)$ is updated toward the $n$-step return, defined as, 
\begin{equation}
r_t + \gamma r_{t+1} + \cdots + \gamma^{n-1} r_{t+n-1} + \gamma^n V(s_{t+n}).
\end{equation}
The eligibility trace from the backward view provides an online, incremental implementation, resulting in TD($\lambda$), Q($\lambda$) and SARSA($\lambda$) algorithms, where $\lambda \in[0,1]$. TD(1) is the same as the Monte Carlo approach.
Eligibility trace is a short-term memory, usually lasting within an episode, assists the learning process, by affecting the weight vector. The weight vector is a long-term memory, lasting the whole duration of the system, determines the estimated value. Eligibility trace helps with the issues of long-delayed rewards and non-Markov tasks~\citep{Sutton2018}.

TD($\lambda$) unifies one-step TD prediction, TD(0), with Monte Carlo methods, TD(1), using eligibility traces and the decay parameter $\lambda$, for prediction algorithms.
\citet{DeAsis2018} make unification for multi-step TD control algorithms. 

\subsection*{Comparison of DP, MC, and TD}

In the following, we compare dynamic programming (DP), Monte Carlo (MC), and temporal difference (TD) learning, 
based on~\citet{Silver2015Course} and \citet{Sutton2018}.

DP requires the model; TD and MC are model-free.
DP and TD bootstrap; MC does not.
TD and MC work with sample backups, DP and exhaustive search work with full backups.
DP works with one step backups; 
TD works with one or multi-step backups;
MC and exhaustive search work with deep backups, until the episode terminates.

TD can learn online, from incomplete sequences, in continuous environments.
MC learns from complete sequences, in episodic environments.
TD has low variance, some bias, usually more efficient than MC, more sensitive to initialization.
TD(0) converges to value function of a policy, may diverge with function approximation.
MC has high variance, zero bias, simple to understand and use, insensitive to initilization.
MC has good convergence properties, even with function approximation.

TD exploits Markov property, so it is more efficient in Markov environments.
MC does not assume Markov property, so it is usually more effective in non-Markov environments.

Table~\ref{table-DPTD} compare DP with TD~\citep{Silver2015Course, Sutton2018}.
\begin{table}[h]
\centering
\begin{tabular}{| l | l | }
\hline
\hline
Full Backup (DP) & Sample Backup \\ \hline
iterative policy evaluation & TD learning \\
$v(s) \leftarrow \mathbb{E}[r + \gamma v(s') | s]$ & $v(s) \overset{\alpha}{\leftarrow} r+ \gamma v(s')$ \\ \hline
Q policy iteration & SARSA \\
$Q(s,a) \leftarrow \mathbb{E}[r + \gamma Q(s',a') | s,a]$ & $Q(s,a) \overset{\alpha}{\leftarrow} r+ \gamma Q(s',a')$ \\ \hline
Q value iteration & Q-learning \\
$Q(s,a) \leftarrow \mathbb{E}[r + \gamma \max_{a' \in \mathcal{A}}Q(s',a') | s,a]$ & $Q(s,a) \overset{\alpha}{\leftarrow} r+ \gamma \max_{a' \in \mathcal{A}}Q(s',a')$ \\ \hline
\hline
\end{tabular}
\caption{Comparison between DP and TD Learning, where $x \overset{\alpha}{\leftarrow} y \doteq x \leftarrow \alpha(y-x)$.}
\label{table-DPTD}
\end{table}

\subsubsection{Model-based RL}

\citet{Sutton1990} proposes Dyna-Q to integrate learning, acting, and planning, by not only learning from real experience, but also planning with simulated trajectories from a learned model. Learning uses real experience from the environment; and planning uses experience simulated by a model. Algorithm~\ref{Dyna-algo} presents the pseudo code for tabular Dyna-Q. We will discuss more about model-based RL in Chapter~\ref{model}.

\begin{algorithm}[h]
\SetAlgoNoLine
\hrulefill

// $Model(s,a)$ denotes predicted reward and next state for state action pair $(s,a)$ \\
initialize $Q(s,a)$ and $Model(s,a)$ for all $s \in \mathcal{S}$ and $a \in \mathcal{A}$\\
\For{true}{
	$s \leftarrow $ current, nonterminal, state\\
	$a \leftarrow $ action for $s$ derived by $Q$, e.g., $\epsilon$-greedy\\
	//acting\\
	take action $a$; observe reward $r$, and next state $s'$\\
	// direct reinforcement learning\\
	$Q(s, a) \leftarrow Q(s, a) + \alpha [r + \gamma  \max_{a'} Q(s', a') - Q(s, a)]$\\
	// model learning\\
	$Model(s,a) \leftarrow r, s'$\\ 
	// planning\\
	\For{$N$ iterations}{
	        $s \leftarrow$ random state previously observed\\ 
	        $a \leftarrow$ random action previously taken\\ 
	        $r, s' \leftarrow Model(s,a)$\\
	        $Q(s, a) \leftarrow Q(s, a) + \alpha [r + \gamma  \max_{a'} Q(s', a') - Q(s, a)]$
	}
}

\hrulefill
\caption{Dyna-Q, adapted from \citet{Sutton2018}}
\label{Dyna-algo}
\end{algorithm}

\subsubsection{Function Approximation}
\label{functionapproximation}

We discuss tabular cases above, where a value function or a policy is stored in a tabular form. Function approximation is a way for generalization when the state and/or action spaces are large or continuous. Function approximation aims to generalize from examples of a function to construct an approximate of the entire function; it is usually a concept in supervised learning, studied in the fields of machine learning, patten recognition, and statistical curve fitting; function approximation in reinforcement learning usually treats each backup as a training example, and encounters new issues like nonstationarity, bootstrapping, and delayed targets~\citep{Sutton2018}. Linear function approximation is a popular choice, partially due to its desirable theoretical properties, esp. before the work of deep Q-network~\citep{Mnih-DQN-2015}.  However, the integration of reinforcement learning and neural networks dates back a long time ago~\citep{Sutton2018,Bertsekas96, Schmidhuber2015-DL}. 

Algorithm~\ref{TD-FA-algo} presents the pseudo code for TD(0) with function approximation. $\hat{v}(s, \bm{w})$ is the approximate value function, $\bm{w}$ is the value function weight vector, $\nabla \hat{v}(s, \bm{w})$ is the gradient of the approximate value function w.r.t. the weight vector, which is updated following the update rule, 
\begin{equation}
\bm{w} \leftarrow \bm{w} + \alpha [r + \gamma  \hat{v}(s', \bm{w}) - \hat{v}(s, \bm{w})] \nabla \hat{v}(s, \bm{w}). 
\end{equation}

\begin{algorithm}[h]
\SetAlgoNoLine

\hrulefill

\textbf{Input: } the policy $\pi$ to be evaluated\\
\textbf{Input: } a differentiable value function $\hat{v}(s, \bm{w})$, $\hat{v}(terminal, \cdot) = 0$\\
\textbf{Output: } value function $\hat{v}(s, \bm{w})$\\

\hrulefill

initialize value function weight $\bm{w}$ arbitrarily, e.g., $\bm{w} = 0$\\
\For{each episode}{
	initialize state $s$\\
	\For{each step of episode, state $s$ is not terminal}{
	        $a \leftarrow \pi(\cdot|s)$\\
	        take action $a$, observe $r$, $s'$\\
	        $\bm{w} \leftarrow \bm{w} + \alpha [r + \gamma  \hat{v}(s', \bm{w}) - \hat{v}(s, \bm{w})] \nabla \hat{v}(s, \bm{w}) $\\
	        $s \leftarrow s'$
	}
}

\hrulefill
\caption{TD(0) with function approximation, adapted from \citet{Sutton2018}}
\label{TD-FA-algo}
\end{algorithm}



When combining off-policy, function approximation, and bootstrapping, instability and divergence may occur~\citep{Tsitsiklis97}, which is called the deadly triad issue~\citep{Sutton2018}. All these three elements are necessary: function approximation for scalability and generalization, bootstrapping for computational and data efficiency, and off-policy learning for freeing behaviour policy from target policy.  What is the root cause for the instability? 
Learning or sampling is not, since dynamic programming suffers from divergence with function approximation; 
exploration, greedification, or control is not, since prediction alone can diverge; 
local minima or complex non-linear function approximation is not, since linear function approximation can produce instability~\citep{Sutton2016Course}. 
It is unclear what is the root cause for instability -- each single factor mentioned above is not -- there are still many open problems in off-policy learning~\citep{Sutton2018}. 

Table~\ref{table-rl}  presents various algorithms that tackle various issues~\citep{Sutton2016Course}.  
ADP algorithms refer to approximate dynamic programming algorithms like policy evaluation, policy iteration, and value iteration, with function approximation. 
Least square temporal difference (LSTD)~\citep{Bradtke96} computes TD fix-point directly in batch mode. 
LSTD is data efficient, yet with squared time complexity. 
LSPE~\citep{Nedic2003} extends LSTD. 
Fitted-Q algorithms~\citep{Ernst2005, Riedmiller2005} learn action values in batch mode.
Residual gradient algorithms~\citep{Baird1995} minimize Bellman error. 
Gradient-TD~\citep{Sutton2009GTD-ICML, Sutton2009GTD-NIPS, Mahmood2014} methods are true gradient algorithms, perform SGD in the projected Bellman error (PBE), converge robustly under off-policy training and non-linear function approximation. 
Expected SARSA~\citep{vanSeijen2009} has the same convergence guarantee as SARSA, with lower variance.
Emphatic-TD~\citep{Sutton2016} emphasizes some updates and de-emphasizes others by reweighting, improving computational efficiency, yet being a semi-gradient method. See~\citet{Sutton2018} for more details. 
\citet{Du2017PE} propose variance reduction techniques for policy evaluation to achieve fast convergence. 
\citet{Liu2018PGTD} study proximal gradient TD learning.
\citet{White2016} perform empirical comparisons of linear TD methods, and make suggestions about their practical use.
\citet{JinChi2018NIPS} study the sample efficiency of Q-learning.
\citet{LuTyler2018NIPS} study non-delusional Q-learning and value-iteration.

\begin{table}[h]
\centering
\begin{tabular}{cc|c|c|c|c|c|c|l}
\cline{3-8}
& & \multicolumn{6}{ c| }{algorithm} \\ \cline{3-8}
&  & \shortstack{TD($\lambda$)\\ SARSA($\lambda$)} &  \shortstack{ADP} & \shortstack{LSTD($\lambda$)\\ LSPE($\lambda$)} & Fitted-Q & \shortstack{Residual\\ Gradient} &  \shortstack{GTD($\lambda$)\\ GQ($\lambda$)}\\ \cline{1-8}
\multicolumn{1}{ |c  }{ } &
\multicolumn{1}{ |c| }{\shortstack{linear\\ computation}} & $\checkmark$  & $\checkmark$ &  &  & $\checkmark$  &  $\checkmark$&    \\ \cline{2-8}
\multicolumn{1}{ |c  }{\multirow{4}{*}{\begin{turn}{-270}issue\end{turn}}}                        & 
\multicolumn{1}{ |c| }{\shortstack{nonlinear\\ convergent}} &  &  &  & $\checkmark$ & $\checkmark$& $\checkmark$&     \\ \cline{2-8}
\multicolumn{1}{ |c  }{}                        & 
\multicolumn{1}{ |c| }{\shortstack{off-policy\\ convergent}} &  &  &  $\checkmark$&  & $\checkmark$& $\checkmark$&     \\ \cline{2-8}
\multicolumn{1}{ |c  }{}                        & 
\multicolumn{1}{ |c| }{\shortstack{model-free,\\ online}} & $\checkmark$ &  & $\checkmark$ &  &$\checkmark$ & $\checkmark$&     \\ \cline{2-8}
\multicolumn{1}{ |c  }{}                        & 
\multicolumn{1}{ |c| }{\shortstack{converges to\\ PBE = 0}} &  $\checkmark$&  $\checkmark$&  $\checkmark$&  $\checkmark$& &  $\checkmark$&     \\ \cline{1-8}
\end{tabular}
\caption{RL Issues vs. Algorithms, adapted from \citet{Sutton2016Course}}
\label{table-rl}
\end{table}



\subsubsection{ Policy Optimization}
\label{bg:policy}

In contrast to value-based methods like TD learning and Q-learning, policy-based methods optimize the policy $\pi(a|s; \bm{\theta})$ (with function approximation) directly, and update the parameters $\bm{\theta}$ by gradient ascent.
Comparing with value-based methods, 
policy-based methods
usually have better convergence properties, 
are effective in high-dimensional or continuous action spaces,
and can learn stochastic policies.
However, policy-based methods
usually converge to local optimum,
are inefficient to evaluate,
and encounter high variance~\citep{Silver2015Course}. 
Stochastic policies are important since some problems have only stochastic optimal policies,
e.g., in the rock-paper-scissors game, 
an optimal policy for each player is to take each action (rock, paper, or scissors) with probability 1/3.

For a differentiable policy $\pi(a|s; \bm{\theta})$, 
we can compute the policy gradient analytically, whenever it is non-zero,
\begin{equation}
\nabla_{\bm{\theta}} \pi(a|s; \bm{\theta})  = \pi(a|s; \bm{\theta})  \frac{\nabla_{\bm{\theta}} \pi(a|s; \bm{\theta})}{\pi(a|s; \bm{\theta})}  = \pi(a|s; \bm{\theta})  \nabla_{\bm{\theta}} \log \pi(a|s; \bm{\theta})
\end{equation}
We call $\nabla_{\bm{\theta}} \log \pi(a|s; \bm{\theta})$ score function, or likelihood ratio.
Policy gradient theorem~\citep{Sutton2000} states that for a differentiable policy  $\pi(a|s; \bm{\theta})$, 
the policy gradient is,  
\begin{equation}
\mathbb{E}_{\pi_{\bm{\theta}}}[\nabla_{\bm{\theta}} \log \pi(a|s; \bm{\theta}) Q^{\pi_{\bm{\theta}}}(s,a)].
\end{equation}
We omit $\pi_{\bm{\theta}}$ in value functions below for simplicity.

REINFORCE~\citep{Williams1992} 
updates $\bm{\theta}$ in the direction of 
\begin{equation}
\nabla_{\bm{\theta}} \log \pi(a_t|s_t; \bm{\theta}) R_t,
\end{equation}
by using return $R_t$ as an unbiased sample of $Q(s_t,a_t)$.
Usually a baseline $b_t(s_t)$ is subtracted from the return to reduce the variance of gradient estimate, 
yet keeping its unbiasedness, to yield the gradient direction, 
\begin{equation}
\nabla_{\bm{\theta}} \log \pi(a_t|s_t; \bm{\theta}) (Q(a_t, s_t)- b_t(s_t)).
\end{equation} 
Using $V(s_t)$ as the baseline $b_t(s_t)$, we have the advantage function, 
\begin{equation}
A(a_t, s_t) = Q(a_t, s_t) - V(s_t).
\end{equation}
Algorithm~\ref{REINFORCE} presents the pseudo code for REINFORCE in the episodic case.

\begin{algorithm}[h]
\SetAlgoNoLine

\hrulefill

\textbf{Input: } policy $\pi(a|s, \bm{\theta})$, $\hat{v}(s, \bm{w})$ \\
\textbf{Parameters: } step sizes, $\alpha > 0$, $\beta > 0$\\
\textbf{Output: } policy $\pi(a|s, \bm{\theta})$\\

\hrulefill

initialize policy parameter $\bm{\theta}$ and state-value weights $\bm{w}$\\
\For{true}{
	generate an episode $s_0$, $a_0$, $r_1$,  $\cdots$, $s_{T-1}$, $a_{T-1}$, $r_T$, following $\pi(\cdot|\cdot, \bm{\theta})$\\
	\For{each step $t$ of episode 0, $\cdots$, $T-1$}{
	        $G_t \leftarrow $ return from step $t$\\
	        $\delta \leftarrow G_t - \hat{v}(s_t, \bm{w})$\\
	        	$\bm{w} \leftarrow \bm{w} + \beta \delta \nabla_{\bm{w}} \hat{v}(s_t, \bm{w}) $\\
		$\bm{\theta} \leftarrow \bm{\theta} + \alpha \gamma^t \bm{\delta} \nabla_{\bm{\theta}}  log \pi(a_t|s_t, \bm{\theta})$
	}
}

\hrulefill

\caption{REINFORCE with baseline (episodic), adapted from \citet{Sutton2018}}
\label{REINFORCE}
\end{algorithm}

In actor-critic algorithms, the critic updates action-value function parameters, and the actor updates policy parameters, in the direction suggested by the critic. 
Algorithm~\ref{Actor-Critic} presents the pseudo code for one-step actor-critic algorithm in the episodic case.

\begin{algorithm}[h]
\SetAlgoNoLine

\hrulefill

\textbf{Input: } policy $\pi(a|s, \bm{\theta})$, $\hat{v}(s, \bm{w})$ \\
\textbf{Parameters: } step sizes, $\alpha > 0$, $\beta > 0$\\
\textbf{Output: } policy $\pi(a|s, \bm{\theta})$\\

\hrulefill

initialize policy parameter $\bm{\theta}$ and state-value weights $\bm{w}$\\
\For{true}{
        initialize $s$, the first state of the episode\\
        $I \leftarrow 1$\\
	\For{$s$ is not terminal}{
	        $a \sim \pi(\cdot|s, \bm{\theta})$\\
	        take action $a$, observe $s'$, $r$\\
	        $\delta \leftarrow r + \gamma  \hat{v}(s', \bm{w}) - \hat{v}(s, \bm{w})$  (if $s'$ is terminal, $\hat{v}(s', \bm{w}) \doteq 0$)\\
	        	$\bm{w} \leftarrow \bm{w} + \beta \delta \nabla_{\bm{w}} \hat{v}(s_t, \bm{w}) $\\
		$\bm{\theta} \leftarrow \bm{\theta} + \alpha I \delta \nabla_{\bm{\theta}}  log \pi(a_t|s_t, \bm{\theta})$\\
		$I \leftarrow \gamma I$\\
		$s \leftarrow s'$
	}
}

\hrulefill
\caption{Actor-Critic (episodic), adapted from \citet{Sutton2018}}
\label{Actor-Critic}
\end{algorithm}

As summarized in \citet{Silver2015Course}, policy gradient may take various forms:
$\nabla_{\bm{\theta}} \log \pi(a|s; \bm{\theta}) R_t$ for REINFORCE,
$\nabla_{\bm{\theta}} \log \pi(a|s; \bm{\theta}) Q(s, a; \bm{w})$ for Q actor-critic,
$\nabla_{\bm{\theta}} \log \pi(a|s; \bm{\theta}) A(s, a; \bm{w})$ for advantage actor-critic,
$\nabla_{\bm{\theta}} \log \pi(a|s; \bm{\theta}) \delta$ for TD actor-critic,
$\nabla_{\bm{\theta}} \log \pi(a|s; \bm{\theta}) \delta e$ for TD($\lambda$) actor-critic, 
and $G_{\bm{\theta}} \bm{w}$ for natural gradient decent, 
where
$G_{\bm{\theta}} = \mathbb{E}_{\pi_{\bm{\theta}}}[\nabla_{\bm{\theta}} \log \pi(a|s; \bm{\theta}) \nabla_{\bm{\theta}} \log \pi(a|s; \bm{\theta})^T]$
is the Fisher information matrix;
and the critic may use Monte Carlo or TD learning for policy evaluation to estimate the value functions $Q$, $A$ or $V$.

\subsubsection{ Deep RL}

We obtain deep reinforcement learning (deep RL) methods when we use deep neural networks 
to represent the state or observation,
and/or to approximate any of the following components of reinforcement learning: value function, $\hat{v}(s; \bm{\theta})$ or $\hat{q}(s,a; \bm{\theta})$, policy $\pi(a|s; \bm{\theta})$, and model (state transition function and reward function). 
Here, the parameters $\bm{\theta}$ are the weights in deep neural networks. When we use "shallow" models, like linear function, decision trees, tile coding and so on as the function approximator, we obtain "shallow" RL, and the parameters $\bm{\theta}$ are the weight parameters in these models. Note, a shallow model, e.g., decision trees, may be non-linear. The distinct difference between deep RL and "shallow" RL is what function approximator is used. This is similar to the difference between deep learning and "shallow" machine learning. We usually utilize stochastic gradient descent to update weight parameters in deep RL. When off-policy, function approximation, in particular, non-linear function approximation, and bootstrapping are combined together, instability and divergence may occur~\citep{Tsitsiklis97}. However, recent work like deep Q-network~\citep{Mnih-DQN-2015} and AlphaGo~\citep{Silver-AlphaGo-2016} stabilize the learning and achieve outstanding results.
There are efforts for convergence proof of control with non-linear function approximation, e.g., \citet{Dai2018SBEED, Nachum2018TrustPCL}.

\subsubsection{ Brief Summary}

An RL problem is formulated as an MDP when the observation about the environment satisfies the Markov property. An MDP is defined by the 5-tuple $(\mathcal{S}, \mathcal{A}, \mathcal{P},\mathcal{R}, \gamma)$. A central concept in RL is value function. Bellman equations are cornerstone for developing RL algorithms. Temporal difference learning algorithms are fundamental for evaluating/predicting value functions.  Control algorithms find optimal policies. 
Policy-based methods become popular recently.
Reinforcement learning algorithms may be based on value function and/or policy, model-free or model-based, on-policy or off-policy, with function approximation or not, with sample backups (TD and Monte Carlo) or full backups (dynamic programming and exhaustive search), and about the depth of backups, either one-step return (TD(0) and dynamic programming) or multi-step return (TD($\lambda$), Monte Carlo, and exhaustive search). When combining off-policy, function approximation, and bootstrapping, we face instability and divergence~\citep{Tsitsiklis97}, the deadly triad issue~\citep{Sutton2018}. Theoretical guarantee has been established for linear function approximation, e.g., Gradient-TD~\citep{Sutton2009GTD-ICML, Sutton2009GTD-NIPS, Mahmood2014}, Emphatic-TD~\citep{Sutton2016} and \citet{Du2017PE}. 
When RL integrates with deep neural networks, either for representation or function approximation, we have deep RL. 
Deep RL algorithms like Deep Q-Network~\citep{Mnih-DQN-2015} and AlphaGo~\citep{Silver-AlphaGo-2016, Silver-AlphaGo-2017} stabilize the learning and achieve stunning results.






\subsection{Resources}
\label{resources}

We present some resources for deep RL in the following.   
We maintain a blog titled Resources for Deep Reinforcement Learning at \url{https://medium.com/@yuxili/}.


Sutton and Barto's RL book~\citep{Sutton2018} covers fundamentals and reflects new progress, e.g., in deep Q-network, AlphaGo, policy gradient methods, as well as in psychology and neuroscience. 
David Silver's RL course~\citep{Silver2015Course} and
Sergey Levine's Deep RL course~\citep{Levine2018}
are highly recommended.

\citet{Goodfellow2016} is a recent deep learning book.
\citet{Bishop2011},  \citet{Hastie2009}, and \citet{Murphy2012} are popular machine learning textbooks.
\citet{James2013} is an introduction book for machine learning.
\citet{Domingos2012}, \citet{Zinkevich2017}, and \citet{Ng2018} are about practical machine learning advices.
See \citet{Ng2016Nuts} and \citet{Schulman2017Nuts} for practical advices for deep learning and deep RL respectively.


There are excellent summer schools and bootcamps, e.g., 
Deep Learning and Reinforcement Learning Summer School: 
2018 at \url{https://dlrlsummerschool.ca},
2017 at \url{https://mila.umontreal.ca/en/cours/deep-learning-summer-school-2017/};
Deep Learning Summer School: 
2016 at \url{https://sites.google.com/site/deeplearningsummerschool2016/},
and, 2015 at \url{https://sites.google.com/site/deeplearningsummerschool/}; 
and, 
Deep RL Bootcamp: at \url{https://sites.google.com/view/deep-rl-bootcamp/}. 


Common benchmarks for general RL algorithms are
Atari games in  the Arcade Learning Environment (ALE) for discrete control,
and simulated robots using the MuJoCo physics engine in OpenAI Gym for continuous control.

The Arcade Learning Environment (ALE) \citep{Bellemare2013, Machado2017ALE} is a framework composed of Atari 2600 games to develop and evaluate AI agents.
OpenAI Gym, at \url{https://gym.openai.com}, is a toolkit for the development of RL algorithms, consisting of environments, e.g., Atari games and simulated robots, and a site for the comparison and reproduction of results. 
MuJoCo, Multi-Joint dynamics with Contact, at \url{http://www.mujoco.org}, is a physics engine.
DeepMind Lab~\citep{Beattie2016} is a first-person 3D game platform, at \url{https://github.com/deepmind/lab}.  
DeepMind Control Suite~\citep{Tassa2018} provides RL environments with the MuJoCo physics engine, at \url{https://github.com/deepmind/dm_control}.
Dopamine~\citep{Bellemare2018Dopamine} is  a Tensorflow-based RL framework from Google AI.
ELF, at \url{https://github.com/pytorch/ELF}, is a platform for RL research~\citep{Tian2017ELF}.
ELF OpenGo is a reimplementation of AlphaGo Zero/Alpha Zero using the ELF framework.

\clearpage


\newpage

\section*{Part I: Core Elements}

\addcontentsline{toc}{section}{Part I:  Core Elements}

A reinforcement learning (RL) agent observes states, executes actions, and  receives rewards,  
with major components of value function, policy and model. 
A RL problem may be formulated as a prediction, control, or planning problem, 
and solution methods may be model-free or model-based, 
and value-based or policy-based. 
Exploration vs. exploitation is a fundamental tradeoff in RL. 
Representation is relevant to all elements in RL  problems.

\begin{figure}[h]
\centering
\includegraphics[width=0.7\linewidth]{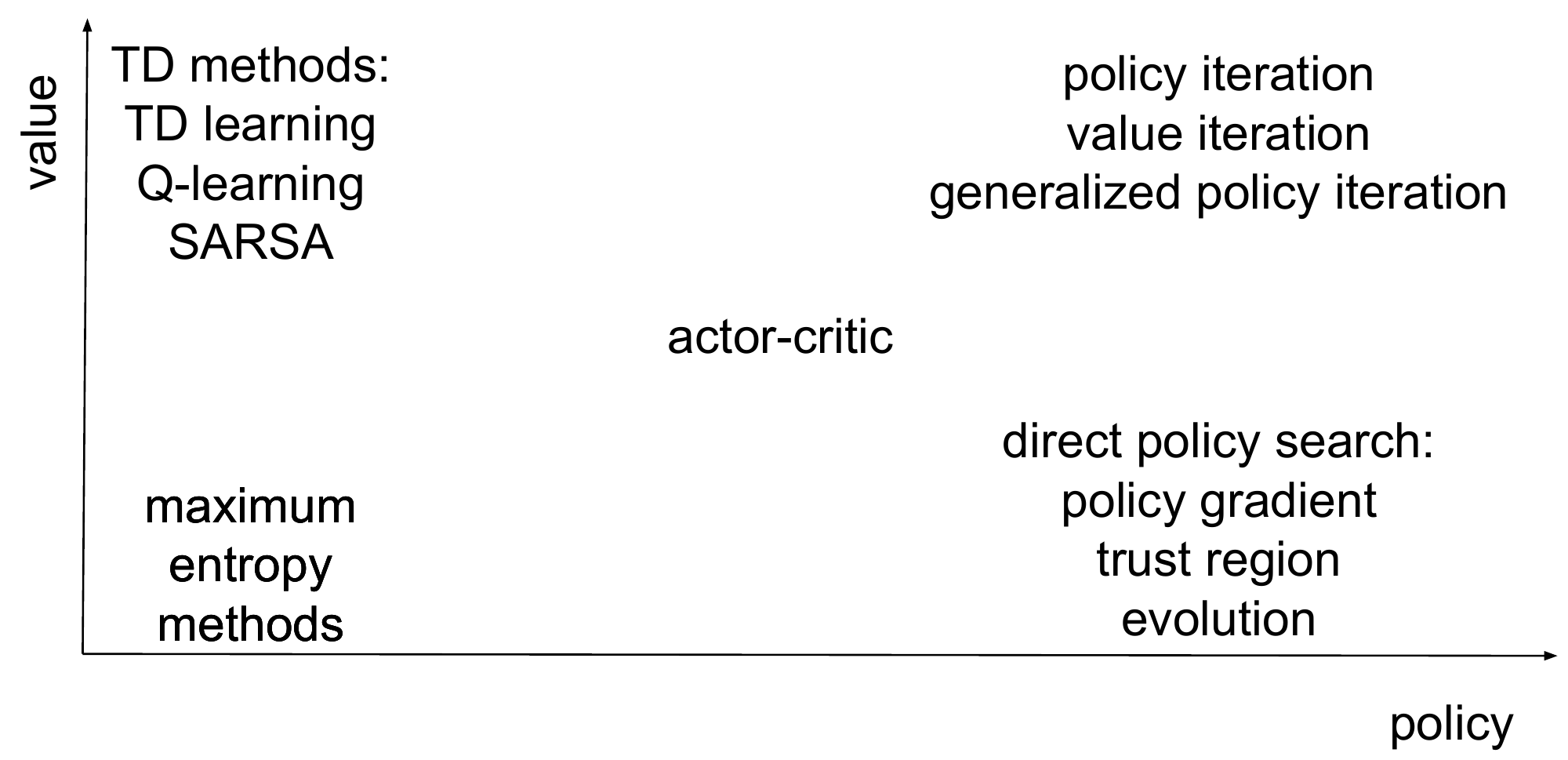}
\caption{Value- and Policy-based RL Methods}
\label{fig:policy}
\end{figure}

Figure~\ref{fig:policy} illustrates value- and policy-based methods.
TD methods, e.g., TD-learning, Q-learning, SARSA, and,
Deep Q-network (DQN)~\citep{Mnih-DQN-2015}, are purely value-based.
Direct policy search methods include
policy gradient, e.g., REINFORCE~\citep{Williams1992}, 
trust region methods, e.g.,
Trust Region Policy Optimization (TRPO)~\citep{Schulman2015}, and,
Proximal Policy Optimization (PPO) \citep{Schulman2017PPO},
and, 
evolution methods, e.g., Covariance Matrix Adaptation Evolution Strategy (CMA-ES)~\citep{Hansen2016}.
Actor-critic methods combine value function and policy.
Maximum entropy methods further bridge the gap between value- and policy-based methods,
e.g., 
soft Q-learning~\citep{Haarnoja2017},
Path Consistency Learning (PCL)~\citep{Nachum2017Gap},
and, 
trust-PCL~\citep{Nachum2018TrustPCL}.
Policy iteration, value iteration, and, generalized policy iteration, e.g.,
AlphaGo~\citep{Silver-AlphaGo-2016, Silver-AlphaGo-2017}
and DeepStack~\citep{Moravcik2017},
are based on (approximate) dynamic programming.

In this part, we discuss RL core elements: value function in Chapter~\ref{value}, policy in Chapter~\ref{policy}, reward in Chapter~\ref{reward}, model-based RL in Chapter~\ref{model},  exploration vs exploitation in Chapter~\ref{exploration}, and representation in Chapter~\ref{representation}.

\clearpage

\newpage

\section{Value Function}
\label{value}

Value function is a fundamental concept in reinforcement learning.
A value function is a prediction of the expected, accumulative, discounted, future reward, measuring the goodness of each state, or each state-action pair. 
Temporal difference (TD) learning~\citep{Sutton1988} and its extension, Q-learning~\citep{Watkins1992}, are classical algorithms for learning state and action value functions respectively. 
Once we have an optimal value function, we may derive an optimal policy.

In the following, we first introduce Deep Q-Network (DQN)~\citep{Mnih-DQN-2015}, a recent breakthrough, and its extensions.
DQN ignited this wave of deep reinforcement learning, combating the stability and convergence issues with experience replay and target networks, which make Q-learning closer to supervised learning.
Next we introduce value distribution, rather than value expectation as in classical TD and Q learning.
Then we discuss general value function, usually with the goal as a parameter of a value function, besides the state or the state action pair.
General value functions hold great promise for further development of RL and AI.

Recently, there are more and more work using policy-based methods to obtain an optimal policy directly. 
There are also work combing policy gradient with off-policy Q-learning, 
e.g., \citet{Gu2017QProp, ODonoghue2017,Gu2017Interpolated, Haarnoja2017, Haarnoja2018, Nachum2017Gap, Nachum2018TrustPCL, Dai2018SBEED}.
\citet{Dai2018SBEED} propose to solve the Bellman equation using primal-dual optimization; instead TD learning and Q-learning are based on fixed point iteration.
We discuss policy optimization in next Chapter.

\subsection{Deep Q-Learning}
\label{DQN}

\citet{Mnih-DQN-2015} introduce Deep Q-Network (DQN)  and ignite the field of deep RL.
There are early work to integrate neural networks with RL, e.g.~\citet{Tesauro1994} and~\citet{Riedmiller2005}.
Before DQN, it is well known that RL is unstable or even divergent when action value function is approximated with a nonlinear function like neural networks.  
That is, the deadly triad issue, when combining off-policy, function approximation, and, bootstrapping.
 DQN makes several contributions: 1)  stabilizing the training of action value function approximation with deep neural networks, in particular, CNNs, using experience replay~\citep{Lin1992} and target network; 2) designing an end-to-end RL approach, with only the pixels and the game score as inputs, so that only minimal domain knowledge is required; 3) training a flexible network with the same algorithm, network architecture and hyperparameters to perform well on many different tasks, i.e., 49 Atari games~\citep{Bellemare2013}, outperforming previous algorithms, and performing comparably to a human professional tester. 
Note, different games are trained separately, so the network weights are different. 
 
DQN uses a CNN to approximate the optimal action value function,
\begin{equation}
Q^*(s,a) = \max_\pi \mathbb{E}[\sum_0^\infty \gamma^i r_{t+1}| s_t = s, a_t = a, \pi]. 
\end{equation} 
In Section~\ref{functionapproximation}, we discuss deadly triad, i.e.,  instability and divergence may occur, when combining off-policy, function approximation, and bootstrapping. Several factors cause the instability: 1) correlations in sequential observations; 2) small updates to action value function $Q$ may change the policy dramatically, and consequently change the data distribution; 3) correlations between action values $Q$ and $r+ \gamma \max_{a'} Q(s', a')$, which is usually used as the target value in batch Q-learning.

DQN uses experience replay and target networks to address the instability issues. In experience replay, observation sequences $(s_t, a_t, r_t, s_{t+1})$ are stored in the replay buffer, and sampled randomly, to remove correlations in the data, and to smooth data distribution changes. In DQN, experiences are sampled uniformly, and as we discuss later, prioritized experience replay~\citep{Schaul2016} samples experiences according to their importance. A target network keeps its separate network parameters, and update them only periodically, to reduce the correlations between action values $Q$ and the target $r+ \gamma \max_{a'} Q(s', a')$.     
The following is the loss function Q-learning uses to update network parameters at iteration $i$, 
\begin{equation}
(r+ \gamma \max_{a'} Q(s', a'; \bm{\theta}_i^-) - Q(s, a, \bm{\theta}_i) )^2
\end{equation} 
where  $\bm{\theta}_i$ are parameters of the $Q$-network at iteration $i$, $\bm{\theta}_i^-$ are parameters of the target network at iteration $i$. The target network parameters $\bm{\theta}_i^-$ are updated periodically, and held fixed in between.
We present DQN pseudo code in Algorithm~\ref{DQN-algo}. 

\begin{algorithm}[h]
\SetAlgoNoLine

\hrulefill

\textbf{Input: } the pixels and the game score\\
\textbf{Output: } Q action value function (from which we obtain a policy and select actions)\\

\hrulefill

initialize replay memory $D$\\
initialize action-value function $Q$ with random weight $\bm{\theta}$\\
initialize target action-value function $\hat{Q}$ with weights $\bm{\theta}^{-} = \bm{\theta}$\\    
\For{episode = 1 to $M$}{
	initialize sequence $s_1 = \{x_1\}$ and preprocessed sequence $\phi_1 = \phi(s_1)$\\
	\For{t = 1 to $T$}{
		following $\epsilon$-greedy policy, select $a_t =
  		\begin{cases}
    		\text{a random action}      & \text{with probability } \epsilon\\
    		\argmax_{a}Q(\phi(s_t), a; \bm{\theta})   & \text{otherwise}\\
  		\end{cases} $\\
		execute action $a_i$ in emulator and observe reward $r_t$ and image $x_{t+1}$\\
		set $s_{t+1} = s_t, a_t, x_{t+1}$ and preprocess $\phi_{t+1} = \phi(s_{t+1})$\\
		store transition $(\phi_t, a_t, r_t, \phi_{t+1})$ in $D$\\
		\tcp{experience replay}
		sample random minibatch of transitions $(\phi_j, a_j, r_j, \phi_{j+1})$ from $D$\\
		set $y_j =
  		\begin{cases}
    		r_j       & \text{if episode terminates at step } j+1\\
    		r_j + \gamma \max_{a'} \hat{Q}(\phi_{j+1}, a'; \bm{\theta}^-)   & \text{otherwise}\\
  		\end{cases} $\\
		perform a gradient descent step on $(y_j - Q(\phi_j, a_j; \bm{\theta}))^2$ w.r.t. the network parameter $\bm{\theta}$\\
		\tcp{periodic update of target network}
		in every $C$ steps, reset $\hat{Q} = Q$, i.e., set $\bm{\theta}^- = \bm{\theta}$
	}
}

\hrulefill

\caption{Deep Q-Nework (DQN), adapted from \citet{Mnih-DQN-2015}}
\label{DQN-algo}
\end{algorithm}

DQN has a preprocessing step to reduce the input dimensionality. DQN also uses error clipping, clipping the update $r+ \gamma \max_{a'} Q(s', a'; \bm{\theta}_i^-) - Q(s, a, \bm{\theta}_i)$ in $[-1,1]$, to help improve stability. (This is not reflected in the pseudocode.) 
\citet{Mnih-DQN-2015} also present visualization results using t-SNE~\citep{vanderMaaten2008}.

DQN makes a breakthrough by showing that Q-learning with non-linear function approximation, in particular, deep convolutional neural networks, can achieve outstanding results over many Atari games. 
Following DQN, many work improve DQN in various aspects, e.g., in the following, we will discuss over-estimation in Q-learning, prioritized experience replay, and a dueling network to estimate state value function and associated advantage function, and then combine them to estimate action value function. 
We also discuss an integration method called Rainbow to combine several techniques together. 

In later chapters, we will discuss more extensions, like asynchronous advantage actor-critic (A3C)~\citep{Mnih-A3C-2016} in Section~\ref{policy:AC}, better exploration strategy to improve DQN~\citep{Osband2016} in Chapter~\ref{exploration},  and hierarchical DQN in Chapter~\ref{hierarchical}, etc. 
Experience replay uses more memory and computation for each interaction, and it requires off-policy RL algorithms. This motivates the asynchronous methods as we will discuss in Section~\ref{policy:AC}.

See more work as the following. 
\citet{Anschel2017} propose to reduce variability and instability by an average of previous Q-values estimates.
\citet{Farebrother2019} study generalization and regularization in DQN.
\citet{Guo2014} combine DQN with offline Monte-Carlo tree search (MCTS) planning.
\citet{Hausknecht2015} propose deep recurrent Q-network (DRQN) to add recurrency to DQN for the partial observability issue in Atari games.
\citet{He2017} propose  to accelerate DQN by optimality tightening, a constrained optimization approach,  to propagate reward faster, and to improve accuracy over DQN.
\citet{Kansky2017} propose schema networks and empirically study variants of Breakout in Atari games.
\citet{Liang2016} propose to replicate DQN with shallow features w.r.t. spatial invariance, non-Markovian features, and object detection, together with basic tile-coding features.
\citet{Zahavy2018NIPS} study action elimination.

See a blog at \url{https://deepmind.com/research/dqn/}. 
See Chapter 16 in \citet{Sutton2018} for a detailed and intuitive description of Deep Q-Network.   
See Chapter 11 in \citet{Sutton2018} for more details about the deadly triad. 

\subsection*{Double DQN}
\label{D-DQN}

\citet{vanHasselt2016} propose Double DQN (D-DQN) to tackle the over-estimate problem in Q-learning~\citep{vanHasselt2010}. 
In standard Q-learning, as well as in DQN, the parameters are updated as follows:  
\begin{equation}
\bm{\theta}_{t+1} = \bm{\theta}_t + \alpha (y_t ^Q- Q(s_t, a_t; \bm{\theta}_t) ) \nabla_{\bm{\theta}_t}  Q(s_t, a_t; \bm{\theta}_t),
\end{equation}
where 
\begin{equation}
y_t^Q = r_{t+1} + \gamma \max_{a}Q(s_{t+1}, a; \bm{\theta}_t),
\end{equation}
so that the max operator uses the same values to both select and evaluate an action. As a consequence, it is more likely to select over-estimated values, and results in over-optimistic value estimates. \citet{vanHasselt2016}  propose to evaluate the greedy policy according to the online network, but to use the target network to estimate its value. This can be achieved with a minor change to the DQN algorithm, replacing $y_t^Q$ with 
\begin{equation}
y_t^{D-DQN} = r_{t+1} + \gamma Q(s_{t+1}, \argmax_{a} Q(s_{t+1}, a; \bm{\theta}_t); \bm{\theta}_t^-),
\end{equation}
where $\bm{\theta}_t$ is the parameter for online network and $\bm{\theta}_t^-$ is the parameter for target network. For reference, $y_t^Q$ can be written as 
\begin{equation}
y_t^{Q} = r_{t+1} + \gamma Q(s_{t+1}, \argmax_{a} Q(s_{t+1}, a; \bm{\theta}_t); \bm{\theta}_t).
\end{equation}

\subsection*{Prioritized Experience Replay} 

In DQN, experience transitions are uniformly sampled from the replay memory, regardless of the significance of experience. \citet{Schaul2016} propose to prioritize experience replay, so that important experience transitions can be replayed more frequently, to learn more efficiently. The importance of experience transitions are measured by TD errors. The authors design a stochastic prioritization based on the TD errors, using importance sampling to avoid the bias in the update distribution. The authors use prioritized experience replay in DQN and D-DQN, and improve their performance on Atari games. In planning, prioritized sweeping~\citep{Moore93} is used to set priorities for state action pairs according to TD errors to achieve more efficient updates.

\subsection*{Dueling Architecture} 
\label{dueling}

\citet{Wang-Dueling-2016} propose the dueling network architecture to estimate state value function $V(s)$ and the associated advantage function $A(s,a)$, and then combine them to estimate action value function $Q(s,a)$, to converge faster than Q-learning. 
In DQN, a CNN layer is followed by a fully connected (FC) layer. 
In dueling architecture, a CNN layer is followed by two streams of FC layers, to estimate value function and advantage function separately; then the two streams are combined to estimate action value function. Usually we use the following to combine $V(s)$ and $A(s,a)$ to obtain $Q(s,a)$,
\begin{equation}
Q(s,a; \bm{\theta}, \alpha, \beta) = V(s; \bm{\theta}, \beta) + \big( A(s,a; \bm{\theta}, \alpha) - \max_{a'}A(s,a'; \bm{\theta}, \alpha)\big) 
\end{equation}
where $\alpha$ and $\beta$ are parameters of the two streams of FC layers. \citet{Wang-Dueling-2016} propose to replace max operator with average as below for better stability,
\begin{equation}
Q(s,a; \bm{\theta}, \alpha, \beta) = V(s; \bm{\theta}, \beta) + \big( A(s,a; \bm{\theta}, \alpha) - \frac{a}{|\mathcal{A}|}A(s,a'; \bm{\theta}, \alpha)\big) 
\end{equation}
Dueling architecture implemented with D-DQN and prioritized experience replay improves previous work, DQN and D-DQN with prioritized experience replay, on Atari games.

\subsection*{Rainbow} 
\label{rainbow}

\citet{Hessel2018} propose Rainbow to combine DQN, double Q-learning, prioritized replay, dueling networks, multi-step learning, value distribution (discussed in Section~\ref{distributional} below), and noisy nets (\citet{Fortunato2018}, an exploration technique by adding parametric noises to network weights), and achieve better data efficiency and performance on Atari games. 
The ablation study show that removing either prioritization or multi-step learning worsens performance for most games; 
however, the contribution of each component vary significantly for different games.

\subsection*{Retrace}

\citet{Remi2016} propose Retrace($\lambda$) for a safe and efficient return-based off-policy control RL algorithm,
for low variance, safe use of samples from any behaviour policy, and efficiency with using samples from close behaviour policies.
The authors analyze the property of convergence to the optimal action value $Q^*$,
without the assumption of Greedy in the Limit with Infinite Exploration (GLIE)~\citep{Singh2000},
and the convergence of Watkins' $Q(\lambda)$. 
The authors experiment with Atari games.
\citet{Gruslys2017} extend Retrace~\citep{Remi2016} for actor-critic.

\subsection{Distributional Value Function}
\label{distributional}


Usually, value-based RL methods use expected values, like TD learning and Q-learning. 
However, expected values usually do not characterize a distribution, and more information about value distribution may be beneficial.
Consider a commuter example. For 4 out of 5 days, she takes 15 minutes to commute to work, and for the rest 1 out of 5 days, she takes 30 minutes. On average, she takes 18 minutes to work. However, the commute takes either 15 or 30 minutes, but never 18 minutes. 

\citet{Bellemare2017Distributional} propose a value distribution approach to RL problems. 
A value distribution is the distribution of the random return received by a RL agent.
In contrast to, and analogous with, the Bellman equation for action value function,
\begin{equation}
Q(s,a) = \mathbb{E} R(s,a) + \gamma \mathbb{E} Q(S', A')
\end{equation}
\citet{Bellemare2017Distributional} establish the distributional Bellman equation,
\begin{equation}
Z(s,a) = R(s,a) + \gamma Z(S', A')
\end{equation}
Here, $Z$ is the random return, and its expectation is the value $Q$.
Three random variables characterize the distribution of $Z$: the reward $R$, the next state action pair $(S', A')$, and the random return $Z(S', A')$.

\citet{Bellemare2017Distributional}  prove the contraction of the policy evaluation Bellman operator for the value distribution, so that, for a fixed policy, the Bellman operator contracts in a maximal form of the Wasserstein metric. 
The Bellman operator for the value distribution, $\mathcal{T}^\pi: \mathcal{Z} \rightarrow \mathcal{Z}$, was defined as,
\begin{equation}
\mathcal{T}^\pi Z(s,a) = R(s,a) + \gamma P^\pi Z(s,a) 
\end{equation}
where $P^\pi Z(s,a) = Z(S', A'), S' \sim P(\cdot| s,a), A' \sim \pi(\cdot|S')$.
The authors also show instability in the control setting, i.e., in the Bellman optimal equation for the value distribution. 
The authors argue that value distribution approximations have advantages over expectation approximations, for preserving multimodality in value distributions, and mitigating the effects of a nonstationary policy on learning, and demonstrate the importance of value distribution for RL both theoretically and empirically with Atari games.  

\citet{Bellemare2017Distributional} propose a categorical algorithm for approximate distributional reinforcement learning.
The value distribution is modelled using a discrete distribution, with a set of atoms as the support, $\{z_i = V_{MIN} + i \triangle z : 0 \leq i < N\}$, $\triangle z = \frac{V_{MAX} - V_{MIN} }{N-1}$, where $N \in \mathbb{N}, V_{MIN} \in \mathbb{R}, V_{MAX} \in \mathbb{R}$ are parameters. 
The atom probabilities are then given by
\begin{equation}
Z_\theta(s,a)= z_i \mbox{, with probability, }  p_i(s,a) = \frac{e^{\theta_i(s,a)}}{\sum_j e^{\theta_j(s,a)}}.
\end{equation}
To deal with the issue of disjoint supports caused by Bellman update $\mathcal{T} Z_\theta$, and learning with sample transitions, \citet{Bellemare2017Distributional} project the sample update  $\hat{\mathcal{T}} Z_\theta$ onto the support of $Z_\theta$, by computing the Bellman update $\hat{\mathcal{T}} z_j = r + \gamma z_j$ for each atom $z_j$, then distributing its probability to the immediate neighbours of the update $\hat{\mathcal{T}} z_j$. Use $\Phi \hat{\mathcal{T}} Z_\theta(s,a)$ to denote the operations of a sample Bellman operator following the projection of distributing probabilities. The sample loss function is the cross-entropy term of the KL divergence, and can be optimized with gradient descent,
\begin{equation}
\mathcal{L}_{s,a}(\theta) = D_{KL}(\Phi \hat{\mathcal{T}} Z_{\theta}(s,a) || Z_\theta(s,a)).
\end{equation}

Algorithm~\ref{C51-algo} presents pseudo-code for the categorial algorithm, which computes the sampling Bellman operator followed by a projection of distributing probabilities for one transition sample.

\begin{algorithm}[h]
\SetAlgoNoLine

\hrulefill

\textbf{Input: } a transition $s_t, a_t, r_t, s_{t+1}, \gamma_t$\\
\textbf{Output: } cross-entropy loss

\hrulefill

// $\{z_i\}$ are atom support, $p_i(s, a)$ are atom probabilities, $i\in\{0, \dots, N-1\}$\\
$Q(s_{t+1}, a) = \sum_i z_i p_i(s_{t+1}, a)$\\
$a^* \leftarrow \argmax_a Q(s_{t+1}, a)$\\
$m_i = 0, i \in \{0, \dots, N-1\}$\\ 
\For{$j \in \{0, \dots, N-1\}$}{
	//compute the projection of $\hat{\mathcal{T}} z_j$ onto the support of $\{z_i\}$\\
	// $[\cdot]_a^b$ bounds the argument in the range of $[a,b]$\\
        	$\hat{\mathcal{T}} z_j \leftarrow [r+t + \gamma_t z_j]_{V_{MIN}}^{V_{MAX}}$\\
	// $b_j \in [0, N-1]$\\
	$b_j \leftarrow (\hat{\mathcal{T}} - V_{MIN})/\triangle z $ \\
	$l \leftarrow \lfloor b_j \rfloor, u \leftarrow \lceil b_j \rceil$\\
	//distribute probability of $\hat{\mathcal{T}} z_j$\\
	$m_l \leftarrow m_l + p_j(x_{t+1}, a^*) (u - b_j)$\\
	$m_u \leftarrow m_u + p_j(x_{t+1}, a^*) ( b_j -l)$
}
output $-\sum_i m_i \log p_i(x_t, a_t)$ \\
\hrulefill
\caption{Categorical Algorithm, adapted from \citet{Bellemare2017Distributional}}
\label{C51-algo}
\end{algorithm}

\citet{Morimura2010ICML, Morimura2010UAI} propose risk sensitive algorithms.
\citet{Rowland18} analyze the categorical distributional reinforcement learning proposed in~\citet{Bellemare2017Distributional}.
\citet{Doan2018} propose GAN Q-learning for distributional RL.
\citet{Dabney2018Quantile} propose to utilize quantile regression for state-action return distribution.

See a talk by Marc Bellemare at \url{https://vimeo.com/235922311}.
See a blog at \url{https://deepmind.com/blog/going-beyond-average-reinforcement-learning/},
with a video about Atari games experiments.

\subsection{General Value Function}
\label{value:GVF}

\subsection*{Horde}

\citet{Sutton2011}  discuss that value functions provide semantics for predictive knowledge and goal-oriented (control) knowledge,
and propose to represent knowledge with general value function, where policy, termination function, reward function, and terminal reward function are parameters. 
The authors then propose Horde, a scalable real-time architecture for learning  in parallel general value functions for independent sub-agents from unsupervised sensorimotor interaction, i.e., nonreward signals and observations. Horde can learn to predict the values of many sensors, and  policies to maximize those sensor values, with general value functions, and answer predictive or goal-oriented questions. 
Horde is off-policy, i.e., it learns in real-time while following some other behaviour policy, and learns with gradient-based temporal difference learning methods, with constant time and memory complexity per time step.

\subsection*{Universal Value Function Approximators}

\citet{Schaul2015} propose Universal Value Function Approximators (UVFAs) $V(s,g; \theta)$ to generalize over both states $s$ and goals $g$, to extend normal state value function approximators. 
The aim of UVFAs is to exploit structure across both states and goals, by taking advantages of similarities encoded in the goal representation, and the structure in the induced value function.  
\citet{Schaul2015} show that such UVFAs can be trained using direct bootstrapping with a variant of Q-learning, and the derived greedy policy can generalize to previously unseen action-goal pairs. 
UVFAs can be regarded as an infinite Horde of demons, without scalability issue as in Horde.

\subsection*{Hindsight Experience Replay}

\citet{Andrychowicz2017} propose Hindsight Experience Replay (HER) to combat with the sparse reward issue, inspired by UVFAs~\citep{Schaul2015}.
The idea is, after experiencing some episodes, storing every transition in the replay buffer, with not only the original goal for this episode, but also some other goals. 
HER can replay each trajectory with any goal with an off-policy RL algorithm, since the original goal pursued does not influence the environment dynamics, albeit it influences the actions.
\citet{Andrychowicz2017} combine HER with DQN and DDPG (as will be discussed in Section~\ref{policy:PG}) on several robotics arm tasks, push, slide and pick-and-place, and perform well.

An OpenAI blog describes HER, with videos about HER experiments, introduces a baseline HER implementation and simulated robot environment, and proposes potential research topics with HER, 
\url{https://blog.openai.com/ingredients-for-robotics-research/}.

\clearpage


\section{Policy}
\label{policy}


A policy maps a state to an action, or, a distribution over actions,
and policy optimization is to find an optimal mapping. 
Value-based methods optimize value functions first, then derive optimal policies.
Policy-based methods directly optimize an objective function, usually cumulative rewards.

REINFORCE~\citep{Williams1992} is a classical policy gradient method, 
with a Monte Carlo approach to estimate policy gradients. 
Policy gradient methods can stay stable when combined with function approximation, under some conditions.
However, sample inefficiency is a major issue;
and Monte Carlo sampling with rollouts usually results in high variance in policy gradient estimates.

Incorporating a baseline or critic can help reduce variance.  
In actor-critic algorithms~\citep{Barto1983, Vijay2003}, the critic updates action value function parameters, 
the actor updates policy parameters, in the direction suggested by the critic,
and the value function can help reduce the variance of policy parameter estimates,
by replacing rollout estimates, with the possibility of encountering a higher bias.
We discuss policy gradient, actor critic, and REINFORCE in Section~\ref{bg:policy}.

On-policy methods, like TD learning, are usually sample inefficient by using data only once, 
with estimations based on trajectories from the current policy.
Off-policy methods, like Q-learning, can learn from any trajectories from any policies, 
e.g., expert demonstrations, from the same environment.  
Recently, experience replay regains popularity after DQN, as we discuss in Chapter~\ref{value}.
This usually makes off-policy methods more sample efficient than on-policy methods.
Importance sampling is a variance reduction technique in Monte Carlo methods.
\citet{Precup2001} study off-policy TD learning with importance sampling.
\citet{Degris2012} propose off-policy actor-critic with importance sampling.
\citet{LiuQiang2018NIPS} propose an off-policy estimation method with importance sampling to avoid the exploding variance issue.

\citet{Kakade2002} introduce natural policy gradient to improve stability and convergence speed of policy based methods.
This leads to trust region methods, like Trust Region Policy Optimization (TRPO)~\citep{Schulman2015},
and Proximal Policy Optimization (PPO)~\citep{Schulman2017PPO}, two on-policy methods.
Trust-PCL~\citep{Nachum2018TrustPCL} is an off-policy trust region method.

It is desirable to improve data efficiency of policy gradient, while keeping its stability and unbiasedness.
Asynchronous advantage actor-critic (A3C)~\citep{Mnih-A3C-2016} integrates policy gradient with on-line critic.
There are recent work for policy gradient with off-policy critic, like    
deep deterministic policy gradient (DDPG)~\citep{Lillicrap2016},
and policy gradient with Q-learning (PGQL)~\citep{ODonoghue2017}.
Interpolated policy gradient~\citep{Gu2017Interpolated},
and Q-Prop~\citep{Gu2017QProp} are proposed to 
combine stability of trust region methods with off-policy sample efficiency.   
However, the deadly triad issue results from the combination off-policy, function approximation, and bootstrapping, 
so that instability and divergence may occur~\citep{Sutton2018}.

There are efforts to establish theoretical guarantees, tackling the deadly triad, e.g.,   
Retrace~\citep{Remi2016},
path consistency learning (PCL)~\citep{Nachum2017Gap},
Trust-PCL~\citep{Nachum2018TrustPCL}, and 
SBEED~\citep{Dai2018SBEED}, etc.

Maximum entropy methods integrate policy gradient with off-policy learning
and attempt to bridge the gap between value- and policy-based methods,
e.g., \citet{Ziebart2008}, \citet{Ziebart2010}, G-learning~\citep{Fox2016}, soft Q-learning~\citep{Haarnoja2017}, 
and several mentioned above,  including PGQL, PCL, and Trust-PCL.

In conventional RL, we are content with deterministic policies~\citep{Sutton2018}.
However, stochastic policies are desirable sometimes, for reasons like,
partial observable environments,
handling uncertainty~\citep{Ziebart2008, Ziebart2010},
convergence and computation efficiency~\citep{Gu2017QProp},
exploration and compositionality~\citep{Haarnoja2017},
and optimal solutions for some games like rock-paper-scissors, etc. 
Policy gradient methods usually obtain stochastic policies.
So are maximum entropy regularized methods.

Here we focus on model-free policy optimization algorithms. 
We will discuss model-based ones, like stochastic value gradients (SVG)~\citep{Heess2015}, 
and normalized advantage functions (NAF)~\citep{Gu2016}, 
in Chapter~\ref{model}.
we discuss guided policy search (GPS)~\citep{Levine2016} in Chapter~\ref{robotics}.

Evolution strategies achieve excellent results, e.g.,~\citet{Petroski2017}, \citet{Salimans2017}, \citet{Lehman2017}. 
See \citet{Hansen2016} for a tutorial. 
\citet{Khadka2018NIPS} propose evolutionary reinforcement learning.

Policy search methods span a wide spectrum from direct policy search to value-based RL, includes:
evolutionary strategies, 
covariance matrix adaptation evolution strategy (CMA-ES)~\citep{Hansen2001, Hansen2016}, 
episodic relative entropy policy search (REPS)~\citep{Peters2010},
policy gradients, 
probabilistic inference for learning control (PILCO)~\citep{Deisenroth2011}, 
model-based REPS~\citep{Abdolmaleki2015}, 
policy search by trajectory optimization~\citep{Levine2014},
actor critic, natural actor critic~\citep{Kakade2002},
episodic natural actor critic (eNAC),
advantage weighted regression~\citep{Peters2007},
conservative policy iteration~\citep{Kakade2002ICML},
least square policy iteration (LSPI)~\citep{Lagoudakis03},
Q-learning, and fitted Q-learning~\citep{Riedmiller2005}.
See \citet{Peters2015} for more details.
AlphaGo~\citep{Silver-AlphaGo-2016, Silver-AlphaGo-2017}, as well as \citet{SunWen2018NIPS}, 
follows the scheme of generalized policy iteration.
We will discuss AlphaGo in Section~\ref{games:AlphaGo}.

See \citet{Abbeel2017} for a tutorial on policy optimization.
\citet{Levine2018inference} discusses connections between RL and control, in particular, maximum entropy RL, and probabilistic inference.
See NIPS 2018 Workshop on Infer to Control: Probabilistic Reinforcement Learning and Structured Control, 
at \url{https://sites.google.com/view/infer2control-nips2018}.


In the following,  we discuss 
policy gradient
in Section~\ref{policy:PG},
actor-critic
in Section~\ref{policy:AC},
trust region methods
in Section~\ref{policy:TR},
policy gradient with off-policy learning
in Section~\ref{policy:PGQ}, and,
benchmark results 
in Section~\ref{policy:benchmark}.


\subsection{Policy Gradient}
\label{policy:PG}

Policy gradients are popular methods in RL, optimizing policies directly. 
Policies may be deterministic or stochastic. 
\citet{Silver-DPG-2014} propose Deterministic Policy Gradient (DPG) and 
\citet{Lillicrap2016}  extend it to deep DPG (DDPG) for efficient estimation of policy gradients. 
\citet{Houthooft2018NIPS} propose evolved policy gradients with meta-learning.

As discussed in \citet{Heess2015}, most policy gradient methods, like REINFORCE, use likelihood ratio method as discussed in Section~\ref{bg:policy}, by sampling returns from interactions with the environment in a model-free manner; another approach, value gradient method, is to estimate the gradient via backpropagation, and DPG and DDPG follow this approach. 

\citet{Silver-DPG-2014} introduce the Deterministic Policy Gradient (DPG) algorithm for RL problems with continuous action spaces. 
The deterministic policy gradient is the expected gradient of the action-value function, which integrates over the state space; whereas in the stochastic case, the policy gradient integrates over both state and action spaces. Consequently, the deterministic policy gradient can be estimated more efficiently than the stochastic policy gradient. 
The authors introduce an off-policy actor-critic algorithm to learn a deterministic target policy from an exploratory behaviour policy, and to ensure unbiased policy gradient with the compatible function approximation for deterministic policy gradients. Empirical results show its superior to stochastic policy gradients, in particular in high dimensional tasks, on several problems: a high-dimensional bandit; standard benchmark RL tasks of mountain car, pendulum, and 2D puddle world with low dimensional action spaces; and controlling an octopus arm with a high-dimensional action space. 
The experiments are conducted with tile-coding and linear function approximators.

\citet{Lillicrap2016} propose an actor-critic, model-free, Deep Deterministic Policy Gradient (DDPG) algorithm in continuous action spaces, by extending DQN~\citep{Mnih-DQN-2015} and DPG~\citep{Silver-DPG-2014}. With actor-critic as in DPG, DDPG avoids the optimization of action value function at every time step to obtain a greedy policy as in Q-learning, which will make it infeasible in complex action spaces with large, unconstrained function approximators like deep neural networks.  
To make the learning stable and robust, similar to DQN, DDPQ  deploys experience replay and an idea similar to target network, a "soft" target, which, rather than copying the weights directly as in DQN, updates the soft target network weights $\bm{\theta}'$ slowly to track the learned networks weights $\bm{\theta}$: $\bm{\theta}' \leftarrow \tau \bm{\theta} + (1-\tau) \bm{\theta}'$, with $\tau \ll 1$. 
The authors adapt batch normalization to handle the issue that the different components of the observation with different physical units.
As an off-policy algorithm, DDPG learns an actor policy with experiences from an exploration policy by adding noises sampled from a noise process to the actor policy. 
More than 20 simulated physics tasks of varying difficulty in the MuJoCo environment are solved with the same learning algorithm, network architecture and hyper-parameters, and obtain policies with performance competitive with those found by a planning algorithm with full access to the underlying physical model and its derivatives. DDPG can solve problems with 20 times fewer steps of experience than DQN, although it still needs a large number of training episodes to find solutions, as in most model-free RL methods. It is end-to-end, with raw pixels as input. 

\citet{Hausknecht2016} extend DDPG by considering parameterization of action spaces, 
and experiment with the domain of simulated RoboCup soccer.



\subsection{Actor-Critic}
\label{policy:AC}

An actor critic algorithm learns both a policy and a state value function, and the value function is used for bootstrapping, i.e.,  updating a state from subsequent estimates, to reduce variance and accelerate learning~\citep{Sutton2018}. 

In the following, we focus on asynchronous advantage actor-critic (A3C)~\citep{Mnih-A3C-2016}. 
\citet{Mnih-A3C-2016} also discuss asynchronous one-step SARSA, one-step Q-learning and n-step Q-learning. 
A3C achieves the best performance among these asynchronous methods, and it can work for both discrete and continuous cases.

 \begin{algorithm}[h]
\SetAlgoNoLine

\hrulefill

global shared parameter vectors $\bm{\theta}$ and $\bm{\theta}_v$, thread-specific parameter vectors $\bm{\theta}'$ and $\bm{\theta}'_v$\\
global shared counter $T=0$, $T_{max}$\\
initialize step counter $t \leftarrow 1$\\
\For{$T \leq T_{max}$}{
	reset gradients, $d \bm{\theta} \leftarrow 0$ and $d \bm{\theta}_v \leftarrow 0$\\
	synchronize thread-specific parameters $\bm{\theta}' = \bm{\theta}$ and $\bm{\theta}'_v = \bm{\theta}_v$\\
	set $t_{start} = t$, get state $s_t$\\
	\For{$s_t$ $\text{ not terminal and }$ $t-t_{start} \leq t_{max}$}{
		take $a_t$ according to policy $\pi(a_t|s_t; \bm{\theta}')$\\
		receive reward $r_t$ and new state $s_{t+1}$\\
		$t \leftarrow t + 1$, $T \leftarrow T + 1$
		}
	$R =
  	\begin{cases}
    	0      & \text{for terminal } s_t\\
    	V(s_t, \bm{\theta}'_v)   & \text{otherwise}\\
  	\end{cases} $\\
	\For{$i \in \{t-1, ..., t_{start}\}$}{
		$R \leftarrow r_i + \gamma R$\\
		accumulate gradients wrt $\bm{\theta}'$: $d\bm{\theta} \leftarrow d\bm{\theta} + \nabla_{\bm{\theta}'} \log  \pi(a_i|s_i; \bm{\theta}') (R - V(s_i; \bm{\theta}'_v))$\\
		accumulate gradients wrt $\bm{\theta}'_v$: $d\bm{\theta}_v \leftarrow d\bm{\theta}_v + \nabla_{\bm{\theta}'_v} (R - V(s_i; \bm{\theta}'_v))^2$
		}
	update asynchronously $\bm{\theta}$ using $d\bm{\theta}$, and $\bm{\theta}_v$ using $d\bm{\theta}_v$
}

\hrulefill

\caption{A3C, each actor-learner thread, based on \citet{Mnih-A3C-2016}}
\label{A3C-algo}
\end{algorithm}

We present pseudo code for A3C for each actor-learner thread in Algorithm~\ref{A3C-algo}. 
A3C maintains a policy $\pi(a_t|s_t;\bm{\theta})$ and an estimate of the value function $V(s_t; \bm{\theta}_v)$, being updated with $n$-step returns in the forward view, after every $t_{max}$ actions or reaching a terminal state, similar to using minibatches. 
In $n$-step update, $V(s)$ is updated toward the $n$-step return, defined as, 
\begin{equation}
r_t + \gamma r_{t+1} + \cdots + \gamma^{n-1} r_{t+n-1} + \gamma^n V(s_{t+n}).
\end{equation}
Lines 15-19 show, each $n$-step update results in a one-step update for the last state, a two-step update for the second last state, and so on for a total of up to $t_{max}$ updates, for both policy and value function parameters.
The gradient update can be seen as 
\begin{equation}
\nabla_{\bm{\theta}'} \log \pi(a_t|s_t; \bm{\theta}')A(s_t, a_t; \bm{\theta}, \bm{\theta}_v),
\end{equation}
where 
\begin{equation}
A(s_t, a_t; \bm{\theta}, \bm{\theta}_v) = \sum_{i=0}^{k-1} \gamma^ir_{t+i} + \gamma^k V(s_{t+k}; \bm{\theta}_v) - V(s_t; \bm{\theta}_v)
\end{equation}
is an estimate of the advantage function, with $k$ upbound by $t_{max}$.

In A3C, parallel actors employ different exploration policies to stabilize training, so that experience replay is not utilized, although experience replay could improve data efficiency. Experience replay uses more memory and computation for each interaction, and it requires off-policy RL algorithms. Asynchronous methods can use on-policy RL methods. Moreover, different from most deep learning algorithms, asynchronous methods can run on a single multi-core CPU. 

For Atari games, A3C runs much faster yet performs better than or comparably with DQN, Gorila~\citep{Nair2015}, D-DQN, Dueling D-DQN, and Prioritized D-DQN. A3C also succeeds on continuous motor control problems: TORCS car racing games and MujoCo physics manipulation and locomotion, and Labyrinth, a navigating task in random 3D mazes using visual inputs. 

\citet{Wang2017} propose ACER, a stable and sample efficient actor-critic deep RL model using experience replay, with truncated importance sampling, stochastic dueling network (\citet{Wang-Dueling-2016} as discussed in Section~\ref{dueling}), and trust region policy optimization (\citet{Schulman2015} as will be discussed in Section~\ref{policy:TR}). \citet{Babaeizadeh2017} propose a hybrid CPU/GPU implementation of A3C.
\citet{Gruslys2017} propose Reactor to extend Retrace~\citep{Remi2016} for the actor-critic scheme. 
\citet{Horgan2018} propose Apex, a distributed version of actor-critic, with prioritized experience replay, 
and improve the performance on Atari games substantially. 
One important factor is that Apex can learn on a large amount of data.
\citet{Espeholt2018} propose IMPALA, a distributed actor-critic agent, and show good performance in multi-task settings.
\citet{Dai2018DualAC} propose dual actor-critic, in which the critic is not learned by standard algorithms like TD but is optimized to help compute gradient of the actor.


\subsection{Trust Region Methods}
\label{policy:TR}


Trust region methods are an approach to stabilize policy optimization by constraining gradient updates.
In the following, we discuss Trust Region Policy Optimization (TRPO)~\citep{Schulman2015}, 
and Proximal Policy Optimization (PPO) \citep{Schulman2017PPO}.
\citet{Nachum2018TrustPCL} propose Trust-PCL, and extension of TRPO for off-policy learning,
which we will discuss in Section~\ref{policy:PGQ}.
\citet{Heess2017DPPO} propose distributed proximal policy optimization.
\citet{Wu2017TRPO} propose scalable TRPO with Kronecker-factored approximation to the curvature.
\citet{Liu2018PGTD} study proximal gradient TD learning.
See a video about TRPO at, \url{https://sites.google.com/site/trpopaper/}.
See a blog about PPO with videos at, \url{https://blog.openai.com/openai-baselines-ppo/}.

\subsection*{Trust Region Policy Optimization (TRPO)}

\citet{Schulman2015} introduce an iterative procedure to monotonically improve policies theoretically, guaranteed by optimizing a surrogate objective function, and then make several approximations to develop a practical algorithm, Trust Region Policy Optimization (TRPO). In brief summary, TRPO iterates the following steps:
\begin{enumerate}
\item collect state action pairs and Monte Carlo estimates of Q values
\item average samples, construct the estimated objective and constraint in the previous optimization problem, 
where, $\theta_{old}$ denotes previous policy parameters, 
$q$ denotes the sampling distribution,
and $\delta$ is the trust region parameter
\begin{equation}
\max_{\theta}  \hat{\mathbb{E}} \Big[\frac{\pi_{\theta}(a|s)}{q(a|s)} Q_{\theta_{old}}(s,a) \Big]  
 \mbox{ subject to }  \hat{\mathbb{E}}[KL(\pi_{\theta_{old}}(\cdot|s) \| \pi_{\theta}(\cdot|s))] \leq \delta
\end{equation}
\item solve the above constrained optimization problem approximately, update the policy parameter $\theta$
\end{enumerate}

\citet{Schulman2015}  unify policy iteration and policy gradient with analysis, and show that policy iteration, policy gradient, and natural policy gradient~\citep{Kakade2002} are special cases of TRPO. 
In the experiments, TRPO methods perform well  on simulated robotic tasks of swimming, hopping, and walking, as well as playing Atari games in an end-to-end manner directly from raw images.

\subsection*{Proximal Policy Optimization (PPO)}

\citet{Schulman2017PPO} propose Proximal Policy Optimization (PPO), 
to alternate between data sampling and optimization, and to benefit the stability and reliability from TRPO, 
with the goal of simpler implementation, better generalization, and better empirical sample complexity.  
In PPO, parameters for policy and value function can be shared in a neural network, 
and advantage function can be estimated to reduce variance of policy parameters estimation.
PPO utilizes a truncated version of Generalized Advantage Estimator (GAE)~\citep{Schulman2016GAE}, reducing to multi-step TD update when $\lambda = 1$,
$\hat{A}_t = \delta_t + (\gamma \lambda) \delta_{t+1} + \cdots + (\gamma \lambda)^{T-t+1} \delta_{T-1}, \mbox{where } \delta_t = r_t + \gamma V(s_{t+1}) - V(s_t)$.
PPO achieves good performance on several continuous tasks in MuJoCo, on continuous 3D humanoid running and steering, and on discrete Atari games. As mentioned in an OpenAI blog about PPO, \url{https://blog.openai.com/openai-baselines-ppo/}, "PPO has become the default reinforcement learning algorithm at OpenAI because of its ease of use and good performance".


\subsection{Policy Gradient with Off-Policy Learning}
\label{policy:PGQ}

It is desirable to combine stability and unbiasedness of policy gradient, 
and sample efficiency of off-policy learning.
\citet{Levine2018inference} connects RL and control with probabilistic inference, 
and discusses that maximum entropy RL is equivalent to exact and variational probabilistic inference in deterministic and stochastic dynamics respectively.
We discuss several recent works following the approach of maximum entropy RL,
including \citet{Haarnoja2017}, \citet{Nachum2017Gap}, \citet{Nachum2018TrustPCL}, \citet{Haarnoja2018}, \citet{Gu2017QProp}, etc.
Maximum entropy RL can help exploration, compositionality, and partial observability~\citep{Levine2018inference}.

\subsection*{Soft Q-Learning}

\citet{Haarnoja2017} design a soft Q-learning algorithm, 
by applying a method for learning energy-based policies to optimize maximum entropy policies. 
In soft Q-learning, an optimal policy is expressed with a Boltzmann distribution,
and a variational method is employed to learn a sampling network to approximate samples from this distribution.
Soft Q-learning can improve exploration, 
and help stochastic energy-based policies achieve compositionality for better transferability. 

\citet{Haarnoja2018} propose soft actor-critic, based on the maximum energy RL framework in~\citep{Haarnoja2017},
so that the actor aims to maximize both expected reward and entropy.
\citet{Schulman2017} show equivalence between entropy-regularized Q-learning and policy gradient. 
\citet{Asadi2017} propose a new Q-value operator.

\subsection*{Path Consistency Learning (PCL)}

\citet{Nachum2017Gap} introduce the notion of softmax temporal consistency, to generalize  the hard-max Bellman consistency as in off-policy Q-learning, and in contrast to the average consistency as in on-policy SARSA and actor-critic. The authors establish the correspondence and a mutual compatibility property between softmax consistent action values and the optimal policy maximizing entropy regularized expected discounted reward. The authors propose Path Consistency Learning (PCL), attempting to bridge the gap between value and policy based RL, by exploiting multi-step path-wise consistency on traces from both on and off policies. 
The authors experiment with several algorithmic tasks.
Soft Q-learning~\citep{Haarnoja2017} is a one-step special case of PCL.


\citet{Nachum2018TrustPCL} propose Trust-PCL, an off-policy trust region method, 
to address sample inefficiency issue with the on-policy nature of trust region methods like TRPO and PPO.
The authors observe that an objective function maximizing rewards, regularized by relative entropy, 
led to an optimal policy and state value function which satisfy several multi-step pathwise consistencies along any path.
Therefore, Trust-PCL achieves stability and off-policy sample efficiency by employing relative entropy regularization.
The authors design a method to determine the coefficient for the relative entropy regularization term,
to simplify the task of hyperparameter tuning.
The authors experiment on standard continuous control tasks.

\citet{Dai2018SBEED} reformulate the Bellman equation into a primal-dual optimization problem,
and propose smoothed Bellman error embedding (SBEED) to solve it.
The authors provide "the first convergence guarantee for general non-linear function approximation, and analyze the algorithm's sample complexity", and experiment with several control tasks.
SBEED can be viewed as a de-biased version of PCL and generalizes PCL.

\subsection*{Recent Work}

\citet{Gu2017QProp} propose Q-Prop to take advantage of the stability of policy gradient and the sample efficiency of off-policy learning.
Q-Prop utilizes a Taylor expansion of the off-policy critic as a control variate, 
which gives an analytical gradient term through critic, and a Monte Carlo policy gradient term.

\citet{ODonoghue2017} propose PGQL to combine policy gradient with off-policy Q-learning, 
to benefit from experience replay.  
The authors also show that action value fitting techniques and actor-critic methods are equivalent, and interprete regularized policy gradient techniques as advantage function learning algorithms. 

\citet{Gu2017Interpolated} show that the interpolation of off-policy updates with value function estimation and 
on-policy policy gradient updates can satisfy performance guarantee.
The authors employ control variate methods for analysis, and design a family of policy gradient algorithms,
with several recent ones as special cases, including Q-Prop, PGQL, and 
ACER~\citep{Wang2017}.
The author study the correspondence between the empirical performance and the degree of mixture of off-policy gradient estimates with on-policy samples, 
on several continuous tasks. 

\subsection{Benchmark Results}
\label{policy:benchmark}

\citet{Duan2016} present a benchmark study for continuous control tasks, including classic tasks like cart-pole, tasks with very large state and action spaces such as 3D humanoid locomotion and tasks with partial observations, and tasks with hierarchical structure.
The authors implement and compare various algorithms, including batch algorithms: REINFORCE, Truncated Natural Policy Gradient (TNPG), Reward-Weighted Regression (RWR), Relative Entropy Policy Search (REPS), Trust Region Policy Optimization (TRPO), Cross Entropy Method (CEM), Covariance Matrix Adaption Evolution Strategy (CMA-ES); and online algorithms: Deep Deterministic Policy Gradient (DDPG); and recurrent variants of batch algorithms. 
See the open source at  \url{https://github.com/rllab/rllab}.


\citet{Henderson2018} investigate reproducibility, experimental techniques, and reporting procedures for deep RL.
The authors show that hyperparameters, including 
network architecture and reward scale, 
random seeds and trials,
environments (like Hopper or HalfCheetah etc. in OpenAI Baseline), 
and codebases
influenced experimental results.
This causes difficulties for reproducing deep RL results.

\citet{Tassa2018} present the DeepMind Control Suite, a set of continuous tasks,
implemented in Python, based on the MuJoCo physics engine.
\citet{Tassa2018} include benchmarks for A3C, DDPG, and 
distributed distributional deterministic policy gradients (D4PG)~\citep{Barth-Maron2018}.
The open source is at, \url{https://github.com/deepmind/dm_control},
and a video showing the tasks is at, \url{https://youtu.be/rAai4QzcYbs}.

\clearpage

\section{Reward}
\label{reward}

Rewards provide evaluative feedbacks for a RL agent to make decisions. 
Reward function is a mathematical formulation for rewards. 

Rewards may be sparse so that it is challenging for learning algorithms, 
e.g., in computer Go, a reward occurs at the end of a game.  
Hindsight experience replay~\citep{Andrychowicz2017} is a way to handle sparse rewards, as we discuss in Chapter~\ref{value}.
Unsupervised auxiliary learning~\citep{Jaderberg2017} is an unsupervised ways to harness environmental signals, as we discuss in Chapter~\ref{unsupervised}.  
Reward shaping is to modify reward function to facilitate learning while maintaining optimal policy. 
\citet{Ng1999} show that potential-based reward shaping can maintain optimality of the policy.
Reward shaping is usually a manual endeavour.
\citet{Jaderberg2018Quake} employ a learning approach in an end-to-end training pipeline.  

Reward functions may not be available for some RL problems.
In imitation learning,  an agent learns to perform a task from expert demonstrations, with samples of trajectories from the expert, without reinforcement signal.
Two main approaches for imitation learning are behavioral cloning and inverse reinforcement learning. 
Behavioral cloning, 
or learning from demonstration, maps state-action pairs from expert trajectories to a policy, maybe as supervised learning, without learning the reward function~\citep{Ho2016Imitation, Ho2016}. 

\citet{Levine2018} discusses about imitation learning and RL.
Pure imitation learning
is supervised learning, stable and well-studied;
however, it encounters the issue of distributional shift,
and it can not perform better than the demonstrations.
Pure RL is unbiased, and can improve until optimal,
however, with challenging issues of exploration and optimization.
Initialization with imitation learning then fine-tuning with RL can take advantage of both approaches;
however, it can forget initialization from demonstration due to distributional shift.
Pure RL with demonstrations as off-policy data is still RL, keeping advantages of RL;
however, demonstrations may not always help.
A hybrid objective including both RL and imitation objectives, 
can take advantage of both, do not forget demonstrations;
however, it is not pure RL, may be biased, and may require considerable tuning.

Inverse reinforcement learning (IRL) is the problem of determining a reward function given observations of optimal behaviour~\citep{Ng2000}. 
\citet{Abbeel2004} approach apprenticeship learning via IRL. 
\citet{Finn2016} study inverse optimal cost.
Probabilistic approaches are developed for IRL with maximum entropy~\citep{Ziebart2008} and maximum causal entropy~\citep{Ziebart2010} to deal with uncertainty in noisy and imperfect demonstrations.
\citet{Ross2010} reduce imitation learning and structured prediction~\citep{Daume2009} to no-regret online learning, and propose DAGGER, which requires interaction with the expert. 
\citet{Syed2007}, \citet{Syed2008}, and \citet{Syed2010} study apprenticeship learning with linear programming, game theory, and reduction to classification.
 
A reward function may not represent the intention of the designer.
A negative side effect of a misspecified reward refers to potential poor behaviours resulting from missing important aspects~\citep{Amodei2016}.
\citet{Hadfield-Menell2017} give an old example about the wish of King Midas, that everything he touched, turned into gold. Unfortunately, his intention did not include food, family members, and many more.
\citet{Russell2009} give an example that a vacuum cleaner collects more dust to receive more rewards by ejecting collected dust.
  
\citet{Singh2009} and \citet{Singh2010} 
discuss fundamental issues like, 
homeostatic vs. non-homeostatic (heterostatic) theories,
primary rewards vs. conditioned or secondary rewards, 
internal vs. external environments, 
intrinsic vs. extrinsic motivation, and,
intrinsic vs. extrinsic reward.
The authors then formulate an optimal reward computational framework.
\citet{Oudeyer2007} present a typology of computational approaches to these concepts.
 
See~\citet{Yue2018ICMLtutorial} for  a tutorial on imitation learning,
\citet{Rhinehart2018} for a tutorial on IRL for computer vision,
and, \citet{Argall2009} for a survey of robot learning from demonstration. 
See NIPS 2018 Workshop on Imitation Learning and its Challenges in Robotics, at \url{https://sites.google.com/view/nips18-ilr}.


In the following, we first discuss 
RL methods with and without reward learning respectively, when there is no reward function given. 
We then discuss an approach to handle complex reward functions.

See also 
\citet{Amin2017}, 
\citet{HuangJiexi2018NIPS},
\citet{Leike2018NIPS},
\citet{Merel2017}, 
\citet{Stadie2017}, 
\citet{Su2016}, 
\citet{Wang2017Imitation},
and, \citet{ZhengZeyu2018NIPS}.

We discuss robotics with imitation learning in Section~\ref{robotics:imitation},
including \citet{Duan2017OneShot, Finn2017One, Yu2018One, Finn2016}.
\citet{HuZhiting2018NIPS} leverage IRL techniques to learn knowledge constraints in deep generative models. 

\subsection*{Reward Learning}
\citet{Hadfield-Menell2016} observe flaws with IRL: 
a robot may take a human's reward function as its own, 
e.g., a robot may learn to drink coffee, rather than learn to cook coffee;
and, by assuming optimal demonstrations which achieve a task efficiently,
a robot may miss chances to learn useful behaviours,
e.g., a robot learns to cook coffee by passive observing would miss chances to learn many useful skills during the process of cooking coffee by active teaching and learning.

The authors then propose a cooperative inverse reinforcement learning (CIRL) game for the value alignment problem.
CIRL is a two-player game of partial information, with a human, knowing the reward function, a robot, not knowing it,
and the robot's payoff is human's reward.
An optimal solution to CIRL maximizes the human reward, 
and may involve active teaching by the human and active learning by the robot.
The authors reduce finding an optimal policy pair for human and robot to the solution of a single agent POMDP problem,
prove that optimality in isolation, like apprenticeship learning and inverse reinforcement learning, is suboptimal in CIRL,
and present an approximate algorithm to solve CIRL.


\citet{Hadfield-Menell2017} introduce inverse reward design (IRD),
to infer the true reward function, based on a designed reward function, an intended decision problem, e.g., an MDP, and a set of possible reward functions. 
The authors propose approximate algorithms to solve the IRD problem, 
and experiment with the risk-averse behaviour derived from planning with the resulting reward function.
Experiments show that IRD reduces chances of undesirable behaviours like
misspecified reward functions and reward hacking.

\citet{Christiano2017} propose to learn a reward function based on human preferences by comparing pairs of trajectory segments.
The proposed method maintains a policy and an estimated reward function, approximated by deep neural networks.
The networks are updated asynchronously with three processes iteratively: 
1) produce trajectories with the current policy, and the policy is optimized with a traditional RL method, 
2) select pairs of segments from trajectories, obtain human preferences, and,
3) optimize the reward function with supervised learning based on human preferences.
Experiments show the proposed method can solve complex RL problems like Atari games and simulated robot locomotion.

\subsection*{Learning from Demonstration}
\label{demonstration}

Here we discuss several recent papers without reward learning.


We first discuss deep Q-learning from demonstrations.
\citet{Hester2018} propose deep Q-learning from demonstrations (DQfD) to attempt to accelerate learning by leveraging demonstration data, using a combination of temporal difference (TD), supervised, and regularized losses. In DQfQ, reward signal is not available for demonstration data; however, it is available in Q-learning. The supervised large margin classification loss enables the policy derived from the learned value function to imitate the demonstrator; the TD loss enables the validity of value function according to the Bellman equation and its further use for learning with RL; the regularization loss function on network weights and biases prevents overfitting on small demonstration dataset. In the pre-training phase, DQfD trains only on demonstration data, to obtain a policy imitating the demonstrator and a value function for continual RL learning. After that, DQfD self-generates samples, and mixes them with demonstration data according to certain proportion to obtain training data. Experiments on Atari games show DQfD in general has better initial performance, more average rewards, and learns faster than DQN.

In AlphaGo~\citep{Silver-AlphaGo-2016}, as we discuss in Section~\ref{games:AlphaGo}, the supervised learning policy network is learned from expert moves as learning from demonstration; the results initialize the RL policy network. See also \citet{Kim2014, Perez2017, Vecerik2017}. 


Now we discuss generative adversarial imitation learning.
With IRL, an agent learns a reward function first, then from which derives an optimal policy.  Many IRL algorithms have high time complexity, with a RL problem in the inner loop.
\citet{Ho2016} propose the generative adversarial imitation learning (GAIL) algorithm to learn policies directly from data, bypassing the intermediate IRL step. 
Generative adversarial training is deployed to fit the discriminator, about the distribution of states and actions that defines expert behavior, and the generator, representing the policy. 
Generative adversarial networks (GANs) are a recent unsupervised learning framework, 
which we discuss in Section~\ref{unsupervised:GAN}. 

GAIL finds a policy $\pi_{\bm{\theta}}$ so that a discriminator $\mathcal{D}_R$ can not distinguish states following the expert policy $\pi_E$ and states following the imitator policy $\pi_{\bm{\theta}}$, hence forcing $\mathcal{D}_R$ to take 0.5 in all cases and $\pi_{\bm{\theta}}$ not distinguishable from $\pi_E$ in the equillibrium. Such a game is formulated as:

\[\max_{\pi_{\bm{\theta}}} \min_{\mathcal{D}_R} -E_{\pi_{\bm{\theta}}} [\log \mathcal{D}_R(s)] -E_{\pi_E} [\log (1-\mathcal{D}_R(s))]\]  

The authors represent both $\pi_{\bm{\theta}}$ and $\mathcal{D}_R$ as deep neural networks, and find an optimal solution by repeatedly performing gradient updates on each of them.  $\mathcal{D}_R$ can be trained with supervised learning with a data set formed from traces from a current  $\pi_{\bm{\theta}}$ and expert traces. For a fixed $\mathcal{D}_R$, an optimal $\pi_{\bm{\theta}}$ is sought. Hence it is a policy optimization problem, with $-\log \mathcal{D}_R(s)$ as the reward. 
The authors train $\pi_{\bm{\theta}}$ by trust region policy optimization~\citep{Schulman2015}, 
and experiment on various physics-based control tasks with good performance. 

\citet{Li2017InfoGAIL} extend GAIL to InfoGAIL 
for not only imitating, but also learning interpretable representations of complex behaviours,
by augmenting the objective with a mutual information term between latent variables and trajectories, i.e., the observed state-action pairs.
InfoGAIL learns a policy in which more abstract, high level latent variables would control low level actions.
The authors experiment on autonomous highway driving using TORCS driving simulator~\citep{TORCS}.
See the open source at \url{https://github.com/ermongroup/InfoGAIL}.
\citet{SongJiaming2018NIPS} extend GAIL to the multi-agent setting.

\subsection*{Reward Manipulating}

\citet{vanSeijen2017} propose hybrid reward architecture (HRA)
to tackle the issue that optimal value function may not be embedded in low dimensional representation,
by decomposing reward function into components, and learning value functions for them separately.
Each component may be embedded in a subset of all features,
so its value function may be learned and represented in a low dimensional space relatively easily.
HRA agents learn with sample trajectories using off-policy learning in parallel, similar to Horde~\citep{Sutton2011}.
Experiments on Atari game Ms. Pac-Man show above-human performance.
See open source at \url{https://github.com/Maluuba/hra}.

\clearpage

\section{Model}
\label{model}

A model is an agent's representation of the environment, including the state transition model and the reward model. Usually we assume the reward model is known. We discuss how to handle unknown reward models in Chapter~\ref{reward}.

Model-free RL approaches handle unknown dynamical systems, which usually requires large number of samples.
This may work well for problems with good simulators to sample data, e.g., Atari games and the game of Go.
However,  this may be costly or prohibitive for real physical systems. 

Model-based RL approaches learn  value function and/or policy in a data-efficient way, however, they may suffer from issues of model identification, so that the estimated models may not be accurate, and the performance is limited by the estimated model.
Planning constructs a value function or a policy usually with a model.

Combining model-free RL with on-line planning can improve value function estimation. 
\citet{Sutton1990} proposes Dyna to integrate learning and planning, 
by learning from both real experiences and simulated trajectories from a learned model.


Monte Carlo tree search (MCTS)~\citep{Browne2012, Gelly2007, Gelly12} and upper confidence bounds (UCB)~\citep{Auer2002} applied to trees (UCT)~\citep{Kocsis2006} are important techniques for planning.
A typical MCTS builds a partial tree starting from the current state, in the following stages: 1) select a promising node to explore further, 2) expand a leaf node and collect statistics, 3) evaluate a leaf node, e.g., by a rollout policy, or some evaluation function, 4) backup evaluations to update the action values.  An action is then selected.   
A prominent example is AlphaGo~\citep{Silver-AlphaGo-2016, Silver-AlphaGo-2017} as we will discuss in Section~\ref{games:AlphaGo}. 
\citet{SunWen2018NIPS} study dual policy iteration.

Several papers design new deep neural networks architectures for RL problems, e.g., 
value iteration networks (VIN), 
predictron, 
value prediction network (VPN), 
TreeQN and ATreeC, 
imagination-augmented agents (IA2), 
temporal difference models (TDMs), 
MCTSnets, 
which we will discuss below,
as well as dueling network 
as we discuss in  Chapter~\ref{value}.
VIN, predictron, and, MCTSnets follow the techniques of learning to learn, which we discuss in Chapter~\ref{meta}.

R-MAX \citep{Brafman2002} and E$^3$ \citep{Kearns2002} achieve guaranteed efficiency for tabular cases.
\citet{Li2011KWIK} present the "know what it knows" framework. 
\citet{Deisenroth2011} present probabilistic inference for learning control (PILCO),
and \citet{McAllister2017} extend PILCO to POMDPs.
See papers about model predictive control (MPC)~\citep{Amos2018NIPS, Finn2017MPC, Lenz2015}.
See more papers, e.g., 
\citet{Berkenkamp2017}, 
\citet{Buckman2018NIPS}, 
\citet{Chebotar2017}, 
\citet{Chua2018NIPS},  
\citet{deAvilaBelbutePeres2018NIPS}, 
\citet{Farahmand2018NIPS}, 
\citet{Ha2018world}, 
\citet{Haber2018NIPS},  
\citet{Henaff2017}, 
\citet{Watter2015}.
We will discuss guided policy search (GPS)~\citep{Levine2016}, 
and, \citet{Chebotar2017} in Chapter~\ref{robotics}


\citet{Sutton2018TTT} discusses that "planning with a learned model" means "Intelligence is just knowing a lot about the world, being able to use that knowledge flexibly to achieve goals", and,
mentions that "planning $\approx$ reasoning $\approx$ thought", 
and "world model $\approx$ knowledge $\approx$ propositions $\approx$ facts".
He quotes from Yann LeCun, "obstacles to AI: learning models of the world, learning to reason and plan",
and "predictive learning $\approx$ unsupervised learning $\approx$ model-based RL".
He also quotes from Yoshua Bengio's most important next step in AI, "learning how the world ticks", 
and "predictive, causal, explanatory models with latent variables ...".
He lists the following as some answers to the problem of planning with a learned model:
function approximation, 
off-policy learning,
Dyna, linear Dyna, 
non-linear Dyna,
GVFs, Horde,
options, option models,
prioritized sweeping,
intrinsic motivation,
curiosity,
recognizers,
predictive state representations,
TD networks~\citep{Sutton2004},
TD networks with options~\citep{Sutton2005}, and,
propagation with valuableness.
\citet{LeCun2018} also talks about world model, highlighting the role of self-supervised learning.


\citet{Geffner2018} discusses model-free learners and model-based solvers, and planners as particular solvers.
Learners are able to infer behaviour and functions from experience and data, 
solvers are able to address well-defined but intractable models like classical planning and POMDPs, 
and, planners are particular solvers for models with goal-directed behaviours.
\citet{Geffner2018} makes connections between model-free learners vs. model-based solvers,
and the two systems in current theories of human mind~\citep{Kahneman2011}:
System 1 with a fast, opaque, and inflexible intuitive mind, 
vs. System 2 with a slow, transparent, and flexible analytical mind.

See \citet{Finn2017TalkModelRL} for a tutorial,
and \citet{Silver2015Course} and  \citet{Levine2018} for lectures,
 about model-based (deep) RL.
See NIPS 2018 Workshop on Modeling the Physical World: Perception, Learning, and Control
at \url{http://phys2018.csail.mit.edu}. 
We discuss \citet{Lake2016} in Section~\ref{representation:knowledge},
and, scene understanding and physics model learning in Section~\ref{CV:scene}. 


\subsection*{Model-based RL}

We discuss several recent papers about model-based RL.
We may roughly have
methods with RL flavour, e.g., Dyna, VPN, IA2, using TD methods;
methods with optimal control flavour, e.g., GPS, NAF, using local models like linear quadratic regulator (LQR) and MPC;
and, methods with physics flavour, e.g., those with physics models as discussed in \citet{Lake2016}. 


In policy gradient methods, the gradient can be estimated either by likelihood ratio method as in REINFORCE, or by value gradient methods with backpropagation as in DPG/DDPG. 
Value gradients methods are used for deterministic policies. 
\citet{Heess2015} propose stochastic value gradients (SVG) for stochastic policies,
to combine advantages of model-free and model-based methods, and to avoid their shortcomings.
SVG treats noise variables in Bellman equation as exogenous inputs, 
allowing direct differentiation w.r.t. states.
This results in a spectrum of policy gradient algorithms, 
ranging from model-free ones with value functions, 
to model-based ones without value functions.
SVG learns model, value function, and policy jointly,
with neural networks trained using experience from interactions with the environment.
SVG mitigates compounded model errors by computing value gradients 
with real trajectories from the environment rather than those from an inaccurate, estimated model.
SVG uses models for computing policy gradient, but not for prediction.
\citet{Heess2015} experiment with physics-based continuous control tasks in simulation.


\citet{Oh2017VPN} propose value prediction network (VPN) to integrate model-free and model-based RL into a neural network. 
VPN learns a dynamic model to train abstract states to predict future values, rewards, and discount, rather than future observations as in typical model-based methods.
The author propose to train VPN by TD search~\citep{Silver2012} and multi-step Q learning. 
In VPNs, values are predicted with Q-learning, rewards are predicted with supervised learning, and lookahead planning are performed for choosing actions and computing target Q-values.
VPN is evaluate on a 2D navigation collect task and Atari games. 

 
\citet{Pong2018} propose temporal difference models (TDMs) to combine benefits of model-free and model-based RL.  
TDMs are general value functions, as discussed in Section~\ref{value:GVF}.
TDMs can be trained with model-free off-policy learning, and be used for model-based control.
TDM learning interpolates between model-free and model-based learning,
seeking to achieve sample efficiency in model-based learning, and at the same time, avoiding model bias.
\citet{Pong2018} evaluate TDMs on both simulated and real-world robot tasks.


\citet{Farquhar2018} propose TreeQN, using a differentiable, recursive tree structure neural network architecture to map the encoded state to the predicted action value Q function, for discrete actions. 
TreeQN uses such recursive model to refine the estimate of Q function, with the learned transition model, reward function, and value function, by tree transitioning, and value prediction \& backup steps in the recursive tree structure neural network.
TreeQN takes advantage of the prior knowledge that Q values are induced by MDPs.
In contrast, DQN uses fully connected layers, not implementing such inductive bias.  
\citet{Farquhar2018} also propose ATreeC, an actor-critic variant.
The authors evaluate TreeQN and ATreeC in a box-pushing environment and on Atari games.


\citet{Weber2017} propose imagination-augmented agents (IA2), a neural network architecture, to combine model-free and model-based RL.
IA2 learns to augment  model-free decisions by interpreting environment models.


\citet{Gu2016} propose normalized advantage functions (NAF) to enable experience replay with Q-learning for continuous task, and  propose to refit local linear models iteratively. 
NAF extends Dyna to continuous tasks.
The authors evaluate NAF on several simulated MuJoCo robotic tasks.


\citet{Hester2017texplore} propose variance-and-novelty-intrinsic-rewards algorithm (TEXPLORE-VANAIR), a model-based RL algorithm with intrinsic motivations. 
The authors study two intrinsic motivations, one for model uncertainty, and another for acquiring novel experiences new to the model. 
The authors conduct empirical study on a simulated domain and a real world robot and show good results.


\citet{Leonetti2016} propose domain approximation for reinforcement learning (DARLING), 
taking advantage of both RL and planning, 
so that the agent can adapt to the environment, 
and the agent's behaviour is constrained to reasonable choices. 
The authors perform evaluation on a service robot for tasks in an office building.

\subsection*{Planning}

We discuss several recent papers about planning.


\citet{Guez2018} propose to learn to search with MCTSnets, a neural network architecture that incorporates simulation-based search, using a vector embedding for expansion, evaluation, and backup.
The authors propose to jointly optimize in MCTSnets a simulation policy for where to traverse in the simulation, an evaluation network for what to evaluate in the reached states, and a backup network for how to backup evaluations, end-to-end with a gradient-based approach. 
The authors experiment MCTSnets with a classical planning problem Sokoban.


\citet{Tamar2016} introduce value iteration networks (VIN), a fully differentiable CNN planning module to approximate the value iteration algorithm, to learn to plan, e.g, policies in RL. In contrast to conventional planning, VIN is model-free, where reward and transition probability are part of the neural network to be learned.
VIN can be trained end-to-end with backpropagation. VIN can generalize in a diverse set of tasks: simple gridworlds, Mars Rover Navigation, continuous control and WebNav Challenge for Wikipedia links navigation~\citep{Nogueira2016}. 
VIN plans via value iteration over the full state space, and with local transition dynamics for states, e.g., 2D domains,
which limits applicability of VIN.    
\citet{LeeLisa2018ICML} propose gated path planning networks to extend VIN. 


\citet{Silver-Predictron-2016} propose the predictron to integrate learning and planning into one end-to-end training procedure with raw input in Markov reward process (MRP), 
which can be regarded as Markov decision process without actions. 
Predictron rolls multiple planning steps of an abstract model represented by an MRP for value prediction;
in each step, predictron applies the model to an internal state, 
and generates a next state, reward, discount, and value estimate.
The predictron focuses on evaluation tasks in uncontrolled environments; however, it can be used as Q-network, e.g., in DQN, for control tasks.

\citet{Silver2012} propose temporal difference search (TD search) to combine TD learning with simulation based search.
TD search updates value function from simulated experience, and generalizes among states using value function approximation and bootstrapping. 
\citet{Xiao2018} propose memory-augmented MCTS.
\citet{Srinivas2018} propose universal planning networks.

\clearpage

\section{Exploration vs. Exploitation}
\label{exploration}

A fundamental tradeoff in RL is between exploration of uncertain policies and exploitation of the current best policy.
Online decision making faces a central issue: 
either exploiting the information collected so far to make the best decision, or exploring for more information.
In sequential decision making, we may have to sacrifice short-term losses to achieve long-term gains.
It is essential to collect enough information to make the best overall decisions.

There are several principles in trading off between exploration and exploitation, namely,
naive methods, 
optimism in the face of uncertainty, including,
optimistic initialization,
upper confidence bounds, and,
probability matching, 
and, information state search~\citep{Silver2015Course}.
These are developed in the settings of multi-armed bandit,
but are applicable to RL problems.

The multi-armed bandit problem is classical for studying exploration and exploitation.
It is defined by a tuple $<\mathcal{A}, \mathcal{R}>$,
where $\mathcal{A}$ is a given set of arms, or actions, 
and $\mathcal{R}(r|a) = \mathbb{P}(r|a)$ is a probability distribution over rewards, unknown to the agent.
At each step $t$, the agent selects an action $a_t \in \mathcal{A}$,
receives a reward $r_t \sim \mathcal{R}(\cdot|a_t)$ from the environment,
and the goal is to maximize the cumulative reward $\sum_{\tau=1}^t r_{\tau}$.

The action-value function is the expected reward for action $a$, 
$Q(a) = \mathbb{E}[r|a]$.
The optimal value is 
$V^* = Q(a^*) = \max_{a \in \mathcal{A}} Q(a)$.
The regret is one step loss, 
\begin{equation}
l_t = \mathbb{E}[V^* - Q(a_t)].
\end{equation}
The total regret until time step $t$ is then 
\begin{equation}
L_t = \mathbb{E}[\sum_{\tau=1}^t V^* - Q(a_{\tau})].
\end{equation}
The maximum cumulative reward is the minimum total regret.

Denote $N_t(a)$ as the expected number of selecting action $a$ until time step $t$.
The greedy algorithm selects the action with the highest value, $a_t^* = \argmax_{a \in \mathcal{A}} \hat{Q}_t(a)$,
where $\hat{Q}_t(a)$ is an estimate of $Q(a)$, 
e.g., by Monte Carlo evaluation,  
\begin{equation}
\hat{Q}_t(a) = \frac{1}{N_t(a)} \sum_{\tau=1}^t r_{\tau} 1( a_{\tau} = a).
\end{equation}

The greedy algorithm may stick to a suboptimal action.
However, $\epsilon$-greedy, where $\epsilon \in (0,1)$, can ensure a minimum regret with a constant $\epsilon$.
In $\epsilon$-greedy, an agent selects a greedy action $a = \argmax_{a \in \mathcal{A}} \hat{Q}(a)$, 
with probability $1-\epsilon$;
and selects a random action with probability $\epsilon$.

A simple and practical idea for optimistic initialization is to initialize action value $Q(a)$ to a high value,
then update it with incremental Monte Carlo evaluation,
\begin{equation}
\hat{Q}_t(a) = \hat{Q}_{t-1}(a) + \frac{1}{N_t(a)} (r_t - \hat{Q}_{t-1}(a)).
\end{equation}
This encourages exploration in an early stage, but may stick to a suboptimal action.

Next we discuss upper confidence bounds (UCB)~\citep{Auer2002},
an important result in bandit problems.
Its extension, UCT for search trees~\citep{Kocsis2006}, in particular, in Monte Carlo tree search (MCTS)~\citep{Browne2012, Gelly2007, Gelly12}, plays important roles in many problems, 
including AlphaGo~\citep{Silver-AlphaGo-2016, Silver-AlphaGo-2017}.

We estimate an upper confidence $\hat{U}_t(a)$ for each action,
to have $Q(a) \leq \hat{Q}_t(a) + \hat{U}_t(a)$ with a high probability.
When $N_t(a)$ is small, $\hat{U}_t(a)$ is large, i.e., the estimated value is uncertain;
when $N_t(a)$ is large, $\hat{U}_t(a)$ is small, i.e., the estimated value is close to true value.
We want to select an action to maximize the upper confidence bound,
\begin{equation}
a_t = \max_{a \in \mathcal{A}}  \hat{Q}_t(a) + \hat{U}_t(a).
\end{equation}
We need to establish a theoretical guarantee with Hoeffding's inequality theorem, which is: 
let $X_1, \dots, X_t$ be i.i.d. random variables in [0,1], 
and let $\bar{X}_t = \frac{1}{\tau} \sum_{\tau = 1}^t X_{\tau}$ be the sample mean. Then,
\begin{equation}
\mathbb{P}[ \mathbb{E}[X] > \bar{X}_t ] \leq e^{-2tu^2}.
\end{equation}
 With Hoeffding's inequality, conditioned on selecting action $a$, we have,
 \begin{equation}
 \mathbb{P} [Q(a) > \hat{Q}_t(a) + \hat{U}_t(a) ]< e^{-2N_t(a)U_t(a)^2}.
 \end{equation}
Choose a probability $p = e^{-2N_t(a)U_t(a)^2}$, 
so that $Q(a) > \hat{Q}_t(a) + \hat{U}_t(a)$, i.e., the true value exceeds UCB, 
hence, $U_t(a) = \sqrt{\frac{-\log p}{2N_t(a)}}$. 
Choose a schedule for $p$ as observing more rewards, e.g., $p = t^{-4}$, 
hence, $U_t(a) = \sqrt{\frac{2 \log p}{N_t(a)}}$.
This guarantees that, as $t \rightarrow \infty$, we select optimal actions. 
Thus, we obtain the UCB1 algorithm, 
\begin{equation}
a_t = \argmax_{a \in \mathcal{A}} Q(a) + \sqrt{\frac{2 \log t}{N_t(a)}}.
\end{equation}
The UCB algorithm can achieve logarithmic asymptotic total regret,
better than linear total regret achievable by $\epsilon$-greedy and optimistic initialization~\citep{Auer2002}.

As shown above, UCB employs $\sqrt{2 \log t/N_t(a)}$ as exploration bonus to encourage less discovered actions.
Model-based interval estimation with exploration bonuses (MBIE-EB)~\citep{Strehl2008} employs $\sqrt{1/N_t(a)}$;
and Bayesian exploration bonus (BEB)~\citep{Kolter2009} employs $1/N_t(a)$.

In probability matching, 
we select an action $a$ according to the probability that $a$ is the optimal action with the largest value.
It is optimistic under uncertainty, and uncertain actions tend to have higher probabilities of having the largest value.
Thompson sampling implements probability matching, 
dealing with the difficulty of analytical computation with posterior distributions.
Thompson sampling uses Bayes law to compute the posterior distribution,
samples a reward distribution from the posterior,
evaluates action value function,
and, selects the action that maximizes value estimates on samples.

Gittins indices are a Bayesian model-based RL method for solving information state space bandits.
It is known as Bayes-adaptive RL, and finds Bayes-optimal exploration and exploitation trade off w.r.t. a prior distribution.
The computation complexity may be prohibitive for large problems.

The above discussions are in the setting of multi-armed bandits.
They do not pay attention to additional information.
However, such feature-based exploration vs exploitation problems are challenging~\citep{Auer2002N, Langford2007}. 
\citet{Li2010} introduce contextual bandit and design algorithms based on UCB. 

The techniques for multi-armed bandits are also applicable for full RL problems.
A naive exploration technique, $\epsilon$-greedy is widely used.
In model-free RL, we can initialize action value function $Q(s,a) \leftarrow \frac{r_{max}}{1-\gamma}$,
where $r_{max}$ is the maximal value of reward. 
\citet{Brafman2002} present R-MAX, a model-based RL, 
optimistically initializing all actions in all states to the maximal reward. 
UCB can work with model-free and model-based RL methods.
\citet{Lipton2018BBQ} propose to add variance information to DQN, which is then used to guide exploration following the principle of Thompson sampling.
\citet{Guez2014} propose a simulation-based search approach for Bayes-adaptive MDPs augmented with information states.

A RL agent usually uses  exploration to reduce its uncertainty about the reward function and transition probabilities of the environment. In tabular cases, this uncertainty can be quantified as confidence intervals or posterior of environment parameters, which are related to the state-action visit counts. 
An example is MBIE-EB~\citep{Strehl2008}.
\citet{Brafman2002}, \citet{Jaksch2010}, and \citet{Strehl2008} provide theoretical guarantee for tabular cases.

Intrinsic motivation~\citep{Barto2013, Schmidhuber2010, Oudeyer2007} suggests to explore based on the concepts of curiosity and surprise, so that actions will transition to surprising states that may cause large updates to the environment dynamics models. One particular example of measuring surprise is by the change in prediction error in learning process~\citep{Schmidhuber1991}, as studied recently in \citet{Bellemare2016}. 
Intrinsic motivation methods do not require Markov property and tabular representation as count-based methods. 
Watch a video~\citep{Barto2017}.
See ICML 2018 workshop on Exploration in RL at, \url{https://sites.google.com/view/erl-2018/home}, and \url{https://goo.gl/yxf16n} for videos.

\citet{Levine2018} discusses three classes of exploration methods in deep RL:
1) optimistic exploration methods, 
which estimate state visitation frequencies or novelty,
typically with exploration bonuses, e.g.,  \citet{Bellemare2016}, \citet{Fu2017}, \citet{Schmidhuber1991}, and \citet{Tang2017};
2)  Thompson sampling methods,
which learn distribution over Q-functions or policies,
then act according to samples, e.g., \citet{Osband2016};
3) information gain methods,
which reason about information gain from visiting new states, e.g., \citet{Houthooft2016}.
All these three methods follow the principle of optimism in the face of uncertainty.

In the above, we discuss background of exploration and exploitation largely based on
Lecture 9: Exploration and Exploitation in \citet{Silver2015Course},
as well as,
the lecture about exploration in \citet{Levine2018},
Chapter 2 in~\citep{Sutton2018} about multi-armed bandits,
and, relevant papers. 
\citet{Lattimore2018} is about bandit algorithms.
\citet{Li2012exploration} surveys theoretical approaches to exploration efficiency in RL.

In the following we discuss several recent work about exploration in the setting of large scale RL problems, in particular deep RL.

\subsection*{Count-based Methods}


With the count-based exploration, a RL agent uses visit counts to guide its behaviour to reduce uncertainty. 
However, count-based methods are not directly useful in large domains. 
\citet{Bellemare2016} propose pseudo-count, a density model over the state space, for exploration with function approximation, to unify count-based exploration and intrinsic motivation, by introducing information gain, to relate to confidence intervals in count-based exploration, 
and to relate  learning progress in intrinsic motivation. 
The authors establish pseudo-count's theoretical advantage over previous intrinsic motivation methods, 
implement it with a density model, 
use it as exploration bonuses in MBIE-EB~\citep{Strehl2008}, 
in experience replay and actor-critic settings,
and study its empirical performance with Atari games. 

\citet{Ostrovski2017} further study the approach of pseudo-count~\citep{Bellemare2016}
w.r.t. importance of density model selection, modelling assumptions, and role of mixed Monte Carlo update, 
and propose to use a neural density model for images, PixelCNN~\citep{vandenOord2016}, 
for supplying pseudo-count, and combine it with various agent architectures, 
including DQN~\citep{Mnih-DQN-2015} and Reactor~\citep{Gruslys2017}.
The authors observe that mixed Monte Carlo update facilitate exploration in settings with sparse rewards, 
like in the game of Montezuma's Revenge.


\citet{Tang2017} propose to implement the count-based exploration method 
by mapping states to hash codes to count occurrences with a hash table,
and the counts are used for reward bonus to guide exploration.
The authors experiment with simple hash functions and a learned domain-dependent hash code
 on both Atari games and continuous control tasks.

\subsection*{Intrinsic Motivation}


\citet{Houthooft2016} propose variational information maximizing exploration (VIME),
a curiosity-driven exploration approach
to simultaneously optimizing both external reward and intrinsic surprise, 
for continuous state and action spaces.
The authors propose to measure the information gain with variance inference, 
and to approximate the posterior distribution of an agent's internal belief of environment dynamics, 
represented with Bayesian neural networks.
The authors evaluate VIME with various continuous control tasks and algorithms.

\citet{Pathak2017} study curiosity-driven exploration with intrinsic reward signal for predicting the result of its actions 
with self-supervised learning,
and experiment with VizDoom and Super Mario Bros.

\subsection*{More Work}


\citet{Osband2016} propose bootstrapped DQN to combine deep exploration with deep neural networks to achieve efficient learning. 
The authors use randomized value functions to implement Thompson sampling,
to enable exploration with non-linear function approximation, such as deep neural networks, 
and a policy is selected randomly according to its probability being optimal.
The authors implement bootstrapped DQN by building multiple estimates of the action value function in parallel,
and each estimate is trained with its own target network.
The authors evaluate the performance of bootstrapped DQN with Atari games. 


\citet{Nachum2017} propose under-appreciated reward exploration (UREX) to avoid the ineffective, undirected exploration strategies of the reward landscape, 
as in $\epsilon$-greedy and entropy regularization policy gradient, 
and to promote directed exploration of the regions, in which the log-probability of an action sequence under the current policy under-estimates the resulting reward. 
UREX results from importance sampling from the optimal policy, 
and combines a mode seeking and a mean seeking terms to tradeoff exploration and exploitation.  
The authors implement UREX with minor modifications to REINFORCE, 
and validate it, for the first time with a RL method,  on several algorithmic tasks. 
UREX is an effort for symbolic deep RL.

\citet{Azar2017} study the problem of provably optimal exploration for finite horizon MDPs.
\citet{Fu2017} propose novelty detection with discriminative modeling for exploration.
\citet{Fortunato2018} propose NoisyNet for efficient exploration by adding parametric noises to weights of deep neural networks.
\citet{Jiang2017} study systematic exploration for contextual decision processes (CDPs).
See also 
\citet{Dimakopoulou2018NIPS},
\citet{DongShi2018NIPS},
\citet{Gupta2018NIPS},
\citet{Kumaraswamy2018NIPS}, 
\citet{Madhavan2018NIPS}, and,
\citet{Osband2018NIPS}.

Also note that maximum entropy RL helps exploration, as discussed in Section~\ref{policy:PGQ}.


\clearpage

\section{Representation}
\label{representation}

Representation is fundamental to reinforcement learning, machine learning, and AI in general.
For RL, it is relevant not only to function approximation for 
state/observation, action, value function, reward function, transition probability,
but also to agent~\citep{Albrechta2018, Rabinowitz2018ICML}, environment, and any element in a RL problem.
The "representation" in "representation learning"
basically refers to the "feature" in "feature engineering".
Representation learning is an approach to automatically find good features.
Here we discuss "representation" in a broader perspective, i.e., about any element in a RL problem. 
Besides the "feature" for function approximation, 
we also refer "representation" to
problem representation, like Markov decision process (MDP), partially observable Markov decision process (POMDP), and, predictive state representation (PSR),
and, moreover, for representing knowledge, reasoning, causality, and human intelligence,
either in function approximation, 
or in discovering new neural network architectures.
We attempt to use such a notion of "representation" to unify the roles deep learning has played, is playing, and would play in various aspects of deep reinforcement learning.

When the problem is small, both state and action can be accommodated in a table, we can use a tabular representation.
For large-scale problems, we need function approximation, to avoid curse of dimensionality.
One approach is linear function approximation, using basis functions like polynomial bases, tile-coding, radial basis functions, Fourier basis, and proto-value functions (PVFs), etc.
We also discuss representations for state distributions, in particular, successor representation, which is related to value function.

Recently, non-linear function approximations, in particular, deep neural networks, show exciting achievements. 
Common neural network structures include 
multiple layer perceptron (MLP), convolutional neural networks (CNNs), recurrent neural networks (RNNs), in particular long short time memory (LSTM) and gated recurrent unit (GRU), (variational) autoencoder, and capsules, etc.
There are new neural network architectures customized for RL problems, e.g., value iteration networks (VIN), predictron, and value prediction networks (VPN), etc.

General value function (GVF) is an approach to learn, represent, and use knowledge of the world.
Hierarchical representations, like options, feudal networks, and max-Q, etc. handle temporal abstraction.
Relational RL integrates statistical relational learning and reasoning with RL to handle entities and relations.

There are renewed interests in deploying or designing networks for reasoning, including graph neural networks (GNN),  graph networks (GN), relational networks (RNs), and compositional attention networks,  etc.  
There are discussions about addressing issues of current machine learning with causality, and incorporating more human intelligence into artificial intelligence.


Although there have been enormous efforts for representation, 
since reinforcement learning is fundamentally different from supervised learning and unsupervised learning, 
an optimal representation for RL is probably different from generic CNN and RNN,
thus it is desirable to search for an optimal representation for RL.
Our hypothesis is that this would follow a holistic approach, by considering perception and control together,
rather than treating them separately, e.g., by deciding a CNN to handle visual inputs, 
then fixing the network, and designing some procedure to find optimal weights for value function and/or policy.
Learning to learn techniques as we discuss in Chapter~\ref{meta} may play an important role here.

\subsection{Classics}
\label{representation:classics}

In this section, we discuss classical methods for representation,
as well as several papers for recent progress.
We discuss (linear) function approximation, 
which is usually for value function and policy approximation.
We then discuss representations for an RL problem description, 
i.e., state, transitions and reward, including,
partially observable Markov decision process (POMDP),
predictive state representation (PSR), and, 
contextual decision process (CDP).
We also discuss successor representation for state visitation, 
and work for state-action distribution. 

\subsection*{Function Approximation}

Here we discuss linear function approximation.
We discuss neural networks in Section~\ref{representation:NN}.
In linear function approximation, a value function is approximated by a linear combination of basis functions. 
Basis functions may take the form of polynomial bases, tile-coding, radial basis functions, Fourier basis, and proto-value functions, etc. 

In tile coding, tiles partition the input space exhaustively, and each tile is a binary feature. 
A tiling is such a partition. 
Each tiling represents a feature.
There are various ways to tile a space, like grid, log stripes, diagonal strips, and irregular, etc.~\citep{Sutton2018}

With radial basis functions (RBFs), typically, a feature $i$ has a Gaussian response $\phi_s(i) = \exp \big( -\frac{||s-c_i||^2}{2\sigma_i^2}\big)$,
where $s$ is the state, $c_i$ is the feature's prototypical or center state, and $\sigma_i$ is the feature's width~\citep{Sutton2018}.
When using RBFs as features for a linear function approximator, we have an RBF network.

\citet{Mahadevan2007} propose proto-value functions (PVFs), 
using "the eigenvectors of the graph Laplacian on an undirected graph formed from state transitions induced by the MDP". 
The authors then propose to learn PVFs and optimal policies jointly.

There are also papers with Gaussian processes~\citep{Rasmussen2006} and kernel methods~\citep{Scholkopf2001}, e.g.,  \citet{Ghavamzadeh2016} and \citet{Ormoneit2002}.

\subsection*{RL Problem Description}
 
 
Partially observable Markov decision process (POMDP)~\citep{Kaelbling1998} generalizes MDP.
In POMDP, an MDP determines system dynamics, with partial observability of underlying states.
A POMDP maintains a probability distribution over possible states, based on observations, their probabilities, and the MDP.
\citet{Hausknecht2015} propose deep recurrent Q-learning for POMDP.


Predictive state representation (PSR)~\citep{Littman2001} utilizes
vectors of predictions for action-observation sequences to represent states of dynamical systems.
The predictions relate directly to observable quantities, rather than hidden states.
PSRs do not have strong dependence on prior models as POMDP,
and, PSRs are more compact than POMDP, in dynamic systems which linear PSRs can represent. 
In fact, PSRs are more general than $n$th-order Markov models, hidden Markov models (HMM), and POMDP~\citep{Singh2004}.
Recently, \citet{Downey2017} present predictive state RNNs,
and \citet{Venkatraman2017} propose predictive state decoders,
both of which combine PSRs with RNN to take their advantages. 



\citet{Jiang2017} propose contextual decision processes (CDPs), RL problems with rich observations and function approximation, for systematic exploration.
The authors introduce the Bellman rank, a complexity measure, and provide a unified framework for
many problems in RL with low Bellman rank, e.g.,
tabular MDP, low-rank MDP, a POMDP with reactive value-functions, 
linear quadratic regulators (LQR), and reactive PSRs,
and show that these problems are PAC-learnable~\citep{Valiant1984,Strehl2009,Li2012exploration}.

\subsection*{State and State-Action Distribution}


\citet{Dayan1993} introduces successor representation (SR),  
\begin{equation}
\psi = \sum_{t=0}^{\infty} [\gamma \sum_{s'}\mathcal{P}(s'|s, \pi)]^t,
\end{equation}
for expected discounted future state visitations,
w.r.t. a given policy and a discount factor, and being independent of the reward.
In the vector form,
\begin{equation}
\bm{\psi} = \sum_{t=0}^{\infty} (\gamma \bm{P})^t = (\bm{I} - \gamma \bm{P})^{-1},
\end{equation}
where $\bm{P}$ is the transition matrix, and $\bm{I}$ is the identity matrix.
SR captures the dynamics of the MDP,
describing where the agent will traverse in the future, independent of the reward.
SR can be learned with algorithms like TD learning.
For value function, we have
\begin{equation}
v_{\pi}(s) = \mathbb{E}[R_t | s_t = s] = \mathbb{E}[r_{t+1} + \gamma R_{t+1} | s_t = s]  = \mathbb{E}[r_{t+1} | s_t = s] + \gamma \sum_{s'}\mathcal{P}(s'|s, \pi) v_{\pi}(s').
\end{equation}
In vector form, we have 
\begin{equation}
\bm{v} = \bar{\bm{r}} + \gamma \bm{Pv},\mbox{so } \bm{v} = (\bm{I} - \gamma \bm{P})^{-1}\bar{\bm{r}},
\end{equation}
where $\bar{\bm{r}}$ is the average one-step reward from each state.
Thus, we have, $\bm{v} = \bm{\psi} \bar{\bm{r}}$, 
decomposing the value function into environment dynamics (SR) and the reward signal.
With SR, it is much easier to learn the value function.
SR has wide applications in credit assignment, exploration, transfer learning, planning, imitation learning, and continual learning, etc.
There are some recent papers about successor representation, e.g., 
\citet{Barreto2017}, \citet{Kulkarni2016SR}, \citet{Sherstan2018}, and \citet{Zhang2017SF}.
See \citet{Gershman2018} for a review.

Recently, \citet{Chen2018Bilinear} develop a bilinear representation to capture state-action distributions.

\subsection{Neural Networks}
\label{representation:NN}

In this section, we discuss
representation learning,
neural network architectures,
challenges to CNN and RNN,
memory, and a recently proposed
grid-like representation.
We discuss generative query network (GQN) for scene representation~\citep{Eslami2018GQN} in Chapter~\ref{unsupervised}. 
Deep learning, or deep neural networks, have been playing critical roles in many recent successes in AI. 



\subsection*{Representation Learning}

Representation learning is central to the success of deep learning.
An ideal representation captures underlying disentangled, causal factors of data,
and regularization strategies are necessary for generalization, following no free lunch theorem~\citep{Bengio2013, Goodfellow2016}. 
We list these regularization strategies below.
With \emph{smoothness},  training data generalize to close neighbourhood in input space.
\emph{Linearity} assumes linear relationship, and may be orthogonal to smoothness.
\emph{Multiple explanatory factors} govern data generation,
and motivate distributed representation, with separate directions corresponding to separate factors of variation.
\emph{Causal factors} imply that learned factors are causes of observed data, not vice versa.
\emph{Depth, or a hierarchical organization of explanatory factors} defines high level, abstract concepts with simple, hierarchical concepts.
\emph{Shared factors across tasks} enable sharing of statistical strength between tasks.
\emph{Manifolds} represent the concentration of probability mass with lower dimensionality than original space of data.
\emph{Natural clustering} identifies disconnected manifolds, each may contain a single class.
\emph{Temporal and spatial coherence} assumes that critical explanatory factors change more slowly than raw observations, thus easier to predict.
\emph{Sparsity} assumes that most inputs are described by only a few factors. 
And, \emph{simplicity of factor dependencies} assumes simple dependancies among factors, 
e.g., marginal independence, linear dependency, or those in shallow autoencoders.
Watch a talk \citet{Bengio2018}.
See NIPS 2017 Workshop on Learning Disentangled Representations: from Perception to Control
at \url{https://sites.google.com/view/disentanglenips2017}.

\subsection*{Neural Network Architectures}

A CNN is a feedforward deep neural network, with convolutional layers, pooling layers, and fully connected layers. 
CNNs are designed to process data with multiple arrays, with locality and translation  invariance as inductive bias~\citep{LeCun2015}.

A RNN is built with a recurrent cell, and can be seen as a multilayer neural network with all layers sharing the same weights, with temporal invariance as inductive bias~\citep{LeCun2015}.
Long short term memory networks (LSTM)~\citep{Hochreiter1997} and gated recurrent unit (GRU)~\citep{Chung2014} 
are two popular RNNs, to address issues with gradient computation with long time steps.

\citet{Hinton2006} propose an autoencoder to reduce the dimensionality of data with neural networks.
\citet{Sabour2017} and \citet{Hinton2018} propose capsules with dynamic routing, 
to parse the entire object into a parsing tree of capsules, each of which has a specific meaning. 

There are new neural network architectures customized for RL problems, e.g., 
value iteration networks (VIN)~\citep{Tamar2016},
predictron~\citep{Silver-Predictron-2016},
value prediction network (VPN)~\citep{Oh2017VPN},
imagination-augmented agents (IA2)~\citep{Weber2017},
TreeQN and ATreeC~\citep{Farquhar2018},
temporal difference models (TDMs)~\citep{Pong2018}, 
MCTSnets\citet{Guez2018}, and
BBQ-Networks~\citep{Lipton2018BBQ}.
We discuss RL with models in Chapter~\ref{model}.

\subsection*{Challenges to CNN and RNN}

Some recent papers challenge if RNNs are a natural choice for sequence modelling.
\citet{Bai2018TCN} show empirically that CNNs outperform RNNs over a wide range of tasks.
See the open source at \url{https://github.com/locuslab/TCN}.
\citet{Miller2018} show that feed-forward networks approximate stable RNNs well,
for both learning and inference with gradient descent, and validate theoretical results with experiments. 
\citet{Vaswani2017} propose a self-attention mechanism to replace recurrent and convolutional layers, for sequence transduction problems, like language modelling and machine translation.

\subsection*{Memory}

Memory provides long term data storage.
LSTM is a classical approach for equipping a neural network with memory, and its memory is for both storage and computation.
\citet{Weston2015} propose memory networks to combine inference with a long-term memory.
\citet{Grave-DNC-2016} propose differentiable neural computer (DNC) to solve complex, structured problems. 
\citet{Wayne2018MERLIN} propose memory, RL, and inference network (MERLIN) to deal with partial observability.
We discuss attention and memory including above papers in Chapter~\ref{attention}.
Below we discuss briefly neural networks equipped with memory to facilitate reasoning, 
e.g., relational memory core (RMC)~\citep{Santoro2018RMC}, and, compositional attention networks~\citep{Hutson2018}.

\subsection*{Grid-like Representation}

\citet{Banino2018} study vector-based navigation with grid-like representations.
In a process of vector-based navigation, i.e., planning direct trajectories to goals,
animals travel to a remembered goal, following direct routes by calculating goal-directed vectors with a Euclidean spatial metric provided by grid cells.
Grid cells are also important for integrating self-motion, i.e., path integration.
The authors study path integration with a recurrent network,
and find emergent grid-like representations,
which help improve performance of navigation with deep RL in challenging environments,
and also help with mammal-like shortcut behaviors.
\citet{Cueva2018} is a concurrent work.

CNNs are popular neural networks for image processing, and induce the representation to achieve excellent results. CNNs were inspired from visual cortex. 
A popular representation in NLP is word2vec~\citep{Mikolov2013, Mikolov2017}, which is influenced by linguistics, e.g., quoting John Rupert Firth, "You shall know a word by the company it keeps." 
The grid cell representation, with origin from the brain, boosts performance for navigation tasks.

\subsection{Knowledge and Reasoning}
\label{representation:knowledge}

Knowledge and reasoning~\citep{Brachman2004, Russell2009} are fundamental issues in AI.
It is thus important to investigate issues about them,
e.g., how to represent knowledge, 
like the predictive approach with general value function (GVF), 
or a symbolic approach with entities, properties and relations,
how to incorporate knowledge in the learning system, 
like an inductive bias, in particular, using a knowledge base to improve learning tasks~\citep{Chen2018, Yang2017KB}, 
and how to design network architectures to help with reasoning, etc.

\citet{Bottou2014} discuss machine learning and machine reasoning,
and propose to define reasoning as the manipulation of knowledge previously acquired to answer a new question,
to cover first-order logical inference, probabilistic inference, and components in a machine learning pipeline.
\citet{Evans2018} propose a differentiable inductive logic framework to deal with inductive logic programming (ILP) problems with noisy data.
\citet{Besold2017} discuss neural-symbolic learning and reasoning.
There are books about causality~\citep{Pearl2009, Pearl2016, Pearl2018why, Peters2017}.
\citet{Guo2018Causality} present a survey of learning causality with data.

See NIPS 2018 Workshop on Causal Learning.
See NIPS 2018 Workshop on  Relational Representation Learning at \url{https://r2learning.github.io}.
See NIPS 2017 Workshop on Causal Inference and Machine Learning for Intelligent Decision Making 
at \url{https://sites.google.com/view/causalnips2017}.
See 2015 AAAI Spring Symposium Knowledge Representation and Reasoning: Integrating Symbolic and Neural Approaches at \url{https://sites.google.com/site/krr2015/}.


We discuss general value function, hierarchical RL, and relational RL.
We also discuss very briefly several topics, including 
causality, 
reasoning facilitated by neural networks, 
and incorporating human intelligence.

There are recent papers about neural approaches for algorithm induction, e.g.,
\citet{Balog2017, Grave-DNC-2016, Liang2017, Nachum2017, Reed2016, Vinyals2015, Zaremba2015}.

\subsection*{General Value Function (GVF)}

A key problem in AI is to learn, represent, and use knowledge of the world.
\citet{Sutton2011} discuss that high-level representations based on first-order logic and Bayesian networks are expressive, but it is difficult to learn the knowledge and it is expensive to use such knowledge; and low-level representations like differential equations and state-transition matrices, can be learned from unsupervised data, but such representations are less expressive. 
The authors further discuss that value functions provide semantics for predictive knowledge and goal-oriented (control) knowledge. 

\citet{Sutton2011} propose to represent knowledge with General Value Function (GVF), where policy, termination function, reward function, and terminal reward function are parameters. 
\citet{Schaul2015} propose Universal Value Function Approximators (UVFAs) to generalize over both states and goals.
\citet{Andrychowicz2017} propose Hindsight Experience Replay (HER) to combat with the issue of sparse reward, following the idea of GVF.
We discuss GVF in Section~\ref{value:GVF}.

\subsection*{Hierarchical RL}

Hierarchical RL~\citep{Barto2003} is a way to learn, plan, and represent knowledge with temporal abstraction at multiple levels, with a long history, e.g., 
options~\citep{Sutton1999} and
MAXQ~\citep{Dietterich2000}.
Hierarchical RL explores in the space of high-level goals to address issues of sparse rewards and/or long horizons.
Hierarchical RL may be helpful for transfer and multi-task learning, which we discuss in Section~\ref{meta:transfer}.
Hierarchical planning is a classical topic in AI~\citep{Russell2009}.
There are some recent papers, like,
hierarchical-DQN~\citep{Kulkarni2016},
strategic attentive writer~\citep{Vezhnevets2016}, 
feudal network~\citep{Vezhnevets2017}, 
option-critic~\citep{Bacon2017}, 
option discovery with a Laplacian framework~\citep{Machado2017}, and,
stochastic neural networks~\citep{Florensa2017}.
We discuss hierarchical RL in Chapter~\ref{hierarchical}.

\subsection*{Relational RL}

Statistical relational learning and reasoning studies uncertain relations, 
and manipulates structured representations of entities and their relations, 
with rules about how to compose them~\citep{Battaglia2018GN, Getoor2007}.
Inductive logic programming (ILP) learns uncertain logic rules from positive and negative examples,
entailing positive examples but not negative ones. 
Probabilistic ILP~\citep{DeRaedt2008, Manhaeve2018NIPS} is closely related to statistical relational learning.
Probabilistic ILP integrates rule-based learning with statistical learning, 
and tackles the high complexity of ILP. 
Graphical models~\citep{Koller2009} are important approaches for statistical relational learning.

Artificial neural networks have alternative names, 
including connectionism, parallel distributed processing, and neural computation~\citep{Russell2009}.
Symbolism is about a  formal language with symbols and rules, defined by mathematics and logic.
Relational learning and reasoning with neural networks is an approach integrating connectionism and symbolism.

Relational RL integrates RL with statistical relational learning, and connects RL with classical AI, 
for knowledge representation and reasoning.
Relational RL is not new.
\citet{Dzeroski2001} propose relational RL.
\citet{Tadepalli2004} survey relational RL.
\citet{Guestrin2003} introduce relational MDPs.
\citet{Diuk2008} introduce objected-oriented MDPs (OO-MDPs).
Recently,
\citet{Battaglia2018GN} propose graph network (GN) to incorporate relational inductive bias,
\citet{Zambaldi2018} propose deep relational RL,
\citet{Keramati2018} propose strategic object oriented RL,
and there are also deep learning approaches to deal with relations and/or reasoning, 
e.g., \citet{Battaglia2016}, \citet{Chen2018}, \citet{Hutson2018}, \citet{Santoro2017}, and \citet{Santoro2018RMC}.  
We discuss relational RL in Chapter~\ref{relational}.

\subsection*{Causality}

\citet{Pearl2018} discusses that there are three fundamental obstacles for current machine learning systems to exhibit human-level intelligence: adaptability or robustness, explainability, and understanding of cause-effect connections.
The author describes a three layer causal hierarchy: association, intervention, and counterfactual.
Association invokes statistical relationships, with typical questions like "What is?" and "How would seeing $X$ change my belief in $Y$".
Intervention considers not only seeing what is, but also changing what we see, with typical questions like "What if?" and "What if I do $X$?".
Counterfactual requires imagination and retrospection, with typical questions like "Why?" and "What if I had acted differently?".
Counterfactuals subsume interventional and associational questions, and  interventional questions subsume associational questions.

\citet{Pearl2018} proposes structural causal model, which can accomplish seven pillar tasks in automated reasoning:
1) encoding causal assumptions - transparency and testability,  
2) do-calculus and the control of counfounding,
3) the algorithmization of counterfactuals,
4) mediation analysis and the assessment of direct and indirect effects,
5) adaptability, external validity and sample selection bias,
6) missing data, and,
7) causal discovery.

See some recent papers using deep learning to treat causality, e.g.,
\citet{Johansson2016}, \citet{Hartford2017}, and \citet{LopezPaz2017}.
\citet{Lattimore2016Causal} discuss causal bandits.
\citet{Tamar2018NIPS} discuss learning plannable representations with causal InfoGAN.
\citet{LiuYao2018NIPS} study off-policy policy evaluation inspired by causal reasoning.


\subsection*{Reasoning}

\citet{Battaglia2018GN} propose graph network (GN) to incorporate relational inductive bias, to attempt to achieve combinatorial generalization.
GN generalizes graph neural network (GNN), e.g., \citet{Scarselli2009}.
\citet{Santoro2017} propose relation networks (RNs) for relational reasoning.
\citet{Santoro2018RMC} propose a relational memory core (RMC) with self-attention to handle tasks with relational reasoning.
\citet{Hudson2018} propose memory, attention, and control (MAC) recurrent cell for reasoning.
\citet{YiKexin2018NIPS} discuss disentangling reasoning from vision and language understanding.
We discuss relational RL in Chapter~\ref{relational}.
 
\subsection*{Human Intelligence} 
 
\citet{Lake2016} discuss that we should build machines towards human-like learning and thinking. 
In particular, we should build causal world models, to support understanding and explanation, seeing entities rather than just raw inputs or features, rather than just pattern recognition;
we should support and enrich the learned knowledge grounding in intuitive physics and intuitive psychology;
we should represent, acquire, and generalize knowledge, leveraging compositionality and learning to learn, rapidly adapt to new tasks and scenarios, recombining representations, without retraining from scratch.  

\citet{Lake2016} discuss that the following are key ingredients to achieve human-like learning and thinking: 
a) developmental start-up software, or cognitive capabilities in early development, including,
a.1) intuitive physics, and,   
a.2) intuitive psychology;
b) learning as rapid model building, including, 
b.1) compositionality,
b.2) causality, and, 
b.3) learning to learn;
c) thinking fast, including, 
c.1) approximate inference in structured models, and,
c.2) model-based and model-free reinforcement learning.
Watch a video~\citet{Tenenbaum2018ICML}.

We explain some of these key gradients by quoting directly from \citet{Lake2016}.
Intuitive physics refers to that "Infants have primitive object concepts that allow them to track objects over time and to discount physically implausible trajectories".
Intuitive psychology refers to that "Infants understand that other people have mental states like goals and beliefs, and this understanding strongly constrains their learning and predictions".
For causality: "In concept learning and scene understanding, causal models represent hypothetical real-world processes that produce the perceptual observations. In control and reinforcement learning, causal models represent the structure of the environment, such as modeling state-to-state transitions or action/state-to-state transitions."
 
\citet{Botvinick2017} discuss about one additional ingredient, autonomy, 
so that agents can build and exploit their own internal models, with minimal human manual engineering.
 
\clearpage


\newpage

\section*{Part II: Important Mechanisms}

\addcontentsline{toc}{section}{Part II: Important Mechanisms}

In this part, we discuss important mechanisms for the development of (deep) reinforcement learning, 
including 
attention and memory  in Chapter~\ref{attention}, 
unsupervised learning in Chapter~\ref{unsupervised},
hierarchical RL in Chapter~\ref{hierarchical}, 
relational RL in Chapter~\ref{relational},
multi-agent RL in Chapter~\ref{MARL}, and, 
learning to learn in Chapter~\ref{meta}.

Note that we do not discuss some mechanisms, like Bayesian RL~\citep{Ghavamzadeh2015}, 
and semi-supervised RL~\citep{Audiffren2015, Cheng2016, Dai2017, Finn2017, Kingma2014,Papernot2017,YangZ2017, Zhu2009}.




\clearpage

\newpage              


\section{Attention and Memory}
\label{attention}

Attention is a mechanism to focus on the salient parts. 
Memory provides long term data storage.
Attention can be an approach for memory addressing. 

A soft attention mechanism, e.g., \citet{Bahdanau2015}, 
utilizes a weighted addressing scheme to all memory locations, 
or a distribution over memory locations, can be trained with backpropagation.
A hard attention mechanism, e.g., \citet{Zaremba2015}, 
utilizes a pointer to address a memory location, 
following the way conventional computers accessing memory, 
and can be trained with reinforcement learning, in particular, policy gradient.
Attention can help with visualization about where a model is attending to, 
e.g., in machine translation and image captioning.
Most papers follow a soft attention mechanism. 
There are endeavours for hard attention~\citep{ Liang2017, Malinowski2018, Xu2015, Zaremba2015}.


See \citet{olah2016attention} and \citet{Britz2016} for discussions about attention and memory;
the former discusses neural Turing machine~\citep{Graves2014} etc.,
and the latter discusses sequence-to-sequence model~\citep{Bahdanau2015}, etc.

In the following, we discuss several papers about attention and/or memory.

See also \citet{Ba2014, Ba2016,  Danihelka2016, Duan2017OneShot, Eslami2016, Gregor2015, Jaderberg2015, Kaiser2016, Kadlec2016,  Oh2016, Oquab2015, Yang2016CVPR, Zagoruyko2017, Zaremba2015}. 


\subsection*{Attention}


\citet{Cho2014} and \citet{Sutskever2014} propose the sequence to sequence approach by using two RNNs to encode a sentence to a fix-length vector and then decode the vector into a target sentence.  
To address the issues with encoding the whole sentence into a fix-length vector in the basic sequence to sequence approach, 
\citet{Bahdanau2015} introduce a soft-attention technique,
i.e., weighted sum of annotations to which an encoder maps the source sentence,
to learn to jointly align and translate,
by soft-searching for most relevant parts of the source sentence,
and predicting a target word with these parts and previously generated target words.

\citet{Mnih-attention-2014} introduce the recurrent attention model (RAM) to focus on selected sequence of regions or locations from an image or video for image classification and object detection,
to reduce computational cost for handling large video or images. 
The authors utilize REINFORCE to train the model, to overcome the issue that the model is non-differentiable, and experiment on an image classification task and a dynamic visual control problem. 


\citet{Xu2015} integrate attention to image captioning, 
inspired by the papers in neural machine translation~\citep{Bahdanau2015} and object recognition~\citep{Mnih-attention-2014, Ba2014}.
The authors utilize a CNN to encode the image, and an LSTM with attention to generate a caption. 
The authors propose a soft deterministic attention mechanism and a hard stochastic attention mechanism.
The authors show the effectiveness of attention with caption generation tasks on Flickr8k, Flickr30k, and MS COCO datasets.


\citet{Vaswani2017} propose Transformer, using self-attention to replace recurrent and convolutional layers, for sequence transduction problems, like language modelling and machine translation. 
Transformer utilizes a scaled dot-product attention, to map a query and key-value pairs to an output,
and computes a set of queries as matrices simultaneously to improve efficiency.
Transformer further implements a multi-head attention 
by transforming queries, keys, and values  with different, learned linear projections respectively,
performing the attention function in parallel, then concatenating results and yielding final values.
Transformer follows the encoder-decoder architecture.
The encoder is composed of a stack of six identical layers,
with two sub-layers, a multi-head self-attention mechanism, then, 
a position-wise fully connected feed-forward network,
with residual connection around each sub-layer, followed by layer normalization.  
The decoder is the same as the encoder, 
with an additional multi-head attention sub-layer between the two sub-layers,
which takes inputs from the output of the encoder stack and the output from previous multi-head attention sub-layer.
Transformer implements positional encoding to account for the order of the sequence.
The authors evaluate Transformer on two machine translation tasks, achieve good results w.r.t. BLEU score, and show that an attention mechanism has better time efficiency and is more parallelizable than recurrent models. 
\citet{Dehghani2018} extend Transformer.
See open source at \url{https://github.com/tensorflow/tensor2tensor}, in particular, for \citet{Dehghani2018} at \url{https://goo.gl/72gvdq}.
\citet{Tang2018self} study the hypothesis that self-attention and CNNs, rather than RNNs, can extract semantic features to improve long range dependences in texts with NLP tasks.

\subsection*{Memory}


\citet{Weston2015} propose memory networks to combine inference with a long-term memory, 
which could be read from and written to,
and train these two components to use them jointly. 
The authors present a specific implementation of the general framework on the task of question answering (QA),
where the memory works as a dynamic knowledge base,
and evaluate on a large-scale QA task and a smaller yet complex one.

\citet{Sukhbaatar2015} extend \citet{Weston2015} with a recurrent attention model over a large external memory, 
train in an end-to-end way, and experiment with question answering and language modelling tasks. 
See open source at \url{https://github.com/facebook/MemNN}. 


\citet{Grave-DNC-2016} propose differentiable neural computer (DNC), in which, a neural network can read from and write to an external memory, so that DNC can solve complex, structured problems, which a neural network without read-write memory can not solve. DNC minimizes memory allocation interference and enables long-term storage.  Similar to a conventional computer, in a DNC, the neural network is the controller and the external memory is the random-access memory, and a DNC represents and manipulates complex data structures with the memory. Differently, a DNC learns such representation and manipulation end-to-end with gradient descent from data in a goal-directed manner. When trained with supervised learning, a DNC can solve synthetic question answering problems, for reasoning and inference in natural language. Moreover, it can solve the shortest path finding problem between two stops in transportation networks and the relationship inference problem in a family tree. When trained with reinforcement learning, a DNC can solve a moving blocks puzzle with changing goals specified by symbol sequences. DNC outperforms normal neural network like LSTM or DNC's precursor neural Turing machine~\citep{Graves2014}. With harder problems, an LSTM may simply fail. Although these experiments are relatively small-scale, we expect to see further improvements and applications of DNC. 
See a blog at \url{https://deepmind.com/blog/differentiable-neural-computers/}. 


\citet{Wayne2018MERLIN} propose memory, RL, and inference network (MERLIN) to deal with partial observability,
by equipping with extensive memory, and more importantly, formatting memory in the right way for storing right information trained with unsupervised predictive modelling. 
The author evaluate MERLIN on behavioural tasks in psychology and neurobiology,
which may have high dimension sensory input and long duration of experiences.

\clearpage

\section{Unsupervised Learning}
\label{unsupervised}

Unsupervised learning takes advantage of the massive amount of data without labels, and would be a critical mechanism to achieve artificial general intelligence.  

Unsupervised learning is categorized into non-probabilistic models, like sparse coding, autoencoders,  k-means etc, and probabilistic (generative) models, where density functions are concerned, either explicitly or implicitly. Among probabilistic (generative) models with explicit density functions, some are with tractable models, like fully observable belief nets, neural autoregressive distribution estimators, and PixelRNN, etc; some are with non-tractable models, like Botlzmann machines, variational autoencoders,  Helmhotz machines, etc. For probabilistic (generative) models with implicit density functions, we have generative adversarial networks (GANs), moment matching networks, etc. 
See \citet{Salakhutdinov2016} for more details.

\citet{LeCun2018} summarizes the development of deep learning, and outlooks the future of AI, 
highlighting the role of world models and self-supervised learning.
\footnote{\citet{LeCun2018} uses the cake metaphor, as in his NIPS 2016 invited talk titled Predictive Learning.
In this metaphor, "pure" reinforcement learning, as the single cherry on the cake, "predicts a scalar reward given once in a while", with very low feedback information content; supervised learning,  as the icing of the cake, "predicts a category or a few numbers for each input", with medium feedback information content; and, self-supervised learning, as cake genoise, "predicts any part of its input for any observed part", or "predicts future frames in videos", with high but stochastic feedback information content ("self-supervised learning" replacing "unsupervised/predictive learning" in his NIPS 2016 talk).  
As one response from the RL community, 
Pieter Abbeel presents a cake with many cherries in \citet{Abbeel2017NIPStalk}, 
as  a metaphor that RL methods can also have high information content,
e.g., Hindsight Experience Replay (HER)~\citep{Andrychowicz2017} and Universal Value Function Approximators (UVFAs)~\citep{Schaul2015}. }



Self-supervised learning is a special type of unsupervised learning,
in which, no labels are given; however, labels are created from the data.
Unsupervised auxiliary learning~\citep{Jaderberg2017, Mirowski2017}, 
GANs, and \citet{Aytar2018}, as we discuss below, can be regarded as self-supervised learning.
\citet{Pathak2017} propose curiosity-driven exploration by self-supervised prediction.
Watch two talks, \citet{Efros2017} and \citet{Gupta2017talk}, about self-supervised learning.

\citet{Goel2018NIPS} conduct unsupervised video object segmentation for deep RL.
\citet{Mirowski2018NIPS} study learning to navigate in cities without a map.
\citet{Hsu2018} study unsupervised learning with meta-learning.
See also \citet{Artetxe2018}, \citet{Le2012}, \citet{Liu2017unsupervised}, 
\citet{Nair2018NIPS},
and \citet{vandenOord2018}.

In the following, we discuss unsupervised auxiliary learning~\citep{Jaderberg2017, Mirowski2017}, which, together with Horde~\citep{Sutton2011}, are approaches to take advantages of possible non-reward training signals in environments. 
We also discuss generative adversarial networks~\citep{Goodfellow2014},
generative query network (GQN) for scene representation~\citep{Eslami2018GQN}, and,
playing hard exploration games by watching YouTube~\citep{Aytar2018}.


\subsection*{Unsupervised Auxiliary Learning}

Environments may contain abundant possible training signals, which may help to expedite achieving the main goal of maximizing the accumulative rewards, e.g., pixel changes may imply important events, and auxiliary reward tasks may help to achieve a good representation of rewarding states. This may be even helpful when the extrinsic rewards are rarely observed.


\citet{Jaderberg2017} propose unsupervised reinforcement and auxiliary learning (UNREAL) to improve learning efficiency by maximizing pseudo-reward functions, besides the usual cumulative reward, while sharing a common representation.  UNREAL is composed of four components: base agent, pixel control, reward prediction, and value function replay. 
The base agent is a CNN-LSTM agent, and is trained on-policy with A3C~\citep{Mnih-A3C-2016}. 
Experiences of observations, rewards and actions are stored in a replay buffer, for being used by auxiliary tasks. 
The auxiliary policies use the base CNN and LSTM, together with a deconvolutional network, to maximize changes in pixel intensity of different regions of the input images. 
The reward prediction module predicts short-term extrinsic reward in next frame by observing the last three frames, to tackle the issue of reward sparsity. 
Value function replay further trains the value function.  
UNREAL has a shared representation among signals, while Horde trains each value function separately with distinct weights. 
The authors show that UNREAL improves A3C's performance on Atari games, and performs well on 3D Labyrinth game. 
See a blog at \url{https://deepmind.com/blog/reinforcement-learning-unsupervised-auxiliary-tasks/}.

\citet{Mirowski2017} obtain navigation ability by solving a RL problem maximizing cumulative reward and  jointly considering unsupervised tasks to improve data efficiency and task performance.  The authors address the sparse reward issues by augmenting the loss with two auxiliary tasks, 1) unsupervised reconstruction of a low-dimensional depth map for representation learning to aid obstacle avoidance and short-term trajectory planning; 2) self-supervised loop closure classification task within a local trajectory. The authors incorporate a stacked LSTM to use memory at different time scales for dynamic elements in the environments. 
The proposed agent learns to navigate in complex 3D mazes end-to-end from raw sensory inputs, and performs similarly to human level, even when start/goal locations change frequently. In this approach, navigation is a by-product of the goal-directed RL optimization problem, in contrast to conventional approaches such as simultaneous localization and mapping (SLAM), where explicit position inference and mapping are used for navigation. 

\subsection*{Generative Adversarial Nets}
\label{unsupervised:GAN}

\citet{Goodfellow2014} propose generative adversarial nets (GANs) to estimate generative models via an adversarial process by training two models simultaneously, a generative model $G$ to capture the data distribution, and a discriminative model $D$ to estimate the probability that a sample comes from the training data but not the generative model $G$. 

\citet{Goodfellow2014} model $G$ and $D$ with multilayer perceptrons: $G(z: \bm{\theta}_g)$ and $D(x: \bm{\theta}_d)$, where $\bm{\theta}_g$ and $\bm{\theta}_d$ are parameters,  $x$ are data points, and $z$ are input noise variables. Define a prior on input noise variable $p_z(z)$. $G$ is a differentiable function and $D(x)$ outputs a scalar as the probability that $x$ comes from the training data rather than $p_g$, the generative distribution we want to learn.   

$D$ will be trained to maximize the probability of assigning labels correctly to samples from both training data and $G$. Simultaneously, $G$ will be trained to minimize such classification accuracy, $\log(1-D(G(z)))$. As a result, $D$ and $G$ form the two-player minimax game as follows:
\begin{equation}
\min_{G} \max_{D} E_{x \sim p_{data}(x)} [\log D(x)]  + E_{z \sim p_{z}(z)} [\log (1- D(G(z)))].
\end{equation}
\citet{Goodfellow2014} show that as $G$ and $D$ are given enough capacity, generative adversarial nets can recover the data generating distribution, and provide a training algorithm with backpropagation by minibatch stochastic gradient descent. 

GANs are notoriously hard to train. See \citet{Arjovsky2017WGAN} for Wasserstein GAN (WGAN) as a stable GANs model. \citet{Gulrajani2017} propose to improve stability of WGAN by penalizing the norm of the gradient of the discriminator w.r.t. its input,  instead of clipping weights as in \citet{Arjovsky2017WGAN}.  \citet{Mao2016} propose Least Squares GANs (LSGANs), another stable model. \citet{Berthelot2017} propose BEGAN to improve WGAN by an equilibrium enforcing model, and set a new milestone in visual quality for image generation. \citet{Bellemare2017Cramer} propose Cram{\'e}r GAN  to satisfy three machine learning properties of probability divergences: sum invariance, scale sensitivity, and unbiased sample gradients. \citet{Hu2017} unified GANs and Variational Autoencoders (VAEs). 


\citet{Lucic2018NIPS} conduct a large-scale empirical study on GANs models and evaluation measures,
and observe that, by fine-tuning  hyperparameters and random restarts,
most models perform similarly. 
The authors propose more data sets which enable computing of precision and recall.
The authors further observe that the evaluated models do not outperform the original GAN algorithm. 
\citet{Kurach2018} discuss the state of the art of GANs in a practical perspective. 
The authors reproduce representative algorithms, discuss common issues, open-source their code on Github, and provide pre-trained models on TensorFlow Hub.
\citet{Brock2019} study image synthesis.

We discuss generative adversarial imitation learning~\citep{Ho2016, Li2017InfoGAIL} in Chapter~\ref{reward}.
\citet{Finn2016-Connection} establish connections between GANs, inverse RL, and energy-based models. 
\citet{Pfau2016} establish connections between GANs and actor-critic algorithms. 

See \citet{Goodfellow2017} for summary of his NIPS 2016 Tutorial on GANs. GANs have received much attention and many work have been appearing after the publication of \citet{Goodfellow2017}.
See CVPR 2018 tutorial on GANs at \url{https://sites.google.com/view/cvpr2018tutorialongans/},
with several talks by several speakers.

\subsection*{Generative Query Network}
\label{unsupervised:GQN}

\citet{Eslami2018GQN} propose generative query network (GQN) for scene representation,
to obtain a concise description of a 3D scene from 2D visual sensory data, 
with a representation network and a generator network trained jointly in an end-to-end fashion,
without human semantic labels, e.g., object classes, object locations, scene types, or part labels,  and domain knowledge.

In GQN, observations from different 2D viewpoints for 3D scenes are fed into the representation network, 
to obtain a neural scene representation by summing observations' representations element-wise.
The neural scene representation is then fed into the generation network, 
 which is a recurrent latent variable model, 
to make prediction about the scene from a different viewpoint.
GQN is trained with many scenes, with various number of observations, and with back-propagation.
GQN attempts to obtain a concise scene representation, 
to capture the scene contents, e.g., the identities, positions, colors, and object counts, etc.,
and to make a generator successful for predictions,
by maximizing the likelihood to generate ground-truth images from query viewpoints.
Variational approximations are used to deal with the intractability of latent variables.

Experiments show that representations learned by GQN with the properties of viewpoint invariance, compositionality, factorization, "scene algebra", similar to that of word embedding algebra, and decreasing Bayesian surprise  with more observations for both full and partial observability.
Bayesian surprise refers to the KL divergence between conditional prior and posterior.
Robot arm reaching experiments show that the GQN representation helps with data efficiency and robust control.

\subsection*{Playing Games by Watching YouTube}

\citet{Aytar2018} propose to play hard exploration Atari games, including Montezuma's Revenge, Pitfall! and Private Eye, by watching YouTube. YouTube videos are usually noisy and unaligned, without the frame-by-frame alignment between demonstrations, and the information of exact action and reward sequences in demonstrator's observation trajectory, 
which are the properties of demonstrations required by previous imitation learning, e.g., \citet{Hester2018} as we discuss in Chapter~\ref{reward}. 

\citet{Aytar2018} overcome these limitations by a one-shot imitation method in two steps. 
First, a common representation is learned from unaligned videos from multiple sources, with two self-supervised objectives: temporal distance classification (TDC) and cross-model temporal distance classification (CMC). 
In self-supervision, an auxiliary task is proposed to solve among all domains simultaneously, for a network to attempt to learn a common representation.
In TDC, temporal distances between two frames in a single video sequence are predicted, to help learn a representation of the environment dynamics.
In CMC, a representation is learned to correlate visual and audio observations, and to highlight critical game events.  
Furthermore, a new measure of cycle-consistency is proposed to evaluate the quality of the learned representation. 
Second,  a single YouTube video is embedded in such representation, and a reward function is built, so that an agent can learn to imitate human game play. 

Experiments using the distributed A3C RL algorithm IMPALA~\citep{Espeholt2018} show breakthrough results on these three hard Atari games.

\clearpage

\section{Hierarchical RL}
\label{hierarchical}

Hierarchical RL~\citep{Barto2003} is a way to learn, plan, and represent knowledge with temporal abstraction at multiple levels, with a long history, e.g., 
options~\citep{Sutton1999}, 
MAXQ~\citep{Dietterich2000},
hierarchical abstract machines~\citep{Parr1998}, and
dynamic movement primitives~\citep{Schaal2006}.
Hierarchical RL is an approach for issues of sparse rewards and/or long horizons,
with exploration in the space of high-level goals.
The modular structure of hierarchical RL approaches is usually conducive to transfer and multi-task learning, which we discussed in Section~\ref{meta:transfer}.
The concepts of sub-goal, option, skill, and, macro-action are related.
Hierarchical planning is a classical topic in AI~\citep{Russell2009}.


Here we introduce options briefly.
Markov decision process (MDP) is defined by the 5-tuple $(\mathcal{S}, \mathcal{A}, \mathcal{P},\mathcal{R}, \gamma)$,
for the state space, the action space, the state transition probability, the reward function, and the discount factor.
An option $o$ consists of three components:
1) an initiation set of states $\mathcal{I}_o \subseteq \mathcal{S}$,
2) a policy $\pi_o: \mathcal{S} \times\mathcal{A} \rightarrow [0,1]$, guiding the behaviour of an option, 
such that $\pi_o(a|s)$ is the probability of taking action $a$ in state $s$ when following option $o$,
and, 3) a termination condition $\beta_o: \mathcal{S} \rightarrow [0,1]$, roughly determining the length of an option, 
such that $\beta_o(s)$ is the probability of terminating the option $o$ upon entering state $s$.
A policy-over-options calls an option $o$. During the execution of the option $o$, the agent selects an action until a termination condition is met. An  option may call another option. 
$P(s'|s, o)$ is the probability of next state $s'$ conditioned on that option $o$ executes from state $s$.
$r_o(s)$ is the expected return  during the execution of option $o$.
Introducing options over an MDP constitutes a semi-Markov decision process (SMDP).
It can be shown that learning and planning algorithms from MDPs can transfer to options.
It can be shown that the reward model for options is equivalent to a value function,
and a Bellman equation can be written for it, so RL algorithms can be used to learn it.
This also applies to the transition model for options.

Usually options are learned with provided sub-goals and pseudo-rewards,
and good performance is shown for Atari games, e.g. with hierarchical-DQN (h-DQN)~\citep{Kulkarni2016},
and for MineCraft, e.g., with \citet{Tessler2017}.
Automatic options discovery receives much attention recently, e.g.,
strategic attentive writer (STRAW)~\citep{Vezhnevets2016}, 
feudal network (FuN)~\citep{Vezhnevets2017}, 
option-critic~\citep{Bacon2017}, 
option discovery with a Laplacian framework~\citep{Machado2017}, and,
stochastic neural networks~\citep{Florensa2017}.

Hierarchical RL follows the general algorithm design principle of divide and conquer,
so that hard goals, e.g. those with sparse long-term rewards are replaced with easy sub-goals, 
e.g. those with dense short-term rewards, and RL algorithms, 
e.g., policy-based or value-based, combined with representations, are utilized to solve easy sub-goals,
and finally to achieve hard goals. 



Watch recent talks on hierarchical RL, e.g., \citet{Silver2017}, \citet{Precup2018}.
See NIPS 2017 workshop on Hierarchical RL,
at \url{https://sites.google.com/view/hrlnips2017}, and videos at \url{https://goo.gl/h9Mz1a}.

We discuss several recent papers in the following.
See also \citet{Kompella2017}, 
\citet{LeHoang2018ICML}, 
\citet{Nachum2018NIPS}, 
\citet{Peng2017Dialogue},  \citet{Sharma2017}, \citet{Tang2018EMNLP}, \citet{Tessler2017}, and \citet{Yao2014}.

\subsection*{Hierarchical DQN}

\citet{Kulkarni2016} propose hierarchical-DQN (h-DQN) by organizing goal-driven intrinsically motivated deep RL modules hierarchically to work at different time-scales.  
h-DQN integrats a top level action value function and a lower level action value function.
The former learns a policy over intrinsic sub-goals, or options~\citep{Sutton1999}.
And the latter learns policies over raw actions to satisfy the objective of each given sub-goal. 
In particular, h-DQN has a two-stage hierarchy with a meta-controller and a controller.
The meta-controller receives state $s$ and select a goal $g$.
The controller then selects an action $a$ conditioned on state $s$ and goal $g$,
the goal $g$ does not change until it is achieved or a termination condition is met.
The internal critic evaluates if a goal has been achieved,
and produces a reward $r(g)$ to the controller,
e.g., a binary internal reward 1 for achieving the goal, and 0 otherwise.
The objectives for the meta-controller and controller are to maximize cumulative extrinsic and intrinsic rewards, respectively.
The authors evaluate h-DQN on a discrete stochastic decision process, and a hard Atari game, Montezuma's Revenge.
See the open source at \url{https://github.com/mrkulk/hierarchical-deep-RL}

\subsection*{Feudal Networks}

\citet{Vezhnevets2017} propose feudal networks (FuNs) for hierarchical RL, inspired by feudal RL~\citep{Dayan1993Feudal}.
In FuNs, a Manager module discovers and sets abstract sub-goals and operates at a long time scale,
and a Worker module selects atom actions at every time step of the environment
to achieve the sub-goal set by the Manager.
FuNs decouple end-to-end learning across multiple levels at different time scales,
by separating the Manager module from the Worker module, 
to facilitate long time scale credit assignment and emergence of sub-policies for sub-goals set by the Manager.
In FuNs, sub-goals are formulated as directions in the latent space,
so that the Manager selects a subgoal direction to maximize reward, 
and the Worker selects actions to maximize cosine similarity to the direction of the subgoal set by the Manager.
The authors utilize a dilated LSTM to design the Manager to allow backpropagation through long time steps,
and experiment FuNs with a water maze and Atari games.

\subsection*{Option-Critic}

\citet{Bacon2017} derive policy gradient theorems for options, 
and propose an option-critic architecture to learn both intra-option policies and termination conditions gradually, 
at the same time with the policy-over-options, 
combining options discovery with options learning.
The option-critic architecture works with linear and non-linear function approximations, 
with discrete or continuous state and action spaces,
and without rewards or sub-goals.
The authors experiment with a four-room domain, a pinball domain, and Atari games.
See the open source at \url{https://github.com/jeanharb/option_critic}.

\citet{Harutyunyan2018} study the dilemma between efficiency of long options and flexibility of short ones 
in the option-critic architecture.
The authors decouple the behaviour and target termination conditions, 
similar to off-policy learning for policies,
and propose to cast options learning as multi-step off-policy learning,
The authors show benefits of learning short options from longer ones,
by analysis and with experiments.

\citet{Riemer2018NIPS} further study the option-critic architecture.

\subsection*{Option Discovery with A Laplacian Framework}

\citet{Machado2017} propose to discover options with proto-value functions (PVFs)~\citep{Mahadevan2007}, 
which are well-known for representation in MDPs and define options implicitly.
The authors introduce the concepts of eigen-purpose and eigen-behavior.
An eigen-purpose is an intrinsic reward function, 
to motivate an agent to traverse in principle directions of the learned representation of state space.
An eigen-behavior is the optimal policy for an intrinsic reward function.
The authors discover task-independent options, 
since the eigen-purposes  are obtained without reward information.
The authors observe that some options are not helpful for exploration,
although they improve the efficiency of planning.
The authors further show that the options they discover improve exploration, 
since these options operate at different time scales, and they can be sequenced easily.
The authors experiment with tabular domains and Atari games.

\subsection*{Strategic Attentive Writer}

\citet{Vezhnevets2016} propose strategic attentive writer (STRAW), a deep recurrent neural network architecture, for learning high-level temporally abstract macro-actions in an end-to-end manner based on observations from the environment. Macro-actions are sequences of actions commonly occurring.  STRAW builds a multi-step action plan, updated periodically based on observing rewards, and learns for how long to commit to the plan by following it without replanning. STRAW learns to discover macro-actions automatically from data, in contrast to the manual approach in previous work. \citet{Vezhnevets2016} validate STRAW on next character prediction in text, 2D maze navigation, and Atari games.

\subsection*{Stochastic Neural Networks}

\citet{Florensa2017} propose to pre-train a large span of skills using stochastic neural networks with an information-theoretic regularizer, then on top of these skills, to train high-level policies for downstream tasks.  Pre-training is based on a proxy reward signal, which is a form of intrinsic motivation to explore agent's own capabilities, requiring minimal domain knowledge about the downstream tasks.  
Their method combines hierarchical methods with intrinsic motivation, 
and the pre-training follows an unsupervised way.

\clearpage 
\section{Multi-Agent RL}
\label{MARL}

Multi-agent RL (MARL) is the integration of multi-agent systems~\citep{Horling2004,Shoham2009, Stone2000} with RL. 
It is at the intersection of game theory~\citep{Leyton-Brown2008} and RL/AI.  

Besides issues in RL like sparse rewards and sample efficiency, there are new issues like multiple equilibria,\footnote{\citet{Chen2006} show that finding a Nash equilibrium in a two-player game is PPAD-complete, i.e., unless every problem in PPAD is solvable in polynomial time, there is not a fully polynomial-time approximation algorithm for finding a Nash equilibrium in a two-player game. PPAD is a complexity class for polynomial parity arguments on directed graphs.} 
and even fundamental issues like what is the question for multi-agent learning, and whether convergence to an equilibrium is an appropriate goal, etc. 
Consequently, multi-agent learning is challenging both technically and conceptually, and demands clear understanding of the problem to be solved, the criteria for evaluation, and coherent research agendas~\citep{Shoham2007}.

In a fully centralized approach, 
when global state and joint action information are available,
estimating the joint Q action value function becomes possible.
This can address the nonstationary issue.
However, it encounters the issue of curse of dimensionality when the number of agents grows.
Another issue is that it may be hard to extract decentralized policies,
for an agent to make decisions based on its own observation.

\citet{Littman1994}  employ stochastic games as a framework for MARL,
propose minimax-Q learning for zero-sum games, and show convergence under certain conditions. 
\citet{Hu2003} propose Nash Q-learning for general-sum games and show its convergence with a unique Nash equilibrium.
\citet{Bowling2002} propose the win or learn fast (WoLF) principle to vary the learning rate to tackle the issues with learning a moving target, and show convergence with self-play in certain iterated matrix games.
These papers, together with \citet{Foerster2018COMA}, \citet{Lowe2017}, and \citet{Usunier2016}, 
follow centralized approaches. 

\citet{Tan1993} introduces independent Q-learning (IQL) for MARL,
where each agent learns a Q action value function independently.
For Q-learning to be stable and convergent, the environment would be stationary. 
This is usually not the case for multi-agent systems,
where an agent would change its policy according to other agents,
and the environment is usually nonstationary or even adversarial.   
Independent approaches to MARL may not converge.
\citet{Foerster2017} and \citet{Omidshafiei2017} propose to stabilize independent approaches.

\citet{Oliehoek2008} introduce the paradigm of centralized training for decentralized execution.
We discuss several papers following this scheme below.

Along with the success of RL in single agent problems, like, 
\citet{Mnih-DQN-2015}, 
\citet{Jaderberg2017},
\citet{Schulman2015},
\citet{Nachum2018TrustPCL},
and two-player games, like  \citet{Silver-AlphaGo-2016, Silver-AlphaGo-2017, Moravcik2017}, recently, we see some progress in multi-agent RL problems, like \citet{Jaderberg2018Quake} for Quake III Arena Capture the Flag, 
\citet{SunPeng2018StarCraft} and \citet{Pang2018StarCraft} for StarCraft II,
and OpenAI Five for Dota 2. 

\citet{Zambaldi2018} investigate StarCraft II mini-games  with relational deep RL, as discussed in Chapter~\ref{relational}.
\citet{Bansal2018} investigate the emergent complex skills via multi-agent competition.
\citet{Foerster2018opponent} propose learning with opponent-learning awareness, 
so that each agent considers the learning process of other agents in the environment.
\citet{Hoshen2017} present vertex attention interaction network (VAIN), for multi-agent predictive modelling, with an attentional neural network.
\citet{MahdiElMhamdi2017} introduce dynamic safe interruptibility for MARL, in joint action learners and independent learners scenarios.
\citet{Perolat2017} propose to use MARL for the common pool resource appropriation problem.
\citet{Hu2018IJCAI} propose opponent-guided tactic learning for StarCraft micromanagement.
\citet{SongJiaming2018NIPS} extend generative adversarial imitation learning (GAIL)~\citep{Ho2016} to multi-agent settings.
\citet{Lanctot2018NIPS} study actor-critic policy optimization in partially observable multi-agent settings.
See also 
\citet{Hughes2018NIPS},
\citet{WaiHoiTo2018NIPS}, and
\citet{ZhouZhengyuan2018NIPS}

Multi-agent systems have many applications, e.g., as we will discuss, games in Chapter~\ref{games}, robotics in Chapter~\ref{robotics}, energy in Section~\ref{energy}, transportation in Section~\ref{transportation}, and compute systems in Section~\ref{systems}.

\citet{Busoniu2008} and \citet{Ghavamzadeh2006} are surveys for multi-agent RL. 
\citet{Parkes2015} is a survey about economic reasoning and AI.

In the following, we discuss 
centralized training for decentralized execution,
several issues in game theory, 
and games.

\subsection*{Centralized Training for Decentralized Execution}

A centralized critic can learn from all available state information conditioned on joint actions,
and each agent learns its policy from its own observation action history.
The centralized critic is only used during learning, 
and only the decentralized actor is needed during execution.
In the following, several recently propose approaches use StarCraft as the experimental testbed, 
e.g., \citet{Peng2017},
\citet{Foerster2018COMA}, and \citet{Rashid2018}.

\citet{Foerster2018COMA} propose the counterfactual multi-agent (COMA) actor-critic method.
In COMA,  policies are optimized with decentralized actors,
and Q-function is estimated with a centralized critic,
using a counterfactual baseline to marginalize out one agent's action, 
and fixing other agents' actions, for the purpose of multi-agent credit assignment.

Some papers propose communication mechanisms in MARL~\citep{Foerster2016,  Sukhbaatar2016}.
\citet{Peng2017} require communication.

\citet{Peng2017} propose a multiagent actor-critic framework, with a bidirectionally-coordinated network to form coordination among multiple agents in a team, 
deploying the concept of dynamic grouping and parameter sharing for better scalability.  
In the testbed of StarCraft, without human demonstration or labelled data as supervision, 
the proposed approach learns strategies for coordination similar to the level of experienced human players, 
like move without collision, hit and run, cover attack, and focus fire without overkill.

It is desirable to design an algorithm between the two extremes of independent RL and fully centralized RL approaches.
One way is to decompose Q function.
\citet{Sunehag2017} and \citet{Rashid2018} fall into this category.

\citet{Sunehag2017} propose value-decomposition networks (VDN) to represent 
the centralized action value function Q as a sum of value functions of individual agents.
In VDN, each agent trains its value function based on its observations and actions, 
and a decentralized policy is  derived  from its action value function.

\citet{Rashid2018} propose QMIX, 
so that each agent network represents an individual action value function, 
and a mixing network combines them into a centralized action value function,
with a non-linear approach, in contrast to the simple sum in VDN~\citep{Sunehag2017}.
Such factored representation allows complex centralized action value function, 
extraction of decentralized policies with linear time individual  argmax operations, and scalability.

\subsection*{Issues in Game Theory}


\citet{Heinrich2016} propose neural fictitious self-play (NFSP) to combine fictitious self-play with deep RL to learn approximate Nash equilibria for games of imperfect information in a scalable end-to-end approach without prior domain knowledge. NFSP is evaluated on two-player zero-sum games. In Leduc poker, NFSP approaches a Nash equilibrium, while common RL methods diverges. In Limit Texas Hold'em, a real-world scale imperfect-information game, NFSP performs similarly  to state-of-the-art, superhuman algorithms which are based on domain expertise. 


\citet{Lanctot2017} observe that independent RL, in which each agent learns by interacting with the environment, oblivious to other agents,  can overfit the learned policies to other agents' policies. 
The authors propose policy-space response oracle (PSRO), 
and its approximation, deep cognitive hierarchies (DCH),  
to compute best responses to a mixture of policies using deep RL,
and to compute new meta-strategy distributions using empirical game-theoretic analysis.
PSRO/DCH generalizes previous algorithms, like independent RL, iterative best response, 
double oracle, and fictitious play.
The authors present an implementation with centralized training for decentralized execution, as discussed below.
The authors experiment with grid world coordination, a partially observable game, 
and Leduc Poker (with a six-card deck), a competitive imperfect information game,
and show reduced joint-policy correlation (JPC), a new metric to quantify the effect of overfitting.


Social dilemmas, e.g., prisoner's dilemma, reveal the conflicts between collective and individual rationality.
Cooperation is usually beneficial for all.
However, parasitic behaviours like free-riding may result in the tragedy of the commons,
which makes cooperation unstable.
The formalism of matrix game social dilemmas (MGSD) is a popular approach.
However, as discussed in \citet{Leibo2017}, 
MGSD does not consider several important aspects of real world social dilemmas: 
they are temporally extended,
"cooperation and defection are labels that apply to policies implementing strategic decisions",
"cooperative may be a graded quantity",
cooperation and defection may not happen fully simultaneously,  since information about the starting of a strategy by one player would influence the other player,
and, decisions are mandatory although with only partial information about the world and other players.

\citet{Leibo2017} propose a sequential social dilemma (SSD) with MARL to tackle these issues.
The authors conduct empirical game theoretic analyses of two games, 
fruit gathering and wolfpack hunting,
and show that, when treating cooperation and defection as one-shot decisions as in MGSD,
they have empirical payoff matrices as prisoner's dilemma.
However, gathering and wolfpack are two different games, 
with opposite behaviours in the emergence and stability of cooperation.
SSDs can capture the differences between these two games,
using a factor to promote cooperation in gathering and to discourage cooperation in wolfpack.
The sequential structure of SSDs results in complex model to compute or to learn equilibria.
The authors propose to apply DQN to find equilibria for SSDs.

\subsection*{Games}



\citet{Jaderberg2018Quake} approach Capture the Flag, a 3D multi-player first-person video game,  
in an end-to-end manner using raw inputs including pixels and game points, 
with techniques of population based training, optimization of internal reward, and temporally hierarchical RL,
and achieve human-level performance, for the first time for multi-agent RL problems. 

\citet{Jaderberg2018Quake} propose to train a diverse population of agents, 
to form two teams against each other, and to train each agent independently in a decentralized way, only through interaction with the environment, 
without knowledge of environment model, other agents, and human policy prior, and without communication with other agents.  
Each agent learn an internal reward signal, to generate internal goals, such as capturing a flag, to complement the sparse game winning reward. 
The authors propose a two-tier optimization process.    
The inner optimization maximizes agents' expected discounted future internal rewards.
The outer optimization solves a meta-game, to maximize the meta-reward of winning the match,  
w.r.t. internal reward functions and hyperparameters, 
with meta transition dynamics provided by the inner optimization.
The inner optimization is solved with RL.
The outer optimization is solved with population based training (PBT)~\citep{Jaderberg2017PBT}, 
an online evolutionary method adapting internal rewards and hyperparameters and performing model selection by agent mutation, i.e., replacing under-performing agents with better ones.
Auxiliary signals~\citep{Jaderberg2017} and differentiable neural computer memory~\citep{Grave-DNC-2016} are also used to improve performance.
RL agents are trained asynchronously from thousands of concurrent matches on randomly generated maps.

The authors design an agent to achieve strong capacity and avoiding several common RL issues, 
with the integration of learning and evolution.  
Learning from a diverse population of agents on random maps helps achieve skills generalizable to variability of maps, number of players, and choice of teammates, and stability in partially observable multi-agent environments.
Learning an internal reward signal helps tackle the sparse reward problem and further the credit assignment issue.
The multi-timescale representation helps with memory and long term temporal reasoning for high level strategies.
The authors choose PBT instead of self play, since self play may be unstable in multi-agent RL environments, 
and needs more manipulation to support concurrent training for better scalability.

Experiments show that an agent can learn a disentangled representation to encode various knowledge of game situations, 
like conjunctions of flag status, re-spawn state, and room type,
associating with activation of some individual neurones,
and behaving like humans e.g., in navigating, following, and defending.
Such knowledge is acquired through RL training, rather than from explicit models. 
Experiments also show that the generalizable skills for tasks with random maps are supported by rich representation of spatial environments, induced by the temporal hierarchy and explicit memory module.
Experiments show human level performance, and a survey shows that the agents are more collaborative than human participants.
It appears that the training was conducted with less than 2000 commodity computers.

The authors mention the limitations of the current work: 
"the difficulty of maintaining diversity in agent populations, 
the greedy nature of the meta-optimisation performed by PBT, 
and the variance from temporal credit assignment in the proposed RL updates".
See a blog at \url{https://deepmind.com/blog/capture-the-flag/}.

\citet{SunPeng2018StarCraft} and \citet{Pang2018StarCraft} have beaten full-game built-in AI in StarCraft II.
OpenAI Five designs a Dota 2 agent for 5v5 plays, with a common RL algorithm, Proximal Policy Optimization (PPO) and self play, and beat human players.  
However, huge computation is involved, with 256 GPUs and 128,000 CPU cores. 
See \url{https://openai.com/five/}.


\clearpage

\section{Relational RL}
\label{relational}


Integrating reinforcement learning and relational learning~\citep{Getoor2007} 
is a promising approach to problem solving in AI.
Relational RL makes connections between RL and classical AI, 
for knowledge representation and reasoning, which we discuss briefly in Section~\ref{representation:knowledge}.

\citet{Dzeroski2001} propose relational RL.
\citet{Tadepalli2004} give an overview of relational RL, and identify several challenges: suitable function approximation to represent relational knowledge, generalization across objects, transferability across tasks, run-time planning and reasoning, and, incorporating prior knowledge. 
\citet{Tadepalli2004} further propose relational RL as a solution to these challenges.
\citet{Guestrin2003} introduce relational MDPs.
\citet{Diuk2008} introduce objected-oriented MDPs (OO-MDPs), a close approach to relational RL.
See more work, e.g., \citet{Mrowca2018NIPS}, \citet{Palm2018NIPS}, and \citet{Santoro2018RMC}.

See NIPS 2018 Workshop on  Relational Representation Learning at \url{https://r2learning.github.io}.

We discuss some papers about relational learning and relational RL below.

\subsection*{Relational Learning}


\citet{Battaglia2018GN} first discuss ways to incorporate relational inductive bias with deep learning.
In fully connected networks (FC), entities are units, and their relations are all to all, thus have only weak relational inductive bias.  
In convolutional neural networks (CNNs), entities are still units, or grid elements, e.g. pixels in images, and relations are local.  CNNs impose locality and translation  invariance as relational inductive bias. 
In recurrent neural networks (RNN), entities include input and hidden state at each time step, and relations are the mapping from previous hidden state and current input to the hidden state of this step. 
This mapping is reuse over time steps, thus temporal invariance is the relational inductive bias for RNN.

\citet{Battaglia2018GN} then propose graph network (GN) to incorporate relational inductive bias, to attempt to achieve combinatorial generalization, i.e., the capacity to use known elements to build new inferences, predictions and behaviours.
GN can operate on arbitrary relational structure, having explicit representation of entities (nodes) and relationships (edges), grounding in data.  
Node and edge permutations are invariant in GN. GN generalizes previous graph neural networks, e.g., \citet{Scarselli2009}. 


\citet{Santoro2017} propose relation networks (RNs) for relational reasoning in a plug-and-play way in neural networks. The authors experiment RN-augmented networks on visual question answering, text-based question answering, and reasoning about dynamic physical systems. 
\citet{Santoro2018RMC} propose a relational memory core (RMC), a memory module, to handle tasks with relational reasoning, using self attention~\citep{Vaswani2017} to deal with memory interaction. 
The authors test RMC on RL tasks, program evaluation, and language modelling. 
\citet{Battaglia2016} propose interaction network to learning about objects, relations and physics,
and evaluate it with n-body problems, rigid-body collision, and non-rigid dynamics.


\citet{Chen2018} propose a framework for iterative visual reasoning, beyond current recognition systems with CNNs.  
It is composed of a local module to store previous beliefs, a global model for graph reasoning.
It refines estimates by rolling out these models iteratively, and cross-feeding predictions to each other. 
The graph model consists of: 1) a knowledge graph, with nodes and edges to represent classes and their relationships respectively; 2) a region graph of the current image, with nodes  and edges as regions in the images and their spatial relationships; 3) an assignment graph, assigning regions to classes.
An attention mechanism is used to combine both local and global modules for final predictions.


\citet{Hudson2018} propose memory, attention, and control (MAC) cell,
with three units, namely, control, read, and write units,
 to construct recurrent compositional attention networks for reasoning, by imposing structural constraints in the operation of and interaction between cells, for the purpose of interpretability, generalization, computation efficiency, and data efficiency. 
\citet{Hudson2018} evaluate their proposed approach on a visual question answering (VQA) task with the CLEVR data set. Watch a video at \url{https://www.youtube.com/watch?v=jpNLp9SnTF8}.

\subsection*{Relational RL}

\citet{Zambaldi2018} propose to improve sample efficiency, generalization capacity, and interpretability of deep RL with relational reinforcement learning~\citep{Dzeroski2001}. 
The proposed network architecture consists of FC, CNNs, and a relational module, which uses self-attention~\citep{Vaswani2017}
to reason about interactions between entities, and to facilitate model-free policy optimization. 
\citet{Zambaldi2018} construct a navigation and planning task, BOX-World, to test the capacity of relational reasoning, and show good performance. 
\citet{Zambaldi2018} also experiment on StarCraft II mini-games in the the StarCraft II Learning Environment (SC2LE)~\citep{Vinyals2017SC2LE}, and achieve good performance on six mini-games.
Note that \citet{SunPeng2018StarCraft} and \citet{Pang2018StarCraft} have beaten full-game built-in AI in StarCraft II.


\citet{Wang2018NerveNet} propose NerveNet, resembling the neural nervous system to a graph, to learn structured policy for continuous control.
A policy is defined using the graph neural networks (GNN)~\citep{Scarselli2009}, in which, information is propagated over the agent structure, and actions are predicted for different parts of the agent.
The authors evaluate NerveNet in OpenAI Gym environment, and on transfer learning and multi-task learning tasks.
See  \url{http://www.cs.toronto.edu/~tingwuwang/nervenet.html} for open source and demo.


\citet{Keramati2018} propose strategic object oriented RL (SOORL) for model-learning with automatic model selection, and planning with strategic exploration. 
The authors achieve positive reward in Pitfall!, a hard Atari game. 
However, we note that, concurrently, \citet{Aytar2018} achieve much better results in Pitfall! and two other hard Atari games with an approach of self-supervision+RL, albeit relational and object oriented mechanisms are worth more efforts.

\citet{Yang2018PEORL} propose planning-execution-observation-RL (PEORL) to 
integrate hierarchical RL with symbolic planning for dynamic, uncertain environments.

\citet{Zambaldi2018} utilize pixel feature vectors resulting from CNNs as entities, a problem agnostic approach.
\citet{Keramati2018} utilize bounding boxes to detect object.  
\citet{Yang2018PEORL} work with manually crafted symbolic knowledge.
The performance would be further improved with  an advanced reasoning technique about images to find entities and relations, e.g.,~\citet{Chen2018}, \citet{Santoro2017}, \citet{Santoro2018RMC}, or, \citet{Battaglia2016}, etc.


\clearpage 
\section{Learning to Learn}
\label{meta}

Learning to learn, a.k.a. meta-learning, is learning about some aspects of learning.
It includes concepts as broad as transfer learning, multi-task learning,  one/few/zero-shot learning,
learning to optimize, learning to reinforcement learn, learning combinatorial optimization,
hyper-parameter learning, neural architecture design, automated machine learning (AutoML), continual learning, etc.
Learning to learn is a core ingredient to achieve strong AI~\citep{Lake2016},
and has a long history, 
e.g., \citet{Ellis1965}, \citet{Schmidhuber1987}, \citet{Bengio1991}, \citet{Sutton1992},  \citet{Thrun1998}, \citet{Hochreiter2001},  and \citet{Brazdil2009}.

\citet{Li2017} and \citet{Li2017OptNN}, along with the blog at
\url{http://bair.berkeley.edu/blog/2017/09/12/learning-to-optimize-with-rl/},
divide various learning to learn methods into three categories:
learning what to learn,
learning which model to learn, and,
learning how to learn.
The authors mention that, "roughly speaking, 'learning to learn' simply means learning something about learning".
The authors discuss that the term of learning to learn has the root in 
the idea of metacognition by Aristotle in 350 BC (\url{http://classics.mit.edu/Aristotle/soul.html}), 
which describes "the phenomenon that humans not only reason, but also reason about their own process of reasoning".
In the category of learning what to learn, the aim is to learn  values for base-model parameters, 
gaining the meta-knowledge of commonalities across the family of related tasks, 
and to make the base-learner useful for those tasks~\citep{Thrun1998}.
Examples in this category include methods for transfer learning, multi-task learning and few-shot learning.
In the category of learning which model to learn,
the aim is to learn which base-model is most suitable for a task~\citep{Brazdil2009},
gaining the meta-knowledge of correlations of performance between various base-models,
by investigating their expressiveness and searchability.
This learns the outcome of learning.
In the category of learning how to learn,
the aim is to learn the process of learning,
gaining the meta-knowledge of commonalities in learning algorithms behaviours.
There are three components for learning how to learn: 
the base-model, the base-algorithm to train the base-model, and the meta-algorithm to learn the base-algorithm.
The goal of learning how to learn is to design the meta-algorithm to learn the base-algorithm, which trains the base-model.
\citet{Bengio1991}, \citet{Andrychowicz2016}, \citet{Li2017}, and \citet{Li2017OptNN}  fall into this category.
\citet{Fang2017} study learning how to active learn. 
\citet{WangTongzhou2018NIPS} study how to learn MCMC proposals.
\citet{ZhangYu2018NIPS} study learning to multitask.
\citet{Negrinho2018NIPS} study learning beam search policies. 
\citet{Hsu2018} study unsupervised learning with meta-learning.

\citet{Finn2017MAML}, along with the blog at \url{https://github.com/cbfinn/maml},
summarize that there are three categories of methods for learning to learn,
namely, recurrent models, metric learning, and learning optimizers.
In the approach of recurrent models, a recurrent model, e.g. an LSTM, 
is trained to take in data points, e.g., (image, label) pairs for an image classification task, sequentially from the dataset, 
and then processes new data inputs from the task.
The meta-learner usually uses gradient descent to train the learner, 
and the learner uses the recurrent network to process new data.
\citet{Santoro2016}, \citet{Mishra2018}, \citet{Duan2016RL2}, and \citet{Wang2016LearnRL} fall into this category.
In the approach of metric learning, a metric space is learned to make learning efficient,
mostly for few-shot classification.
\citet{Koch2015}, \citet{Vinyals2016}, and \citet{Snell2017} fall into this category.
In the approach of learning optimizers, an optimizer is learned, using a meta-learner network to learn to update the learner network to make the learner learn a task effectively.
\citet{Wichrowska2017}, \citet{Andrychowicz2016}, \citet{Li2017}, and \citet{Li2017OptNN}  fall into this category.
Motivated by the success of using transfer learning for 
initializing computer vision network weights with the pre-trained ImageNet weights~\citep{Donahue2014},
\citet{Finn2017MAML} propose model-agnostic meta-learning (MAML)
to optimize an initial representation for a learning algorithm,
so that the parameters can be fine-tuned effectively from a few examples.

\citet{Duan2017thesis} gives a brief review of meta-learning.
The author discusses meta-learning for supervised learning,
including metric-based models, optimization-based models, and fully generic models,
and other applications.
The author also discusses meta-learning for control,
and proposes to learn reinforcement learning algorithms~\citep{Duan2016RL2}, and
one-shot imitation learning~\citep{Duan2017OneShot}.

Combinatorial optimization is critical for many areas, like social networks, telecommunications, and transportation.
Many combinatorial optimization problems are NP-hard, and algorithms follow three approaches: 
exact algorithms, approximate algorithms, and heuristics,
and all of them require specialized knowledge and human efforts for trail-and-error.
\citet{Dai2017graph} propose to automate combinatorial optimization using deep RL with graph embedding~\citep{Dai2016ICML}. 

We discuss learning to plan, including,
value iteration networks (VIN)~\citep{Tamar2016}, and
predictron~\citep{Silver-Predictron-2016},
and learning to search, MCTSnets~\citep{Guez2018},
in Chapter~\ref{model}.

See NIPS 2018 Workshop on Meta-Learning at \url{http://metalearning.ml/2018/}.

Continual learning~\citep{Chen2016Lifelong, Kirkpatrick2017, Lopez-Paz2017, XuJun2018NIPS} 
is important for achieving general intelligence.
See \citet{Singh2017} for a tutorial about continual learning.
See NIPS 2018 Workshop on Continual Learning at \url{https://sites.google.com/view/continual2018}. 
See ICML 2018 Workshop on Lifelong Learning: A Reinforcement Learning Approach at \url{https://sites.google.com/view/llarla2018/home}.

We discuss
few/one/zero-shot learning in Section~\ref{meta:one-shot-learning},
transfer and multi-task learning in Section~\ref{meta:transfer},
learning to optimize in Section~\ref{meta:learn-to-opt},
learning reinforcement learn in Section~\ref{meta:learnRL},
learning combinatorial optimization in Section~\ref{meta:learn-combinatorial},
and AutoML in Section~\ref{meta:AutoML}.

\subsection{Few/One/Zero-Shot Learning}
\label{meta:one-shot-learning}

As discussed in~\citep{Finn2017MAML}, the aim of few-shot meta-learning is to train a model adaptive to a new task quickly,
using only a few data samples and training iterations. 
\citet{Finn2017MAML} propose model-agnostic meta-learning (MAML)
to optimize an initial representation for a learning algorithm,
so that the parameters can be fine-tuned effectively from a few examples.
To illustrate MAML, consider a model $f_\theta$ with parameter $\theta$.
The model parameter $\theta$ becomes $\theta_i'$, when adapting to a new task $\mathcal{T}_i$.
We compute the new parameter $\theta_i'$ with one gradient descent update,
\begin{equation}
\theta_i' = \theta - \alpha \nabla_\theta \mathcal{L}_{\mathcal{T}_i} (f_\theta),
\end{equation} 
or more updates, on task $\mathcal{T}_i$. 
Here $\alpha$ is the step size.
We train model parameters by optimizing 
the meta-objective, 
\begin{equation}
\min_\theta \sum_{\mathcal{T}_i \sim p(\mathcal{T})} \mathcal{L}_{\mathcal{T}_i} (f_{\theta_i'})  = \min_\theta \sum_{\mathcal{T}_i \sim p(\mathcal{T})} \mathcal{L}_{\mathcal{T}_i} (f_{\theta - \alpha \nabla_\theta \mathcal{L}_{\mathcal{T}_i} (f_\theta)}),
\end{equation}
the performance of $f_{\theta_i'}$, w.r.t. $\theta$ across tasks sampled from $p(\mathcal{T})$.
Note, the meta-optimization is performed over the old parameters $\theta$, 
and the objective is computed using the new parameters $\theta'$.
This  aims to optimize the model parameters so that one or a few gradient steps on a new task will produce maximally effective behaviour on that task.
We perform the meta-optimization across tasks via SGD, and update $\theta$ as, 
\begin{equation}
\theta \leftarrow \theta - \beta \nabla_\theta \sum_{\mathcal{T}_i \sim p(\mathcal{T})} \mathcal{L}_{\mathcal{T}_i} (f_{\theta_i'}),
\end{equation}
where $\beta$ is the meta step size.
The intuition underlying MAML is that some internal representations are effective for adaptation, 
and the goal of MAML is to find model parameters sensitive to changes in the task,
so that after we draw a task from $p(\mathcal{T})$, 
when the direction of the gradient of the loss function changes,
small parameter changes will significantly improve the loss function.

MAML works for both supervised learning and reinforcement learning.
Experiments show good results on few-shot classification, regression, and policy gradients RL with neural network policies.
See a blog about learning to learn and MAML at \url{http://bair.berkeley.edu/blog/2017/07/18/learning-to-learn/}

\citet{Finn2018universal} show that
deep representation integrated with gradient descent has sufficient capacity to approximate any learning algorithms,
i.e., the universality of meta-learning,
and show empirically that gradient-based meta-learning found learning strategies with better generalization capacity than recurrent models. 
\citet{Grant2018} treat gradient-based meta-learning as hierarchical Bayes.
\citet{Finn2018NIPS} study probabilistic MAML.
\citet{Yoon2018NIPS} study Bayesian MAML.

\citet{Al-Shedivat2018meta} propose to 
use the framework of learning to learn for continuous adaptation in non-stationary and competitive environments,
by treating a non-stationary environment as a sequence of stationary tasks.
The authors develope a gradient-based meta-learning algorithm adaptive in dynamically changing and adversarial scenarios based on MAML~\citep{Finn2017MAML}, 
by anticipating changes in environment and updating policies accordingly. 
The proposed approach attempt to handle Markovian dynamics on two level of hierarchy:
at the upper level for dynamics of tasks, and at the lower level for MDPs representing particular tasks.
The authors evaluate the performance of the proposed approach 
1) on a single-agent multi-leg robots locomotion task in MuJoCo physics simulator,
with handcrafted nonstationarity, by selecting a pair of legs to scale down the torques applied to joints, until fully paralyzed,
and, 2) on iterated adaptation games in RoboSumo,
which is in a 3D environment, with simulated physics, having pairs of agents to compete,
and not only non-stationary but also adversarial.
The authors assume that trajectories from the current task contain some information about the next task,
so that tasks become dependant sequentially.
The proposed algorithm may not take advantage of more data,
and it could diverge when there are large distributional changes from iteration to iteration.

\citet{Lake2015} propose an one-shot concept learning model, for handwritten characters in particular, with probabilistic program induction.
\citet{Koch2015} propose siamese neural networks with metric learning for one-shot image recognition.
\citet{Vinyals2016} design matching networks for one-shot classification.
\citet{Duan2017OneShot} propose a model for one-shot imitation learning with attention for robotics.
\citet{Johnson2017} present zero-shot translation for Google's multilingual neural machine translation system.
\citet{Kaiser2017} design a large scale memory module for life-long one-shot learning to remember rare events.
\citet{Kansky2017} propose Schema Networks for zero-shot transfer with a generative causal model of intuitive physics.
\citet{Snell2017} propose prototypical networks for few/zero-shot classification by learning a metric space to compute distances to prototype representations of each class.
\citet{George2017} propose a generative vision model to train with high data efficiency, breaking text-based CAPTCHA.
\citet{LiuShichen2018NIPS} study generalization of zero-shot learning with deep calibration network.

\subsection{Transfer/Multi-task Learning}
\label{meta:transfer}

Transfer learning is about transferring knowledge learned from different domains, possibly with different feature spaces and/or different data distributions~\citep{Taylor09, Pan2010, Weiss2016}. As reviewed in~\citet{Pan2010}, transfer learning can be inductive, transductive, or unsupervised. Inductive transfer learning includes self-taught learning and multi-task learning. Transductive transfer learning includes domain adaptation and sample selection bias/covariance shift.
\citet{Taylor09} compare RL transfer learning algorithms w.r.t. the following performance metrics:
jumpstart, asymptotic performance, total reward, transfer ratio, time to threshold,
and against the following dimensions:
task difference assumption, source task selection, task mapping, transferred knowledge, and allowed learners.
Multitask learning~\citep{Caruana1997, Zhang2018, Ruder2017} learns related tasks with a shared representation in parallel, 
leveraging information in related tasks as an inductive bias, to improve generalization, and to help improve learning for all tasks.
The modular structure of hierarchical RL approaches is usually conducive to transfer and multi-task learning, which we discussed in Chapter~\ref{hierarchical}.


\citet{WhyeTeh2017} propose Distral, distill \& transfer learning, 
for joint training of multiple tasks,
by sharing a distilled policy, trained to be the centroid of policies for all tasks, to capture common behavioural  structure across tasks,
and training each task policy to be close to the shared policy.
The design of Distral is to overcome issues in transfer and multi-task learning,
that in the approach of share neural network parameters, gradients from different tasks may interfere each other negatively,
and that one task may dominate the learning of the shared model, due to different reward functions of different tasks. 
\citet{WhyeTeh2017} design Distral following the techniques of 
distillation~\citep{Bucila2006, Hinton2014}, and  
soft Q-learning~\citep{Haarnoja2017, Nachum2017Gap}, a.k.a., G-learning~\citep{Fox2016}.
The authors observe that,
the distillation arises naturally, when optimize task models towards a distilled model, 
when using KL divergence as a regularization.
Moreover, the distilled model serves as a regularizer for task models training,
and it may help transferability by regularizing neural networks in a space more semantically meaningful, 
e.g., for policies, rather than for network parameters. 
Distral can be instantiated in several forms, and it outperforms empirically the baseline A3C algorithms in 
grid world and complex 3D environments.

\citet{Barreto2017} propose a transfer framework for tasks with different reward functions but the same environment dynamics, based on two ideas, namely, successor features and generalized policy improvement.
Successor features are a value function representation decoupling environment dynamics from rewards,
extending the successor representation~\citep{Dayan1993}, as discussed in Section~\ref{representation:classics}, 
for continuous tasks with function approximation. 
Generalized policy improvement operates on multiple policies, 
rather than one policy as in the policy improvement in dynamic programming.
The authors integrate these two ideas for transfer learning among tasks,
and establish two theorems, one for performance guarantee of a task before any learning,
and another for performance guarantee of a task if it had seen similar tasks.
The author design a method based on these ideas and analysis, 
and show good performance on navigation tasks and a task to control a simulated robotic arm. 

\citet{Gupta2017} formulate the multi-skill problem for two agents to learn multiple skills, 
define the common representation using which to map states and to project the execution of skills, 
and design an algorithm for two agents to transfer the informative feature space maximally to transfer new skills, 
with similarity loss metric, autoencoder, and reinforcement learning.  
The authors validate their proposed approach with two simulated robotic manipulation tasks.

As a practical example of transfer learning in computer vision, 
\citet{Kornblith2018} investigate the transferability of ImageNet architectures and features, 
for 13 classification models on 12 image classification tasks, in three settings: as fixed feature extractors, fine-tuning, and training from random initialization.
The authors observe that, ImageNet classification network architectures generalize well across datasets, 
but fixed ImageNet features do not transfer well.

See recent work in transfer learning/multi-task learning e.g., 
\citet{Andreas2017}, 
\citet{Dong2015}, 
\citet{Kaiser2017OneModel}, 
\citet{Kansky2017}, 
\citet{Killian2017}
\citet{Long2015}, 
\citet{Long2016}, 
\citet{Long2017},
\citet{Mahajan2018},
\citet{Maurer2016}, 
\citet{McCann2017},
\citet{Mo2018}, 
\citet{Parisotto2016}, 
\citet{Papernot2017}, 
\citet{Perez2017}, 
\citet{Rajendran2017},
\citet{Sener2018NIPS}, 
\citet{Smith2017multi}, 
\citet{Sohn2018NIPS},
\citet{Yosinski2014}, and, 
\citet{ZhaoHan2018NIPS}. 



See NIPS 2015 workshop on Transfer and Multi-Task Learning: Trends and New Perspectives
at \url{https://sites.google.com/site/tlworkshop2015/}.

We will discuss sim-to-real transfer learning in robotics in Chapter~\ref{robotics:sim2real}. 

\subsection{Learning to Optimize}
\label{meta:learn-to-opt}

\citet{Li2017} and \citet{Li2017OptNN} propose to automate unconstrained continuous optimization algorithms with RL, in particular, guided policy search~\citep{Levine2016}.
Algorithm~\ref{opt} presents a general structure of optimization algorithms.
\citet{Li2017} and \citet{Li2017OptNN} formulate a RL problem as follows:
1) the state is the current iterate, $x^{(i)}$, and may also include some features along the historical optimization trajectory, 
like the history of gradients, iterates, and objective values;
2) the action is the step vector, $\Delta x$, which updates the iterate $x^{(i)}$;
and, 3) the policy is the update formula $\phi(\cdot)$,
which depends on the current iterate, and the history of gradients, iterates, and objective values.
Thus, learning the policy is equivalent to learning the optimization algorithm.
One possible cost function is the objective function value at the current iterate.
A RL algorithm in this problem does not have access to the state transition model.
See a blog at \url{http://bair.berkeley.edu/blog/2017/09/12/learning-to-optimize-with-rl/}.
See also \citet{Andrychowicz2016} and \citet{Bello2017}.

\begin{algorithm}[h]
\SetAlgoNoLine
\hrulefill

\textbf{Input: } objective function $f$

\hrulefill

$x^{(0)} \leftarrow$ random point in the domain of $f$
\For{$i=1, 2, 3, \dots$}{
	$\Delta x \leftarrow \phi(f, \{x^{(0)}, \dots, x^{(i-1)}\})$
	\If{stopping condition is met}{
	        return $x^{(i-1)}$
	}
	$x^{(i)} \leftarrow x^{(i-1)} + \Delta x$
}

\hrulefill
\caption{General structure of optimization algorithms, adapted from \citet{Li2017OptNN}}
\label{opt}
\end{algorithm}

\subsection{Learning Reinforcement Learn}
\label{meta:learnRL}

\citet{Xu2018meta} investigate a fundamental problem in RL to discover an optimal form of return,
and propose a gradient-based meta-learning algorithm to learn the return function,
by tuning meta-parameters of the return function,  e.g., the discount factor, $\gamma$, the bootstrapping parameter, $\lambda$, etc.,
in an online fashion, when interacting with the environment.
This is in contrast to many recent work in learning to learn that are in a setting of transfer/multi-task learning.
Experiments on Atari games show good results.
The technique would be general for other components of the return function, the learning update rule, and hyperparameter tunning.

\citet{Wang2018cortex} investigate that prefrontal cortex works as a meta-reinforcement learning system.
Learning to learn, or meta-learning, is related to the phenomenon that our brain can do so much with so little.
A hypothesis is that we learn on two timescales: learning on specific examples in short term, and learning abstract skills or rules over long term.
\citet{Kahneman2011} describes  two modes of thought: one, fast, instinctive and emotional; another, slower, more deliberative, and more logical. 
See a blog at \url{https://deepmind.com/blog/prefrontal-cortex-meta-reinforcement-learning-system/}.

\citet{Duan2016RL2} and \citet{Wang2016LearnRL} propose to learn a flexible RNN model to handle a family of RL tasks, to improve sample efficiency, learn new tasks in a few samples, and benefit from prior knowledge. 
The agent is modelled with RNN, with inputs of observations, rewards, actions and termination flags. The weights of RNN are trained with RL, in particular, TRPO in \citet{Duan2016RL2} and A3C in \citet{Wang2016LearnRL}.  
\citet{Duan2016RL2} and \citet{Wang2016LearnRL} achieve similar performance as specific RL algorithms for various problems. 

\citet{Houthooft2018NIPS} propose evolved policy gradients with meta-learning.
\citet{Gupta2018NIPS} propose meta-RL of structured exploration strategies.
\citet{Stadie2018NIPS} study the importance of sampling in meta-RL.

\subsection{Learning Combinatorial Optimization}
\label{meta:learn-combinatorial}

\citet{Dai2017graph} propose to automate algorithm design for combinatorial optimization problems on graphs using deep RL with graph embedding~\citep{Dai2016ICML},
by learning heuristics on problem instances from a distribution $\mathbb{D}$ of a graph optimization problem, to generalize to unseen instances from $\mathbb{D}$. 

The authors propose to construct a feasible solution following a greedy algorithm design pattern, 
by successively adding nodes based on the graph structure and the current partial solution.
A deep graph embedding, structure2vec~\citep{Dai2016ICML}, 
is used to represent nodes, considering their properties in the graph, 
and helps generalize the algorithm to different graph sizes.

The graph embedding is also used to represent state, action, value function, and policy.
A state is a sequence of nodes on a graph.
An action is selecting a node that is not part of the current state.
The reward is the change in the cost function after taking an action, i.e., adding a node, and transition to a new state.
A deterministic greedy policy is used based on the action value function.
For example, in the Minimum Vertex Cover (MVC) problem, we are to find a subset of nodes in a graph, so that every edge is covered and the number of nodes selected is minimized.
In MVC, a state is the subset of nodes selected so far, an action is to add a node to the current subset, the reward is -1 for each action, and the termination condition is all edges are covered. 
The authors propose to learn a greedy policy parameterized by the graph embedding network using $n$-step fitted Q-learning.  

\citet{Dai2017graph} evaluate the proposed approach on three graph combinatorial optimization problems:
Minimum Vertex Cover (MVC), Maximum Cut (MAXCUT), and Traveling Salesman Problem (TSP),  
and compare with pointer networks with actor-critic~\citep{Bello2016}, and strong baseline approximate and heuristics algorithms for each problem respectively. 
\citet{Dai2017graph} achieve good performance on both synthetic and real graphs.
See the open source at \url{https://github.com/Hanjun-Dai/graph_comb_opt}.

\citet{Vinyals2015} propose pointer networks 
to learn the conditional probability of an output sequence of discrete tokens,
corresponding to positions in an input sequence,
by generating output sequence with attention as a pointer to select an element of the input sequence,
and evaluate performance on three combinatorial optimization problems,
finding planar convex hulls, computing Delaunay triangulations, and the planar Travelling Salesman Problem.
Pointer networks are graph-agnostic, in contrast to the graph embedding in \citet{Dai2017graph}.

\subsection{AutoML}
\label{meta:AutoML}

AutoML is about automating the process of machine learning.
See a website for AutoML at \url{http://automl.chalearn.org}.
See NIPS 2018 AutoML for Lifelong Machine Learning Competition at
\url{https://www.4paradigm.com/competition/nips2018}.
See also a usable machine learning project~\citet{Bailis2017}.

Neural network architecture design is one particular task of AutoML.
Neural architecture design is a notorious, nontrivial engineering issue.  
Neural architecture search provides a promising avenue to explore.
See a survey~\citep{Elsken2018}.

\citet{Zoph2017} propose neural architecture search to generate neural networks architectures with an RNN trained by RL, in particular, REINFORCE, searching from scratch in variable-length architecture space, to maximize the expected accuracy of the generated architectures on a validation set. In the RL formulation, a controller generates hyperparameters as a sequence of tokens,  which are actions chosen from hyperparameters spaces. Each gradient update to the policy parameters corresponds to training one generated network to convergence. An accuracy on a validation set is the reward signal. The neural architecture search can generate convolutional layers, with skip connections or branching layers, and recurrent cell architectures. The authors design a parameter server approach to speed up training. Comparing with state-of-the-art methods, the proposed approach achieves competitive results for an image classification task with CIFAR-10 dataset, and better results for a language modeling task with Penn Treebank. 

\citet{Zoph2017Transfer}  propose to transfer the architectural building block learned with the neural architecture search~\citep{Zoph2017} on small dataset to large dataset for scalable image recognition. \citet{Baker2017} propose a meta-learning approach, using Q-learning with $\epsilon$-greedy exploration and experience replay, to generate CNN architectures automatically for a given learning task. \citet{Zhong2017} propose to construct network blocks to reduce the search space of network design, trained by Q-learning. 
See also \citet{Liu2017arch}, \citet{Liu2017H}, \citet{Real2017}, and \citet{Real2018}.
Note that \citet{Real2018} show that evolutionary approaches can match or surpass human-crafted and RL-designed image neural network classifiers.

\citet{Jin2018} propose a Bayesian approach for efficient search.
See \citet{Cai2018NAS} for an approach with limited computation resources (200 GPU hours).
See also
\citet{ChenLiangChieh2018NIPS}, 
\citet{Kandasamy2018NIPS}, 
\citet{LuoRenqian2018NIPS}, and 
\citet{WongCatherine2018NIPS}.

There are recent works exploring new neural architectures (manually). 
\citet{Vaswani2017} propose a new archichitecture for translation that replaces CNN and RNN with attention and positional encoding. 
\citet{Kaiser2017OneModel} propose to train a single model, MultiModel, which is composed of convolutional layers, an attention mechanism, and sparsely-gated layers, to learn multiple tasks from various domains, including image classification, image captioning and machine translation. 
\citet{Wang-Dueling-2016} propose the dueling network architecture to estimate state value function and associated advantage function, to combine them to estimate action value function for faster convergence. 
\citet{Tamar2016} introduce value iteration networks (VIN), a fully differentiable CNN planning module to approximate the value iteration algorithm, to learn to plan. 
\citet{Silver-Predictron-2016} propose the predictron to integrate learning and planning into one end-to-end training procedure with raw input in Markov reward process.
These neural architectures were designed manually.
It would be interesting to see if learning to learn can help automate such neural architecture design.

Neural architecture design has already had industrial impact.
Google, among others, is working on AutoML, in particular, AutoML Vision, and extends AutoML to NLP and contact center, etc.
See blogs about Google AutoML at 
\url{http://goo.gl/ijBjUr},
\url{http://goo.gl/irCvD6}, and, 
\url{http://goo.gl/VUzCNt}.
See Auto-Keras at \url{https://autokeras.com}.

There are recent interesting work in AutoML.
\citet{He2018AMC} propose AutoML for model compression.
\citet{Cubuk2018AutoAugment} propose AutoAugment to automate data augmentation for images.
\citet{ChenTianqi2018} study learning to optimize tensor programs.

See 2018 International Workshop on Automatic Machine Learning (collocated with the Federated AI Meeting, ICML, IJCAI, AMAS, and ICCBR) at \url{https://sites.google.com/site/automl2018icml/}.

One limitation of neural architecture design is that the network components are manually designed, 
and it is not clear if AI has the creativity to discover new components, 
e.g., discovering residual connections before ResNets were designed.


\clearpage


\newpage

\section*{Part III: Applications}

\addcontentsline{toc}{section}{Part III: Applications}

Reinforcement learning has a wide range of applications.  
We discuss games in Chapter~\ref{games} and robotics in Chapter~\ref{robotics}, two classical RL application areas. 
Games are important testbeds for RL/AI. Robotics will be critical in the era of AI.
Natural language processing (NLP) follows in Chapter~\ref{NLP}, which enjoys wide and deep applications of RL recently. 
Next we discuss computer vision in Chapter~\ref{CV}, in which, there are efforts for integration with language.
In Chapter~\ref{fin}, we discuss finance and business management,  
which have natural problems for RL.
We discuss more applications in Chapter~\ref{more-apps},
healthcare in Section~\ref{healthcare}, 
education in Section~\ref{education}. 
energy  in Section~\ref{energy},
transportation in Section~\ref{transportation},
and computer systems in Section~\ref{systems}. 
We attempt to put reinforcement learning in the wide context of science, engineering, and art in Section~\ref{science},

Reinforcement learning is widely utilized in operations research~\citep{Powell11}, e.g., supply chain, inventory management, resource management, etc; we do not list it as an application area --- it is implicitly a component in application areas like energy and transportation. We do not list smart city, an important application area of AI, as it includes several application areas here: healthcare, education, energy, transportation, etc. 


These application areas build on RL techniques as discussed in previous chapters, 
and may overlap with each other, 
e.g., a robot may need skills from application areas like computer vision and NLP. 
 
\begin{figure}[h]
\includegraphics[width=0.9\linewidth]{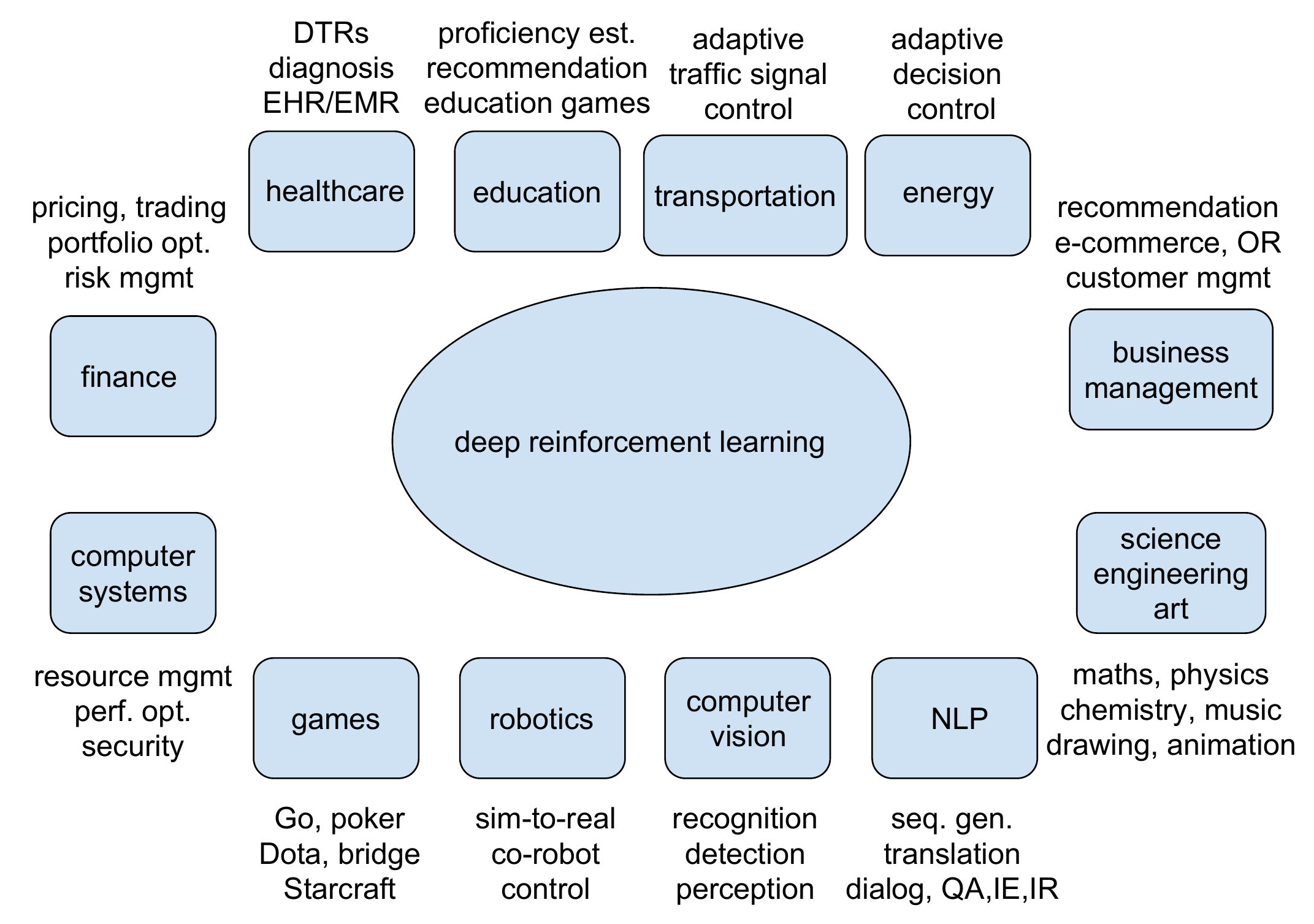}
\caption{Deep Reinforcement Learning Applications}
\label{apps}
\end{figure}

RL is usually for sequential decision making. 
However, some problems, seemingly non-sequential on surface, 
like neural network architecture design~\citep{Zoph2017}, and, model compression~\citep{He2018AMC}, have been approached by RL. 
Creativity would push the frontiers of deep RL further w.r.t. core elements, important mechanisms, and applications.
RL is probably helpful, if a problem can be regarded as or transformed to a sequential decision making problem, 
and states, actions, maybe rewards, can be constructed.
Many problems with manual design of strategies may be automated, and RL is a potential solution method.
We illustrate deep RL applications in Figure~\ref{apps}.
We maintain a blog, Reinforcement Learning Applications, at \url{https://medium.com/@yuxili/}.

\clearpage

\newpage


\section{Games}
\label{games}

Games provide excellent testbeds for AI algorithms,
dated back to the ages of Alan Turing, Claude Shannon, and John von Neumann. 
In games, we have good or even perfect simulators, and we can generate unlimited data.
We have seen great achievements of human-level or super-human performance in computer games, e.g.,
\begin{itemize}
\item Chinook~\citep{Schaeffer1997, Schaeffer2007} for Checkers,
\item Deep Blue~\citep{Campbell2002} for chess,
\item Logistello~\citep{Buro1999} for Othello,
\item TD-Gammon~\citep{Tesauro1994} for Backgammon, 
\item GIB~\citep{Ginsberg2001} for contract bridge,
\item MoHex~\citep{Huang2013,Gao2017} for Hex,
\item DQN~\citep{Mnih-A3C-2016} and \citet{Aytar2018} for Atari 2600 games,
\item AlphaGo~\citep{Silver-AlphaGo-2016} and AlphaGo Zero~\citep{Silver-AlphaGo-2017} for Go,
\item Alpha Zero~\citep{Silver-AlphaZero-2017} for chess, shogi, and Go,
\item Cepheus~\citep{Bowling2015}, DeepStack~\citep{Moravcik2017}, 
and Libratus~\citep{Brown2017,Brown2017Science} for heads-up Texas Hold'em Poker,
\item \citet{Jaderberg2018Quake} for Quake III Arena Capture the Flag, 
\item OpenAI Five, for Dota 2 at 5v5, \url{https://openai.com/five/},
\item \citet{Zambaldi2018}, \citet{SunPeng2018StarCraft}, and \citet{Pang2018StarCraft} for StarCraft II.
\end{itemize}

\citet{Schaeffer1997}, \citet{Buro1999}, \citet{Ginsberg2001}, and \citet{Campbell2002}  employ heuristic search techniques~\citep{Russell2009},  in particular, alpha-beta search.
\citet{Tesauro1994}, \citet{Mnih-A3C-2016}, \citet{Silver-AlphaGo-2016}, \citet{Silver-AlphaGo-2017}, \citet{Silver-AlphaZero-2017}, \citet{Jaderberg2018Quake}, OpenAI Five, \citet{Zambaldi2018}, \citet{SunPeng2018StarCraft}, and \citet{Pang2018StarCraft} are powered with (deep) reinforcement learning; 
the "Alpha series" are also equipped with Monte Carlo tree search (MCTS), a heuristic search technique.
\citet{Aytar2018} follows a self-supervised approach, together with deep RL.
\citet{Huang2013} employ MCTS, and \citet{Gao2017} extend it with deep learning.
\citet{Bowling2015} and \citet{Moravcik2017}, handle imperfect information with counterfactual regret minimization (CFR), and follow generalized policy iteration. \citet{Brown2017,Brown2017Science} utilize classical search techniques.

As deep RL has achieved human-level or superhuman performance for many two-play games,
multi-player games are at the frontier for scientific discovery of AI,
with outstanding achievements already achieved for games like Quake III Arena Capture the Flag and Dota 2 5v5.  

We discuss three categories of games, namely, board games in Section~\ref{games:board}, card games in Section~\ref{games:card}, and video games in Section~\ref{games:video}, loosely for two-player perfect information zero-sum games, imperfect information zero-sum games, and games with video frames as inputs, mostly with partial observability or imperfect information, respectively.

The above classification is not exhaustive, e.g., it does not include single agent, non-board, non-card, non-video games, like Rubik's Cube~\citep{McAleer2018}.
Computer games have much wider topics, e.g., storytelling~\citep{Thue2007}.

See~\citet{Yannakakis2018} for a book on AI and games.
See \citet{Justesen2017Survey} for a survey about applying deep (reinforcement) learning to video games.
See \citet{Ontanon2013} for a survey about Starcraft. Check AIIDE and CIG Starcraft AI Competitions, 
and its history at \url{https://www.cs.mun.ca/~dchurchill/starcraftaicomp/history.shtml}.
See \citet{Lin2017} for a StarCraft dataset.

\subsection{Board Games}
\label{games:board}

Board games like chess, Go and, Backgammon, are classical testbeds for RL/AI algorithms. In such games, players have prefect information of two players. \citet{Tesauro1994} approach Backgammon using neural networks to approximate value function learned with TD learning, and achieve human level performance.  

\subsection*{Computer Go}
\label{games:AlphaGo}

The challenge of solving computer Go comes from not only the gigantic search space of size $250^{150}$, an astronomical number, but also the hardness of position evaluation~\citep{Muller2002}, which was successfully used in solving many other games, like chess and Backgammon.

AlphaGo~\citep{Silver-AlphaGo-2016}, a computer Go program, won the human European Go champion, 5 games to 0, in October 2015, and became the first computer Go program to won a human professional Go player without handicaps on a full-sized 19 $\times$ 19 board. Soon after that in March 2016, AlphaGo defeated Lee Sedol, an 18-time world champion Go player, 4 games to 1, making headline news worldwide. This set a landmark in AI.  AlphaGo defeated Ke Jie 3:0 in May 2017. 
AlphaGo Zero~\citep{Silver-AlphaGo-2017} further improved previous versions by learning a superhuman computer Go program without human knowledge.  
Alpha Zero~\citep{Silver-AlphaZero-2017} generalized the learning framework in AlphaGo Zero to more domains.
See blogs for AlphaGo at \url{https://deepmind.com/research/alphago/},
and for AlphaGo Zero at \url{https://deepmind.com/blog/alphago-zero-learning-scratch/}.

\citet{Tian2016} also investigate computer Go.
See Facebook open source ELF OpenGo, \url{https://github.com/pytorch/ELF/}, and a blog, 
\url{https://research.fb.com/facebook-open-sources-elf-opengo/}.

\subsection*{AlphaGo: Training pipeline and MCTS}

We discuss briefly how AlphaGo works based on \citet{Silver-AlphaGo-2016}  and \citet{Sutton2018}. See \citet{Sutton2018} for an intuitive description of AlphaGo. 

AlphaGo is built with techniques of deep convolutional neural networks,  supervised learning, reinforcement learning, and Monte Carlo tree search (MCTS)~\citep{Browne2012, Gelly2007, Gelly12}. AlphaGo is composed of two phases: neural network training pipeline and MCTS. The training pipeline phase includes training a supervised learning (SL) policy network from expert moves, a fast rollout policy,  a RL policy network, and a RL value network.  

The SL policy network has convolutional layers, ReLU nonlinearities, and an output softmax layer representing probability distribution over legal moves. The inputs to the CNN are 19 $\times$ 19 $\times$ 48 image stacks, where 19 is the dimension of a Go board and 48 is the number of features. State-action pairs are sampled from expert moves to train the network with stochastic gradient ascent to maximize the likelihood of the move selected in a given state. The fast rollout policy uses a linear softmax with small pattern features. 

The RL policy network improves SL policy network, with the same network architecture, and the weights of SL policy network as initial weights, and policy gradient for training. The reward function is +1 for winning and -1 for losing in the terminal states, and 0 otherwise. Games are played between the current policy network and a random, previous iteration of the policy network, to stabilize the learning and to avoid overfitting. Weights are updated by stochastic gradient ascent to maximize the expected outcome.

The RL value network still has the same network architecture as SL policy network, except the output is a single scalar predicting the value of a position. The value network is learned in a Monte Carlo policy evaluation approach. To tackle the overfitting problem caused by strongly correlated successive positions in games, data are generated by self-play between the RL policy network and itself until game termination.  The weights are trained by regression on state-outcome pairs, using stochastic gradient descent to minimize the mean squared error between the prediction and the corresponding outcome. 

In MCTS phase, AlphaGo selects moves by a lookahead search. It builds a partial game tree starting from the current state, in the following stages: 1) select a promising node to explore further, 2) expand a leaf node guided by the SL policy network and collected statistics, 3) evaluate a leaf node with a mixture of the RL value network and the rollout policy, 4) backup evaluations to update the action values.  A move is then selected.    

\subsection*{AlphaGo Zero}

AlphaGo Zero can be understood as following a generalized policy iteration scheme, incorporating MCTS inside the training loop to perform both policy improvement and policy evaluation.
MCTS may be regarded as a policy improvement operator. It outputs move probabilities stronger than raw probabilities of the neural network.
Self-play with search may be regarded as a policy evaluation operator. It uses MCTS to select moves, and game winners as samples of value function.
Then the policy iteration procedure updates the neural network's weights to match the move probabilities and value more closely with the improved search probabilities and self-play winner, and conduct self-play with updated neural network weights in the next iteration to make the search stronger.

The features of AlphaGo Zero~\citep{Silver-AlphaGo-2017}, comparing with AlphaGo~\citep{Silver-AlphaGo-2016}, are: 
1) it learns from random play, with self-play RL, without human data or supervision;
2) it uses black and white stones from the board as input, without any manual feature engineering;
3) it uses a single neural network to represent both policy and value, rather than separate policy network and value network; and
4) it utilizes the neural network for position evaluation and move sampling for MCTS, and it does not perform Monte Carlo rollouts. 
AlphaGo Zero deploys several recent achievements in neural networks:  residual convolutional neural networks (ResNets), batch normalization, and rectifier nonlinearities.

AlphaGo Zero has three main components in its self-play training pipeline executed in parallel asynchronously: 1) optimize neural network weights from recent self-play data continually; 2) evaluate players continually; 3) use the strongest player to generate new self-play data.  

When AlphaGo Zero playing a game against an opponent, MCTS searches from the current state, with the trained neural network weights, to generate move probabilities, and then selects a move.

We present a brief, conceptual pseudo code  in Algorithm~\ref{AlphaGoZero} for training in AlphaGo Zero, conducive for easier understanding. 

\begin{algorithm}
\SetAlgoNoLine
\textbf{Input: } the raw board representation of the position, its history, and the colour to play as 19 $\times$ 19 images; game rules; a game scoring function; invariance of game rules under rotation and reflection, and invariance to colour transposition except for komi\\
\textbf{Output: } policy (move probabilities) $p$, value $v$\\
\vspace{3mm}
initialize neural network weights $\theta_0$ randomly\\
//AlphaGo Zero follows a generalized policy iteration procedure\\
\For{each iteration $i$}{
        \vspace{3mm}
        // termination conditions: \\
        // 1. both players pass \\
        // 2. the search value drops below a resignation threshold \\
        // 3. the game exceeds a maximum length \\
        \vspace{3mm}
        initialize $s_0$\\
        
	\For{each step $t$, until termination at step $T$}{
		\vspace{3mm}
		
		// MCTS can be viewed as a policy improvement operator \\
		// search algorithm: asynchronous policy and value MCTS algorithm (APV-MCTS)\\
	        // execute an MCTS search $\pi_{t} = \alpha_{\theta_{i-1}}(s_t)$ with previous neural network $f_{\theta_{i-1}}$\\
	        // each edge $(s, a)$ in the search tree stores a prior probability $P(s, a)$, a visit count $N(s, a)$, \\
	        // and an action value $Q(s, a)$\\
		\While{computational resource remains} {
		select: each simulation traverses the tree by selecting the edge with maximum upper confidence bound $Q(s, a) + U(s, a)$,
		where $U(s, a) \propto P(s, a) / (1 + N(s, a))$ \\
		expand and evaluate: the leaf node is expanded and the associated position $s$ is evaluated by the neural network, $(P(s, \cdot), V(s)) = f_{\theta_i}(s)$; the vector of $P$ values are stored in the outgoing edges from $s$\\
		backup: each edge $(s, a)$ traversed in the simulation is updated to increment its visit count $N(s, a)$, and to update its action value to the mean evaluation over these simulations, $Q(s,a) = 1/N(s, a) \sum_{s'|s, a \rightarrow s'} V(s')$, where $s'|s, a \rightarrow s'$ indicates that a simulation eventually reached $s'$ after taking move $a$ from position $s$\\
		}
	
		 // self-play with search  can be viewed as a policy evaluation operator: select each move with the improved MCTS-based policy, use the game winner as a sample of the value\\ 
		play: once the search is complete, search probabilities $\pi \propto N^{1/\tau}$ are returned, where $N$ is the visit count of each move from root and $\tau$ is a parameter controlling temperature;      
	        play a move by sampling the search probabilities $\pi_{t}$, transition to next state $s_{t+1}$\\     
	}
	\vspace{3mm}
	score the game to give a final reward $r_{T} \in \{-1, +1\}$\\
	\For{each step $t$ in the last game}{
	        $z_t \leftarrow \pm r_{T}$, the game winner from the perspective of the current player\\
	        store data as $(s_t, \pi_{t}, z_t)$\\     
	}
	sample data $(s, \pi, z)$ uniformly among all time-steps of the last iteration(s) of self-play\\
	\vspace{3mm}
	//train neural network weights $\theta_i$\\
	//optimizing loss function $l$ performs both policy evaluation, via $(z-v)^2$, and policy improvement, via $- \pi^{T} \log p$, in a single step\\
	adjust the neural network $(p, v) = f_{\theta_i}(s)$: \\
	to minimize the error between the predicted value $v$ and the self-play winner $z$, and \\
	to maximize similarity of neural network move probabilities $p$ to search probabilities $\pi$\\
	
	specifically, adjust the parameters $\theta$ by gradient descent on a loss function \\
	$(p, v) = f_{\theta_i}(s)$ and $l = (z-v)^2 - \pi^{T} \log p + c \| \theta_i \|^2 $ \\
	$l$ sums over the mean-squared error and cross-entropy losses, respectively \\
	$c$ is a parameter controlling the level of $L2$ weight regularization to prevent overfitting\\
	\vspace{3mm}
	evaluate the checkpoint every 1000 training steps to decide if replacing the current best player (neural network weights) for generating next batch of self-play games\\ 
}
\caption{AlphaGo Zero training pseudo code, based on~\citet{Silver-AlphaGo-2017}}
\label{AlphaGoZero}
\end{algorithm}

\subsection*{Alpha Zero}

Alpha Zero~\citep{Silver-AlphaZero-2017} generalizes the learning framework in AlphaGo Zero to 
more domains, in particular, perfect information two-player zero-sum games, with a general reinforcement learning algorithm, learning with no human knowledge except the game rules, and achieves superhuman performance for the games of chess, shogi, and Go. 

\subsection*{Discussions}

AlphaGo Zero is a reinforcement learning algorithm. It is neither supervised learning nor unsupervised learning. 
The game score is a reward signal, not a supervision label. 
Optimizing the loss function $l$ is supervised learning.
However, it performs policy evaluation and policy improvement, as one iteration in generalized policy iteration.

AlphaGo Zero is not only a heuristic search algorithm. AlphaGo Zero follows a generalized policy iteration procedure, in which, heuristic search, in particular, MCTS, plays a critical role, but within the scheme of reinforcement learning generalized policy iteration, as illustrated in the pseudo code in Algorithm~\ref{AlphaGoZero}. MCTS can be viewed as a policy improvement operator.


AlphaGo Zero has attained a superhuman level perfromance. It may confirm that human professionals have developed effective  strategies. However, it does not need to mimic human professional plays. Thus it does not need to predict their moves correctly.

The inputs to AlphaGo Zero include the raw board representation of the position, its history, and the colour to play as 19 $\times$ 19 images; game rules; a game scoring function; invariance of game rules under rotation and reflection, and invariance to colour transposition except for komi. 

AlphaGo Zero utilizes 64 GPU workers and 19 CPU parameter servers for training, around 2000 TPUs for data generation, and 4 TPUs for game playing. The computation cost is probably too formidable for researchers with average computation resources to replicate AlphaGo Zero.
ELF OpenGo is a reimplementation of AlphaGoZero/AlphaZero using ELF~\citep{Tian2017ELF}, at \url{https://facebook.ai/developers/tools/elf}.

AlphaGo Zero requires huge amount of data for training, so it is still a big data issue. However, the data can be generated by self play, with a perfect model or precise game rules.

Due to the perfect model or precise game rules for computer Go, AlphaGo algorithms have their limitations. For example, in healthcare, robotics and self driving problems, it is usually hard to collect a large amount of data, and it is hard or impossible to have a close enough or even perfect model. As such, it is nontrivial to directly apply AlphaGo Zero algorithms to such applications. 

On the other hand, AlphaGo algorithms, especially, the underlying techniques, namely, deep learning, reinforcement learning, Monte Carlo tree search, and self-play, have many applications. 
\citet{Silver-AlphaGo-2016} and~\citet{Silver-AlphaGo-2017} recommend the following applications: 
general game-playing (in particular, video games), classical planning, partially observed planning, scheduling, constraint satisfaction,  robotics, industrial control, and online recommendation systems.
AlphaGo Zero blog at \url{https://deepmind.com/blog/alphago-zero-learning-scratch/} mentions the following structured problems: protein folding, reducing energy consumption, and searching for revolutionary new materials.
\footnote{There is a blog titled "AlphaGo, in context" in May 2017 by Andrej Karpathy, after AlphaGo defeated Ke Jie, 
at\url{https://medium.com/@karpathy/alphago-in-context-c47718cb95a5}. 
The author characterizes properties of Computer Go as: fully deterministic, fully observable, discrete action space, accessible perfect simulator, relatively short episode/game, clear and fast evaluation conducive for many trail-and-errors, and huge datasets of human play games, to illustrate the narrowness of AlphaGo. 
(AlphaGo Zero invalidates the last property, "huge datasets of human play games".)
It is true that computer Go has limitations in the problem setting and thus for potential applications, and is far from artificial general intelligence. 
However, we see the success of AlphaGo, in particular, AlphaGo Zero, as the triumph of AI, in particular, the underlying techniques, i.e., deep learning, reinforcement learning, Monte Carlo tree search, and self-play; these techniques are present in many recent achievements in AI. 
AlphaGo techniques will shed light on classical AI areas, like planning, scheduling, and constraint satisfaction~\citep{Silver-AlphaGo-2016}, and new areas for AI, like retrosynthesis~\citep{Segler2018}. Reportedly, the success of AlphaGo's conquering titanic search space inspired quantum physicists to solve the quantum many-body problem~\citep{Carleo2017}.}
Several of these, like planning, scheduling, constraint satisfaction, are constraint programming problems~\citep{Rossi2006}. 

AlphaGo has made tremendous progress, and sets a landmark in AI. However, we are still far away from attaining artificial general intelligence (AGI).

It is interesting to see how strong a deep neural network in AlphaGo can become, i.e., to approximate optimal value function and policy, and how soon a very strong computer Go program would be available on a mobile phone.

\subsection{Card Games}
\label{games:card}

Variants of card games, including Texas Hold'em Poker, majiang/mahjong, Skat, etc.,  
are imperfect information games, which, as one type of game theory problems, have many applications, e.g., security and medical decision support~\citep{Chen2012}.  It is interesting to see more progress of deep RL in such applications, and the full version of Texas Hold'em.

Heads-up Limit Hold'em Poker is essentially solved~\citep{Bowling2015} with counterfactual regret minimization (CFR), which is  an iterative method to approximate a Nash equilibrium of an extensive-form game with repeated self-play between two regret-minimizing algorithms. 

\subsection*{DeepStack}

Recently, significant progress has been made for Heads-up No-Limit Hold'em Poker~\citep{Moravcik2017}, the DeepStack computer program defeated professional poker players for the first time. DeepStack utilizes the recursive reasoning of CFR to handle information asymmetry, focusing computation on specific situations arising when making decisions and use of value functions trained automatically, with little domain knowledge or human expert games, without  abstraction and offline computation of complete strategies  as before.  
DeepStack follows generalized policy iteration. 
Watch a talk by Michael Bowling at \url{https://vimeo.com/212288252}.

It is desirable to see extension of DeepStack to multi-player settings.
The current study of Poker limits to a single hand, while there are usually many hands in Poker.
It is thus desirable to investigate sequential decision making in cases of multiple hands,
and probably treating tournaments and cash games differently.

\subsection{Video Games}
\label{games:video}

Video games are those with video frames as inputs to RL/AI agents.
In video games, information may be perfect or imperfect, and game theory may be deployed or not.  

We discuss algorithms for Atari 2600 games, in particular, DQN and its extensions, mostly in Chapter~\ref{value}, when we talk about value-based methods.
We discuss algorithms for StarCraft in Chapter~\ref{MARL}, when we talk about multi-agent RL.
We discuss \citet{Zambaldi2018}, which investigated StarCraft II mini-games  with relational deep RL, in Chapter~\ref{relational}.
\citet{SunPeng2018StarCraft} and \citet{Pang2018StarCraft} have beaten full-game built-in AI in StarCraft II.
We discuss \citet{Jaderberg2018Quake}, a human-level agent for Quake III Arena Capture the Flag
in Chapter~\ref{MARL}.
We discuss \citet{Mnih-A3C-2016} in Section~\ref{policy:AC}, and \citet{Jaderberg2017} and \cite{Mirowski2017} in Section~\ref{unsupervised}, which use Labyrinth as the testbed.
\citet{Oh2016} and \cite{Tessler2017} study Minecraft.
\citet{Chen2017} and \citet{Firoiu2017} study Super Smash Bros. 

\citet{Wu2017} deploy A3C to train an agent in a partially observable 3D environment, Doom, from recent four raw frames and game variables, to predict next action and value function, following the curriculum learning~\citep{Bengio-2009-Curriculum} approach of starting with simple tasks and gradually transition to harder ones.  
It is nontrivial to apply A3C to such 3D games directly, partly due to sparse and long term reward. 
The authors won the champion in Track 1 of ViZDoom Competition by a large margin. 
\citet{Dosovitskiy2017} approach the problem of sensorimotor control in immersive environments with supervised learning.
\citet{Lample2017} also study Doom. 

\clearpage


\section{Robotics}
\label{robotics}

Robotics is a classical application area for reinforcement learning. 
Robots have wide applications, e.g., manufacture, supply chain, healthcare, etc.

Robotics pose challenges to reinforcement learning,
including dimensionality, real world examples, 
under-modelling (models not capturing all details of system dynamics) and model uncertainty,
and, reward and goal specification~\citep{Kober2013}. 
RL provides tractable approaches to robotics, 
through representation, including state-action discretization, value function approximation, and pre-structured policies;
through prior knowledge, including demonstration, task structuring, and directing exploration; 
and through models, including mental rehearsal, which deals with simulation bias, real world stochasticity, and optimization efficiency with simulation samples, and approaches for learned forward models~\citep{Kober2013}.


Watch \citet{Abbeel2017NIPStalk} for a talk on deep learning for robotics.
See~\citet{Kober2013} for a survey of RL in robotics, \citet{Deisenroth2013} for a survey on policy search for robotics, and
\citet{Argall2009} for a survey of robot learning from demonstration. 
See the journal Science Robotics.
It is interesting to note that from NIPS 2016 Invited Talk titled Dynamic Legged Robots by Marc Raibert, 
Boston Dynamics robots did not use machine learning.
See NIPS 2018 Workshop on Imitation Learning and its Challenges in Robotics at \url{https://sites.google.com/view/nips18-ilr}.

\citet{Ng2004} study autonomous helicopter.
\citet{Reddy2018} study glider soaring.
\citet{Mirowski2017}, \citet{Banino2018}, and \citet{Wayne2018MERLIN} propose methods for learning to navigate.
\citet{Liu2016CoRobot, Liu2017CoRobot} study how to make robots and people to collaborate to achieve both flexibility and productivity in production lines. 
See a blog at \url{http://bair.berkeley.edu/blog/2017/12/12/corobots/}.

In the following, we discuss 
sim-to-real, 
imitation learning,
value-based learning,
policy-based learning, and
model-based learning for robotics.
Then we discuss autonomous driving vehicles, a special type of robots.

\subsection{Sim-to-Real}
\label{robotics:sim2real}

It is easier to train a robot in simulation than in reality. 
Most RL algorithms are sample intensive.
And exploration may cause risky policies to the robot and/or the environment.
However, a simulator usually can not precisely reflect the reality.
How to bridge the gap between simulation and reality, i.e., sim-to-real, is critical and challenging in robotics.
Sim-to-real is a special type of transfer learning, as discussed in Section~\ref{meta:transfer}.

\citet{Peng2017Sim2Real} propose to use dynamics randomization to train recurrent policies in simulation, and deploy the policies directly on a physical robot, achieving good performance on an object pushing task, without calibration. This work does not consider visual observation.

\citet{OpenAI2018} propose to learn dexterity of in-hand manipulation to perform object reorientation for a physical Shadow Dexterous Hand, using Proximal Policy Optimization (PPO), with dynamics randomization in simulation, and transferring the learned policy directly to physical hand,
sharing code base with OpenAI Five for playing Dota 2.
See a blog with video at \url{https://blog.openai.com/learning-dexterity/}.
\citet{ChenTao2018NIPS}, \citet{Popov2017} and \citet{Perez2017} also study dexterity learning.
See also \url{https://bair.berkeley.edu/blog/2018/08/31/dexterous-manip/}.

\citet{Rusu2017} propose to use progressive networks~\citep{Rusu2016} to bridge the reality gap. 
In progressive networks, lateral connections connect layers of network columns learned previously to each new column,
to support compositionality, and to support transfer learning and domain adaptation.
In a progressive network, columns may not have the same capacity or structure,
so that the column for simulation can have sufficient capacity,
and the column for reality may have lower capacity, 
which can be initialized from the column trained with simulation,  
to encourage exploration and fast learning from scarce real data. 
\citet{Rusu2017} use MuJoCo physics simulator to train the first column, for a reaching task with the modelled Jaco; 
and use real Jaco to train the second column with RGB images, to be deployed on a real robot.
The authors also propose to handle dynamic tasks, e.g., dynamic targets, by adding a third column and proprioceptive features, i.e., features for joint angles and velocities for arms and fingers.

\citet{Sadeghi2018} propose a  convolutional recurrent neural network to learn perception and control for servoing a robot arm to desired objects, invariant to viewpoints. 
The proposed method uses the memory of past movements to select actions to reach the target, 
rather than assuming known dynamics or requiring calibration.  
The authors propose to learn the controller with simulated demonstration trajectories and RL.
The supervised demonstration learning usually learns a myopic policy, with distance to goal as the objective; and RL helps learn a policy consider long term effect, by evaluating the action value function to assess if the goal can be reached.
The visual layers are adapted with a small amount of realistic images for better transfer performance.
The work assumes that the robot can move to an object directly, without planning for a sophisticated policy, e.g., with obstacles.

See also \citet{Bousmalis2017}, and a blog,
\url{https://research.googleblog.com/2017/10/closing-simulation-to-reality-gap-for.html}

\subsection{Imitation Learning}
\label{robotics:imitation}

\citet{Finn2016} study inverse optimal cost, or inverse reinforcement learning in control.
The authors propose to utilize nonlinear cost functions, such as neural networks, 
to impose structures on the cost for informative features and effective regularization,
and approximate MaxEnt~\citep{Ziebart2008} with samples for learning with unknown dynamics in high-dimensional continuous environments.

\citet{Duan2017OneShot} propose one-shot imitation learning, in the setting with many tasks, as supervised learning.
The authors train a neural network with pairs of demonstrations for a subset of tasks,
with input as the first demonstration and a state sampled from the second demonstration,
and predicted the action for the sampled state.
The authors utilize the soft attention to process sequence of states and actions of a demonstration, 
and vector components for block locations, as in the block stacking experiments,
for better generalization to unseen conditions and tasks in training data.
See videos at \url{https://sites.google.com/view/nips2017-one-shot-imitation/}.

\citet{Finn2017One} and \citet{Yu2018One} propose one-shot meta-imitation learning methods 
to build vision-based policies fine-tuned end-to-end from one demonstration, 
using model-agnostic meta-learning (MAML)~\citep{Finn2017MAML} for pre-training
on a diverse range of demonstrations from other environments.
Usually learning from raw pixels requires a large amount of data.
MAML optimizes an initial representation for a learning algorithm,
to build a rich prior in the meta-learning phase,
allowing the parameters to adapt to new tasks with a few examples.
We discuss MAML in Section~\ref{meta:one-shot-learning}.
In \citet{Finn2017One}, demonstrations come from a teleoperated robot.
In \citet{Yu2018One}, demonstrations come from a human, posing a challenging issue of domain shift.
\citet{Yu2018One} learn how to learn from both human and teleoperated demonstrations,
and could adapt to a new task with one human demonstration.
Both \citet{Finn2017One} and \citet{Yu2018One} experiment with both simulation and physical robots.
See a blog at \url{http://bair.berkeley.edu/blog/2018/06/28/daml/}.
The open source for \citet{Finn2017One} is at \url{https://github.com/tianheyu927/mil}.

\subsection{Value-based Learning}

Here we discuss two papers using successor representation (SR).
We introduce that a value function can be decomposed into environment dynamics (SR) and reward signal in Chapter~\ref{representation}.
We discuss general value function (GVF) in Section~\ref{value:GVF}, e.g.,
\citet{Sutton2011},
universal function approximators (UVFAs)~\citep{Schaul2015}, and, 
hindsight experience replay (HER)~\citep{Andrychowicz2017}.
\citet{Sutton2011} and \citet{Andrychowicz2017} experiment with robots.
\citet{Barreto2017} extend successor representation to successor features for continuous tasks with function approximation, as discussed in Section~\ref{meta:transfer}. 

\citet{Zhang2017SF} propose a deep RL approach with successor features for robot navigation tasks,
for transferring knowledge obtained from previous navigation tasks to new ones,
without localization, mapping or planning. 
The authors experiment with both simulation and physical robots.

\citet{Sherstan2018} propose to accelerate construction of knowledge represented by GVFs with successor representation in a continual learning setting, so that learning new GVFs are sped up with previous learned GVFs. 

\subsection{Policy-based Learning}

\citet{Levine2016} propose to train perception and control systems jointly end-to-end, to map raw image observations directly to torques at the robot's motors. 
The authors introduce guided policy search (GPS) to train policies represented by CNNs, by transforming policy search into supervised learning to achieve data efficiency, with training data provided by a trajectory-centric RL method operating under unknown dynamics. 
GPS alternates between trajectory-centric RL and supervised learning, to obtain the training data coming from the policy's own state distribution, to address the issue that supervised learning usually does not achieve good, long-horizon performance. 
GPS utilizes pre-training to reduce the amount of experience data to train visuomotor policies. 
Good performance is achieved on a range of real-world manipulation tasks requiring localization, visual tracking, and handling complex contact dynamics, and simulated comparisons with previous policy search methods. 
As the authors mention, "this is the first method that can train deep visuomotor policies for complex, high-dimensional manipulation skills with direct torque control". 

\citet{Yahya2017} propose a distributed and asynchronous guided policy search 
for a vision-based door opening task with four robots.

\citet{Zhu2017} propose target-driven visual navigation in indoor scenes with deep RL 
by treating the policy as a function of both the goal and the current state in an actor-critic model,
to generalize over targets, similar to general value function.
The authors design the house of interactions (AI2-THOR), a simulation framework with high-quality 3D scenes and physics engine, 
which allows visual interactions for agents, to generate large amount of training data.
The authors qualitatively compare AI2-THOR with other simulators, like
ALE, VizDoom, UETorch, Project Malmo, SceneNet, TORCS, SYTHNIA, and, Virtual KITTI.

\subsection{Model-based Learning}

\citet{Gu2016} propose normalized advantage functions (NAF) to enable experience replay with Q-learning for continuous task, and to refit local linear models iteratively.
\citet{Gu-robotics-2017} propose asynchronous NAF algorithm, with safety constraints, 
for 3D manipulation in simulation and door opening for real robots.

\citet{Finn-robotics-2017} propose to combine model predictive control (MPC) with action-conditioned video prediction, in a self-supervised approach without labeled training data, for physical robotic manipulation of previously unseen objects. 

\citet{Chebotar2017} focus on time-varying linear-Gaussian policies, and integrated a model-based linear quadratic regulator (LQR) algorithm with a model-free path integral policy improvement  algorithm. To generalize the method for arbitrary parameterized policies such as deep neural networks, the authors combined the proposed approach with guided policy search (GPS)~\citep{Levine2016}. 

\citet{Lee2017} study visual servoing by combining features learned from object classification, predictive dynamic models, and fitted Q-iteration algorithm.

\subsection{Autonomous Driving Vehicles}
\label{robotics:autodriving}

Autonomous driving is an important topic of  intelligent transportation systems as we discuss in Chapter~\ref{transportation}. 
\citet{OKelly2018NIPS} propose to test autonomous vehicles rare-event simulation.
\citet{Fridman2018} propose  DeepTraffic, a micro-traffic simulator, for deep RL. 
\citet{Yu2018BDD100K} release BDD100K, a large-scale diverse driving video database.
The data is available at, \url{http://bdd-data.berkeley.edu}.
The blog is at, \url{http://bair.berkeley.edu/blog/2018/05/30/bdd/}.
See also \citet{Bojarski2016}, \citet{Bojarski2017}, \citet{Zhou2017Apple}. 
See a website for Tesla Autopilot Miles at, \url{https://hcai.mit.edu/tesla-autopilot-miles/}.

We argue that the current science, engineering, and technology, including AI, 
are not ready for road test of  fully autonomous driving vehicles yet.
One issue is adversarial examples,
as we discuss in Section~\ref{systems}.
From the Insurance Institute for Highway Safety (IIHS) website, \url{https://www.iihs.org/iihs/topics/t/general-statistics/fatalityfacts/state-by-state-overview},
we learn that human drivers in US encounter  roughly one death  per 100 million vehicle miles.
Consider, for an autonomous system, how many decisions to make, 
and what level of accuracy are required, for just one second.
Simulators are probably far from reality;
and road test data collected so far are far from for justification of statistically significant level claims.
Road tests are actually experiments, with humans involved.
For a "self-driving vehicle" to run on roads, a feasible approach is to require a driver to sit on the driver's seat, 
pay attention to the driving, and take control of the vehicle at any time if not always.
For fully self driving vehicles, we propose to conduct "road tests" in a closed environment, 
using robots as pedestrians, etc.,  
until AI has sufficient intelligence, e.g., understanding scenes, with common sense, etc. 

\clearpage

\section{Natural Language Processing}
\label{NLP}

Natural language processing (NLP) learns, understands, and produces human language content using computational techniques~\citep{Hirschberg2015}.

There are many interesting topics in NLP.
we discuss 
sequence generation in Section~\ref{NLP:textgeneration}, 
machine translation in Section~\ref{NLP:translation}, and,
dialogue systems in Section~\ref{NLP:chatbot}.  
We list some NLP topics in the following. 
We also list the integration of computer vision with NLP, like visual captioning, visual dialog, and 
visual relationship and attribute detection, which we discuss in Section~\ref{CV:NLP}.

\begin{itemize}
\item language tree-structure learning, e.g., \citet{Socher2011, Socher2013, Yogatama2017};
\item semantic parsing, e.g., \citet{Liang2017NSM};
\item question answering, e.g., \citet{Shen2017}, \citet{Trischler2016}, \citet{Xiong2017}, \citet{Wang2017RNet}, \citet{Choi2017}
\item summarization, e.g., \citet{Chopra2016, Paulus2017, Zhang2017sentence}
\item sentiment analysis~\citep{Liu2012, Zhang2018sentiment} 
\item information retrieval~\citep{Manning2008}, e.g., and \citet{Mitra2017}, \citet{Wang2017IRGAN}, \citet{Zhang2016}
\item information extraction, e.g., \citet{Narasimhan2016};
\item visual captioning, e.g., \citet{Wang2018storytelling, Xu2015, Liu2016, Lu2016, Ren2017, Pasunuru2017, Wang2018video}; 
\item visual dialog, e.g., \citet{Das2017, Strub2017}; 
\item visual relationship and attribute detection, e.g., \citet{Liang2017X};
\item visual question answering, e.g., \citet{Hudson2018}
\item popular Reddit threads prediction, e.g., \citet{He2016Reddit}
\item automatic query reformulation, e.g., \citet{Nogueira2017};
\item language to executable program, e.g., \citet{Guu2017};
\item knowledge graph reasoning, e.g., \citet{Xiong2017DeepPath};
\item text games, e.g., \citet{WangSida2016}, \citet{He2016-textgame}, \citet{Narasimhan2015};
\item semi-supervised learning, co-training, e.g., \citet{Wu2018co-training}.
\end{itemize}

Deep learning has been permeating into many subareas in NLP, and helping make significant progress. 
It appears that NLP is still a field more about synergy than competition, for deep learning vs. non-deep learning algorithms, and for approaches based on no domain knowledge vs. with linguistics knowledge.
Some non-deep learning algorithms are effective and perform well, e.g., word2vec~\citep{Mikolov2013, Mikolov2017} and fastText~\citep{Joulin2017}, and many papers study syntax and semantics of languages,
with a recent example in semantic role labeling~\citep{He2017SRL}. 
Some deep learning approaches to NLP problems incorporate explicitly or implicitly linguistics knowledge, e.g., \citet{Socher2011, Socher2013, Yogatama2017}. 
\citet{Manning2017} discusses computational linguistics and deep learning.
See ACL 2018 Workshop on Relevance of Linguistic Structure in Neural NLP,
at \url{https://sites.google.com/view/relsnnlp}.

\citet{McCann2018NLPDecathlon} propose natural language decathlon (decaNLP), 
an NLP benchmark suitable for multitask, transfer, and continual learning.
See the website, \url{http://decanlp.com}.
\citet{Devlin2018BERT} propose Bidirectional Encoder Representations from Transformers (BERT)
for language representation model pre-training.

\citet{Melis2018} investigate the evaluation in neural language models,
and observe that standard LSTM outperforms recent models.
\citet{Bai2018TCN} show empirically that CNNs outperforms RNNs over a wide range of tasks.

See \citet{Jurafsky2017}, \citet{Goldberg2017}, \citet{Deng2018} for books on NLP;
\citet{Hirschberg2015,Cho2015, Young2017} for surveys on NLP;
\citet{Deng2013, Gao2018Neural, Hinton2012, He2013, Young2013} for surveys on dialogue systems;
\citet{Neubig2017} for a tutorial on neural machine translation and sequence-to-sequence models,
\citet{Agarwal2018} for a tutorial on end to-end goal-oriented question answering systems,
and, \citet{Monroe2017} for a gentle introduction to translation.

See three ACL 2018 tutorials:
\citet{Wang2018tutorial} for deep reinforcement learning for NLP,
\citet{Gao2018tutorial} for neural approaches to conversational AI (see also \citet{Gao2018Neural}), and,
\citet{Anderson2018tutorial} for connecting language and vision to actions.

See several workshops: 
NIPS Workshop on Conversational AI: Today's Practice and Tomorrow's Potential,
in 2018 at \url{http://alborz-geramifard.com/workshops/nips18-Conversational-AI/}, 
and, in 2017 at \url{http://alborz-geramifard.com/workshops/nips17-Conversational-AI/};
NIPS 2018 Workshop on Wordplay: Reinforcement and Language Learning in Text-based Games;
NIPS 2016 Workshop on End-to-end Learning for Speech and Audio Processing, 
at \url{https://sites.google.com/site/nips2016endtoendspeechaudio/};
and, 
NIPS 2015 Workshop on Machine Learning for Spoken Language Understanding and Interactions,
at \url{http://slunips2015.wixsite.com/slunips2015}.


\subsection{Sequence Generation}
\label{NLP:textgeneration}

A sequence may take the form of text, music, and molecule, etc.
Sequence generation techniques may be applicable to multiple domains, 
e.g., \citet{Jaques2017} experiment with musical melody and computational molecular generation.
Here we focus on text generation,
which is the basis for many NLP problems, like conversational response generation, machine translation, abstractive summarization, etc.

Text generation models are usually based on $n$-gram, feed-forward neural networks, or recurrent neural networks, trained to predict next word given the previous ground truth words as inputs; then in testing, the trained models are used to generate a sequence word by word, using the generated words as inputs. The errors will accumulate on the way, causing the exposure bias issue. Moreover, these models are trained with word level losses, e.g., cross entropy, to maximize the probability of next word; however, the models are evaluated on different metrics like BLEU.

\citet{Ranzato2016} propose mixed incremental cross-entropy reinforce (MIXER) for sequence prediction, with incremental learning and a loss function combining both REINFORCE and cross-entropy. MIXER is a sequence level training algorithm, aligning training and testing objective, such as BLEU, rather than predicting the next word as in previous papers. 

\citet{Bahdanau2017}  propose an actor-critic algorithm for sequence prediction, to improve \citet{Ranzato2016}. 
The authors utilize a critic network to predict the value of a token, i.e., the expected score following the sequence prediction policy, defined by an actor network, trained by the predicted value of tokens. 
Some techniques are deployed to improve performance: SARSA rather than Monter-Carlo method to lessen the variance in estimating value functions; target network for stability; sampling prediction from a delayed actor whose weights are updated more slowly than the actor to be trained, to avoid the feedback loop when actor and critic need to be trained based on the output of each other; and, reward shaping to avoid the issue of sparse training signal.  

\citet{Yu2017} propose SeqGAN,  sequence generative adversarial nets with policy gradient, integrating the adversarial scheme in \citet{Goodfellow2014}.

\subsection{Machine Translation}
\label{NLP:translation}

Neural machine translation~\citep{Kalchbrenner2013, Cho2014, Sutskever2014, Bahdanau2015} utilizes end-to-end deep learning for machine translation, and becomes dominant, against the traditional statistical machine translation techniques.  
In sequence to sequence model~\citep{Cho2014, Sutskever2014}, an input sequence of symbol representation is encoded to a fix-length vector, which is then decoded to symbols one by one, in an auto-regressive way, using symbols generated previously as additional input. 
\citet{Bahdanau2015} introduce a soft-attention technique to address the issues with encoding the whole sentence into a fix-length vector.

\citet{He2016} propose dual learning mechanism to tackle the data hunger issue in machine translation, inspired by the observation that the information feedback between the primal, translation from language A to language B, and the dual, translation from B to A, can help improve both translation models, with a policy gradient method, using the language model likelihood as the reward signal.  
Experiments show, with only 10\% bilingual data for warm start and monolingual data, the dual learning approach perform comparably with previous neural machine translation methods with full bilingual data in English to French tasks. 
The dual learning mechanism may be extended to other tasks, if a task has a dual form, e.g.,  speech recognition and text to speech, image caption and image generation, question answering and question generation, search and keyword extraction, etc. 
\citet{XiaYingce2018ICML} study model-level dual learning.

See \citet{Wu2016}, \citet{Johnson2017} for Google's Neural Machine Translation System, 
\citet{Gehring2017} for convolutional sequence to sequence learning for fast neural machine translation, 
\citet{Klein2017opennmt} for OpenNMT, an open source neural machine translation system, at \url{http://opennmt.net},  
\citet{Cheng2016} for semi-supervised learning for neural machine translation, 
\citet{Wu2017AdversarialNMT} for adversarial neural machine translation, 
\citet{Vaswani2017} for a new approach for translation replacing ConvNets and RNN with self attention and positional encoding, open source at \url{https://github.com/tensorflow/tensor2tensor},
\citet{Dehghani2018} for an extension of \citet{Vaswani2017} at \url{https://goo.gl/72gvdq},
\citet{Artetxe2018} for an unsupervised approach to machine translation, and,
\citet{Zhang2017NMT} for an open source toolkit for neural machine translation.

\subsection{Dialogue Systems}
\label{NLP:chatbot}

In dialogue systems, conversational agents, or chatbots, human and computer interacts with a natural language. We intentionally remove "spoken" before "dialogue systems" to accommodate both spoken and written language user interface (UI). 
\citet{Jurafsky2017} categorize dialogue systems as task-oriented dialog agents and chatbots; the former are set up to have short conversations to help complete particular tasks; the latter are set up to mimic human-human interactions with extended conversations, sometimes with entertainment value.
As in~\citet{Deng2017AIFrontiers}, there are four categories: social chatbots, infobots (interactive question answering), task completion bots (task-oriented or goal-oriented) and personal assistant bots. We have seen generation one dialogue systems: symbolic rule/template based, and generation two: data driven with (shallow) learning. We are now experiencing generation three: data driven with deep learning, in which reinforcement learning usually plays an important role.  A dialogue system usually include the following modules: (spoken) language understanding, dialogue manager (dialogue state tracker and dialogue policy learning), and a natural language generation~\citep{Young2013}. In task-oriented systems, there is usually a knowledge base to query. A deep learning approach, as usual, attempts to make the learning of the system parameters end-to-end. See~\citet{Deng2017AIFrontiers} for more details.
See a survey paper on applying machine learning to speech recognition~\citep{Deng2013}.

\citet{Dhingra2017} propose KB-InfoBot, a goal-oriented dialogue system for multi-turn information access. KB-InfoBot is trained end-to-end using RL from user feedback with differentiable operations, including those for accessing external knowledge database (KB). In previous work, e.g., \cite{Li2017TC} and \citet{WenTH2017}, a dialogue system accesses real world knowledge from KB by symbolic, SQL-like operations, which is non-differentiable and disables the dialogue system from fully end-to-end trainable. KB-InfoBot achieves the differentiability by inducing a soft posterior distribution over the KB entries to indicate which ones the user is interested in.  The authors design a modified version of the episodic REINFORCE algorithm to explore and learn both the policy to select dialogue acts and the posterior over the  KB entries for correct retrievals.The authors deploy imitation learning from rule-based belief trackers and policy to warm up the system.

\citet{Su2016} propose an on-line learning framework to train the dialogue policy jointly with the reward model via active learning with a Gaussian process model, to tackle the issue that it is unreliable and costly to use explicit user feedback as the reward signal. The authors show empirically that the proposed framework reduces manual data annotations significantly and mitigates noisy user feedback in dialogue policy learning. 

\citet{LiJiwei2016} propose to use deep RL to generate dialogues to model future reward for better informativity, coherence, and ease of answering, to attempt to address the issues in the sequence to sequence models based on \citet{Sutskever2014}: the myopia and misalignment of maximizing the probability of generating a response given the previous dialogue turn, and the infinite loop of repetitive responses. The authors design a reward function to reflect the above desirable properties, and deploy policy gradient to optimize the long term reward.  It would be interesting to investigate the reward model with the approach in~\citet{Su2016} or with inverse RL and imitation learning as discussed in Chapter~\ref{reward}, although \citet{Su2016} mention that such methods are costly, and humans may not act optimally. 

\citet{Tang2018EMNLP} propose subtask discovery for hierarchical dialogue policy learning based on a dynamic programming approach to segmentation, 
extending \citet{Peng2017Dialogue}, which assumes subtasks are defined by experts.
\citet{Williams2017} propose to combine an RNN with domain knowledge to improve data efficiency of dialog training.
\citet{Lewis2017} study end-to-end learning for negotiation dialogues; open source at \url{https://github.com/facebookresearch/end-to-end-negotiator}.
\citet{Zhang2016Speech} study end-to-end speech recognition with CNNs.
\citet{XiongW2017MS} describe Microsoft's conversational speech recognition system in 2017.
\citet{Zhou2018Emotional} propose an emotional chatting machine.
\cite{Li2017TC} present an end-to-end task-completion neural dialogue system with parameters learned by supervised and reinforcement learning. See the open source at \url{http://github.com/MiuLab/TC-Bot}.
See \citet{Serban2018} for a survey of corpora for building dialogue systems.

See more recent papers: \citet{ElAsri2016}, \citet{Bordes2017}, \citet{Chen2016},  \citet{Fatemi2016}, \citet{Kandasamy2017},   \citet{LiJiwei2017human}, \citet{LiJiwei2017Q},  \citet{Lipton2018BBQ}, \citet{Mesnil2015}, \citet{Mo2018}, \citet{Saon2016},   \citet{She2017}, \citet{XiongW2017},   \citet{Zhao2016}. 

\clearpage

\section{Computer Vision}
\label{CV}

Computer vision is about how computers gain understanding from digital images or videos. 
Computer vision has been making rapid progress recently, and deep learning plays an important role.
We discuss briefly recent progress of computer vision below.

\citet{Krizhevsky2012} propose AlexNet, almost halving the error rate of an ImagetNet competition task, 
and ignite this wave of deep learning/AI. 
\citet{He2016-ResNets} propose residual nets (ResNets) to ease the training of very deep neural networks by adding shortcut connections to learn residual functions with reference to the layer inputs.
Fast R-CNN~\citep{Girshick2015}, Faster R-CNN~\citep{Ren2015}, and Mask R-CNN~\citep{He2017Mask} are proposed for image segmentation.
Facebook AI Research (FAIR) open source Detectron for object detection algorithms, 
\url{https://research.fb.com/downloads/detectron/}.

Generative adversarial networks (GANs)~\citep{Goodfellow2014, Goodfellow2017} attracts lots of attention recently.
There are fundamental work to improve the stability of learning GANs, e.g., Wasserstein GAN (WGAN)~\citep{Arjovsky2017WGAN}, \citet{Gulrajani2017}, and Least Squares GANs (LSGANs)~\citep{Mao2016}.
Many proposals in GANs are using computer vision testbeds, 
e.g., CycleGAN~\citep{Zhu2017CycleGAN}, DualGAN~\citep{Yi2017DualGAN}, and \citet{Shrivastava2017}.

For disentangled factor learning, many papers use computer vision testbeds.
\citet{Kingma2014VAE} propose variational autoencoders (VAEs).
\citet{Kulkarni2015DC-IGN} propose  deep convolution inverse graphics network (DC-IGN), which follows a semi-supervised way.
\citet{Chen2016InfoGAN} propose InfoGAN, an information-theoretic extension to the GANs, following an unsupervised way.
\citet{Higgins2017betaVAE} propose $\beta$-VAE to automatically discover interpretable, disentangled, factorised, latent representations from raw images in an unsupervised way. 
When $\beta = 1$,  $\beta$-VAE is the same as VAEs.
\citet{Eslami2016} propose the framework of Attend-Infer-Repeat for efficient inference in structured image models to reason about objects explicitly. 
\citet{Zhou2015} show that object detectors emerge from learning to recognize scenes,  without supervised labels for objects.

Reinforcement learning is an effective tool for many computer vision problems, 
like classification, e.g. \citet{Mnih-attention-2014}, detection, e.g. \citet{Caicedo2015}, captioning, e.g. \citet{Xu2015}, etc.
RL is an important ingredient for interactive perception~\citep{Bohg2017}, where perception and interaction with the environment would be helpful to each other, in tasks like object segmentation, articulation model estimation, object dynamics learning, haptic property estimation, object recognition or categorization, multimodal object model learning, object pose estimation, grasp planning, and manipulation skill learning.
See~\citet{Rhinehart2018} for a tutorial on inverse reinforcement learning for computer vision.

\citet{Malik2018IJCAI}  discusses that there are great achievements in the fields of vision, motor control, and language semantic reasoning, and it is time to investigate them together.  
\citet{Zhang2018Interpretability} survey visual interpretability for deep learning.
Lucid, at \url{https://github.com/tensorflow/lucid}, is an open source for interpretability, implementing feature visualization techniques in \citet{olah2017feature}.
\citet{olah2018} discuss building blocks of interpretability.

In the following,  we discuss 
recognition in Section~\ref{CV:recognition},
motion analysis in Section~\ref{CV:motion},
scene understanding in Section~\ref{CV:scene},
integration with NLP in Section~\ref{CV:NLP},
visual control in Section~\ref{CV:control}, and
interactive perception in Section~\ref{CV:perception}.

We list more topics about applying deep RL to computer vision as follows: 
\citet{Liu2017F} for semantic parsing of large-scale 3D point clouds,  
\citet{Kaba2017} for view planning, which is a set cover problem,
\citet{Cao2017} for face hallucination, i.e., generating a high-resolution face image from a low-resolution input image,
\citet{Brunner2018} for learning to read maps,
\citet{Cubuk2018AutoAugment} for data augmentation for images,
\citet{Bhatti2016} for SLAM-augmented DQN.

\subsection{Recognition}
\label{CV:recognition}

RL can improve efficiency for image classification by focusing only on salient parts.
For visual object localization and detection,  RL can improve efficiency over approaches  with exhaustive spatial hypothesis search and sliding windows, and strike a balance between sampling more regions for better accuracy and stopping the search when sufficient confidence is obtained about the target's location.  

\citet{Mnih-attention-2014} introduce the recurrent attention model (RAM), which we discuss in Chapter~\ref{attention}. 
\citet{Caicedo2015} propose an active detection model for object localization with DQN, by deforming a bounding box with transformation actions to determine the most specific location for target objects. 
\citet{Jie2016} propose a tree-structure RL approach to search for objects sequentially, considering both the current observation and previous search paths, by maximizing the long-term reward associated with localization accuracy over all objects with DQN. 
\citet{Mathe2016} propose to use policy search for visual object detection.
\citet{Kong2017} deploy collaborative multi-agent RL with  inter-agent communication for joint object search.
\citet{Welleck2017} propose a hierarchical visual architecture with an attention mechanism for multi-label image classification.
\citet{Rao2017} propose an attention-aware deep RL method for video face recognition.
\citet{Krull2017} study 6D object pose estimation.

\subsection{Motion Analysis}
\label{CV:motion}

In tracking, an agent needs to follow a moving object. 
\citet{Supancic2017} propose online decision-making process for tracking, formulate it as a partially observable decision-making process (POMDP), and learn policies with deep RL algorithms,
to decide where to look for the object, when to reinitialize, and when to update the appearance model for the object,
where image frames may be ambiguous and computational budget may be constrained.
\citet{Yun2017} also study visual tracking with deep RL.

\citet{Rhinehart2017}  propose to discover agent rewards for K-futures online (DARKO) to model and forecast first-person camera wearer's long-term goals, together with states, transitions, and rewards from streaming data, with inverse RL.

\subsection{Scene Understanding}
\label{CV:scene}

\citet{Wu2017De-rendering} study the problem of scene understanding, and attempt to obtain a compact, expressive, and interpretable representation to encode scene information like objects, their categories, poses, positions, etc, in a semi-supervised way. In contrast to encoder-decoder based neural architectures as in previous work, \citet{Wu2017De-rendering} propose to replace the decoder with a deterministic rendering function, to map a structured and disentangled scene description, scene XML, to an image; consequently, the encoder transforms an image to the scene XML  by inverting the rendering operation, a.k.a., de-rendering. The authors deploy a variant of REINFORCE to overcome the non-differentiability issue of graphics rendering engines.

\citet{Wu2017De-animation} propose a paradigm with three major components, a convolutional perception module, a physics engine, and a graphics engine, to understand physical scenes without human annotations. The perception module recovers a physical world representation by inverting the graphics engine, inferring the physical object state for each segment proposal in input and combining them. The generative physics and graphics engines then run forward with the world representation to reconstruct the visual data. 
The authors show results on both neural, differentiable and more mature but non-differentiable physics engines.

We discuss generative query network (GQN)~\citep{Eslami2018GQN} in Chapter~\ref{unsupervised}.
\citet{Chen2018} propose a framework for iterative visual reasoning, which we discuss in Chapter~\ref{relational}.
There are recent papers about physics learning, e.g.,~\cite{Agrawal2016Poke, Battaglia2016,  Denil2017, Watters2017, Wu2015Galileo}.

\subsection{Integration with NLP}
\label{CV:NLP}

Some papers integrate computer vision with natural language processing (NLP). 
\citet{Xu2015} integrate attention to image captioning.
See also \citet{Liu2016}, \citet{Lu2016}, \citet{Rennie2017}, and \citet{Ren2017} for image captioning. 
See \citet{Pasunuru2017, Wang2018video} for video captioning.
\citet{Strub2017} propose end-to-end optimization with deep RL for goal-driven and visually grounded dialogue systems for the GuessWhat?! game. \citet{Das2017} propose to learn cooperative visual dialog agents with deep RL. 
\citet{Wang2018storytelling} propose to use inverse RL for visual storytelling.
See also~\citet{Kottur2017}. 
See \citet{Liang2017X} for visual relationship and attribute detection.

\subsection{Visual Control}
\label{CV:control}

Visual control is about deriving a policy from visual inputs, e.g., in games~\citep{Mnih-DQN-2015, Silver-AlphaGo-2016, Silver-AlphaGo-2017, Oh2015, Wu2017, Dosovitskiy2017, Lample2017, Jaderberg2017}, robotics~\citep{Finn-robotics-2017, Gupta2017CMP, Lee2017, Levine2016, Mirowski2017, Zhu2017}, and self-driving vehicles~\citep{Bojarski2016, Bojarski2017, Zhou2017Apple}. 

For a visual control problem in computer vision, there should be some ingredients of, by, for computer vision, but not just use a CNN or some deep neural network to take image or video as input, without further handling with computer vision techniques, e.g., DQN~\citep{Mnih-DQN-2015} and AlphaGo~\citep{Silver-AlphaGo-2016, Silver-AlphaGo-2017}.

\subsection{Interactive Perception}
\label{CV:perception}

Reinforcement learning is an important ingredient for interactive perception~\citep{Bohg2017}.
\citet{Jayaraman2018} propose a deep RL approach with recurrent neural network for active visual completion, to hallucinate unobserved parts based on a small number of observations. 
The authors attempt to answer the question of how to make decisions about what to observe to acquire more information in visual perception,
without labeled data, rather than making inference decisions based on labeled observations. 
The look-around decisions are rewarded based on the accuracy of the predictions of unobserved views,
in particular, the distance between view predictions and their ground truth for all viewpoints and all time steps.
The authors propose a task agnostic approach for active visual completion, 
and, consider two tasks: panoramic natural scenes and 3D object shapes, for illustration.
The authors also discuss generalization and transferability of their proposed approach to new tasks and environments.

\clearpage
 
\section{Finance and Business Management}
\label{fin}

Machine learning naturally has wide applications in finance, e.g., in fundamental analysis, behavioural finance, technical analysis, financial engineering, financial technology (FinTech), etc. Reinforcement learning is a natural solution to some sequential decision making finance problems, like option pricing, trading, and multi-period portfolio optimization, etc. RL also has many applications in business management, like ads, recommendation, customer management, and marketing, etc. 

Financial engineering is a discipline rooting in finance, computer science and mathematics~\citep{Hull06,Luenberger97}.  Derivative pricing is an essential issue in financial engineering. The values of financial derivatives depend on the values of underlying assets. Options are the most fundamental derivatives. An option gives the holder the right, but not the obligation, to buy or sell an asset at a certain price by a certain time. Portfolio optimization is about how to allocate assets so as to trade off between return and risk.  

There are two schools in finance, Efficient Markets Hypothesis (EMH) and Behavioral Finance~\citep{Lo04}. According to EMH, ``prices fully reflect all available information'' and are determined by the market equilibrium. However, psychologists and economists have found a number of behavioral biases that are native in human decision-making under uncertainty. For example, Amos Tversky and Daniel Kahneman demonstrate the phenomenon of loss aversion, in which people tend to strongly prefer avoiding losses to acquiring gains~\citep{Kahneman2011}. \citet{Prashanth2016} investigate prospect theory with reinforcement learning. Behavioral finance justifies technical analysis \citep{Murphy1999} to some extend. 
\citet{Lo00} propose to use nonparametric kernel regression for automatic  technical pattern recognition.
\citet{Lo04} proposes the Adaptive Market Hypothesis to reconcile EMH and behavioral finance, where the markets are in the evolutionary process of competition, mutation, reproduction and natural selection. RL may play an important role in this fundamental market paradigm. 

Financial technology (FinTech) has been attracting lots of attention, especially after the notion of big data and data science. 
FinTech employs machine learning techniques to deal with issues like fraud detection~\citep{Phua2010},  and consumer credit risk~\citep{Khandani2010}, etc. 

We will discuss applications of deep learning and reinforcement learning to finance and business management. 
We only pick a couple of papers for discussions, and do not include many relevant papers.
Machine learning techniques, like support vector machines (SVM), decision trees, etc, have also been applied to finance and business management. 
We can check them from the reference in the papers we discuss. 
 
It is nontrivial for finance and economics academia to accept machine learning methods like neural networks.
One factor is that neural networks, among many machine learning methods, are black box; however,  interpretability is desirable in finance. 
\citet{Doshi-Velez2017} and \citet{Lipton2018ACM} discuss issues of interpretability. 
\citet{Zhang2018Interpretability} is a survey about interpretability in computer vision.
National Bereau Economic Research (NBER) organizes a meeting on Economics of Artificial Intelligence;  
see \url{http://conference.nber.org/conferences/2018/AIf18/summary.html}.
See \citet{Mullainathan2017} for a lecture in  American Finance Association (AFA) 2017 annual meeting about machine learning and prediction in economics and finance.
In 2018 ASSA Annual Meeting, at \url{https://www.aeaweb.org/conference/2018}, AFA as part of it, we see many sessions with the keywords "artificial intelligence" and/or "machine learning".
We see this as that AI and machine learning are starting to permeate to the mainstream of the field of finance and economics.
It would be natural for economics and finance to marry reinforcement learning, machine learning, and AI,
considering that quantitative approaches for economics and finance share foundations of optimization, statistics, and probability with RL/ML/AI,
and, behavioural approaches for economics and finance share foundations of psychology, neuroscience, and cognitive science  with RL/ML/AI.
We will see more and more applications of RL/ML/AI in finance, economics, and social sciences in general.
See NIPS 2018 Workshop on Challenges and Opportunities for AI in Financial Services: the Impact of Fairness, Explainability, Accuracy, and Privacy.

Before discussing applications of RL to finance and business management, we introduce finance applications with deep learning. Deep learning has a wide applications in finance, e.g., company fundamentals prediction~\citep{Alberg2017}, macroeconomic indicator forecasting~\citep{Cook2017}, and limit order books~\citep{Sirignano2016}, etc.

\citet{Heaton2016} introduce deep learning for finance, in particular, deep portfolios, 
and argue that deep learning methods may be more powerful than the current standard methods in finance., 
e.g., the simplistic traditional financial economics linear factor models and statistical arbitrage asset management techniques. 
The authors show the power of deep learning with a case study of smart indexing for the biotechnology IBB index with a four step algorithm: 1) auto-encoding, find the market-map to auto-encode the input variables with itself and to create a more efficient representation of the input variables; 2) calibrating, find the portfolio-map to create a portfolio from the input variables to approximate an objective; 3) validating, balance the errors in the auto-encoding and calibrating steps; and 4) verifying, choose market-map and portfolio-map to satisfy the validating step. 
The authors make an interesting observation that the univariate activation functions such as ReLU, i.e., $max(0,x)$, where $x$ is a variable, in deep learning can be interpreted as compositions of financial call and put options on linear combination of input assets.   

\citet{Bao2017} investigate the problem of stock price forecasting by combining wavelet transforms (WT), stacked auto-encoders (SAEs) and long-short term memory (LSTM): WT for decomposing stock price time series to reduce noises, SAEs for generating high-level features for stock price prediction, and LSTM for stock price forecasting by taking the denoised features. 
The authors evaluate the performance of the proposed method with six market indices and their corresponding index futures, together with a buy-and-sell trading strategy, using three performance metrics: Mean absolute percentage error (MAPE), correlation coefficient (R) and Theil's inequality coefficient (Theil U), and show promising results in both predictive accuracy and profitability performance.

\subsection{Option Pricing}

Options are fundamental financial instruments, dating back to the ancient time. A challenging problem is option pricing, especially for American type options, which can be exercised before the maturity date. For European options, which can only be exercised at the maturity date, prices can be calculated by the Black-Scholes formula in certain cases.
The key to American option pricing~\citep{Longstaff01, Tsitsiklis01, Li2009Option} is to calculate the conditional expected value of continuation.
This is an optimal stopping problem. 
\citet{Hull06} provides an introduction to options and other derivatives and their pricing methods; and \citet{Glasserman04} provides a book length treatment for Monte Carlo methods in financial engineering.
The least squares Monte Carlo (LSM) method in \citet{Longstaff01},
following approximate dynamic programming,
is a standard approach in the finance literature for pricing American options. 


\subsection{Portfolio Optimization}

Mean-variance analysis by Markowitz is a classical approach to portfolio optimization in one period~\citep{Luenberger97}. Dynamic portfolio optimization in multi-period renews its attraction recently \citep{Campbell02,Brandt05}, following the recent empirical evidence of return predictability \citep{Pastor09}, and with the consideration of practical issues including parameter and model uncertainty, transaction cost and background risks~\citep{Brandt05}. \citet{Brandt05} and \citet{Neuneier1997} deploy the backward dynamic programming approach in \citet{Longstaff01} for the dynamic portfolio problem. It is possible to apply reinforcement learning methods for it. 

\citet{Moody01} learns to trade via direct reinforcement, without any forecasting.
\citet{Deng2016} extend it with deep neural networks. It may be beneficial to take advantage of return predictability in RL methods.

It is critical to control the risk when forming portfolios. Value-at-Risk (VaR) is a popular risk measure; while conditional VaR (CVaR) has desirable mathematical properties~\citep{Hull06}.  
\citet{Yu2009} provide formulations for VaR and CVaR with relaxed probability distributions by worst-case analysis. Deep (reinforcement) learning would provide better solutions in some issues in risk management. The generalization to continuous state and action spaces is an indispensable step for such methods to be applied to dynamic portfolio optimization.

 
\subsection{Business Management}
\label{business}

\citet{Li2010} formulate personalized news articles recommendation as a contextual bandit problem, 
to learn an algorithm to select articles sequentially for users based on contextual information of users and articles, 
such as historical activities of users and descriptive information and categories of content, 
and to take user-click feedback to adapt selection policy to maximize total user clicks in the long run. 

\citet{Theocharous2015} formulate a personalized ads recommendation systems as a RL problem to maximize life-time value (LTV) with theoretical guarantees. This is in contrast to a myopic solution with supervised learning or contextual bandit formulation, usually with the performance metric of click through rate (CTR). As the models are hard to learn, the authors deploy a model-free approach to compute a lower-bound on the expected return of a policy to address the off-policy evaluation problem, i.e., how to evaluate a RL policy without deployment. 

\citet{Jiang2016Doubly} study off-policy value evaluation by extending the doubly robust estimator for bandits. 
The proposed method helps safety in policy improvements and applies to both shallow and deep RL. 
One experiment is about maximizing lifetime value of customers.
\citet{Silver2013} propose concurrent reinforcement learning for the customer interaction problem. 
See
\citet{Cai2018AAAI} for mechanism design for fraudulent behaviour in e-commerce,
\citet{Hu2018ranking} for ranking in e-commerce search engine,
\citet{HuZehong2018NIPS} for incentive mechanism design in crowdsourcing, 
\citet{Lattimore2018NIPS} for ranking,
\citet{Nazari2018NIPS} for vehicle routing in operations research,
\citet{Shi2018} for visualization of online retail environment for RL,
and,
\citet{Zhao2018RecSys}, \citet{Zhao2018KDD} and \citet{Zheng2018WWW} for recommendation.
See \citet{Zhang2017Recommender} for a survey on recommendation.

See Section 16.7 Personalized Web Services in \citet{Sutton2018} for a  detailed and intuitive description of some topics discussed here. 

\clearpage 
\section{More Applications}
\label{more-apps}

In this chapter, we discuss more reinforcement learning applications: 
healthcare in Section~\ref{healthcare}, 
education in Section~\ref{education}. 
energy  in Section~\ref{energy},
transportation in Section~\ref{transportation},
computer systems in Section~\ref{systems},
and, science, engineering and art in Section~\ref{science}. 
\subsection{Healthcare}
\label{healthcare}

There are many opportunities and challenges in healthcare for machine learning \citep{Miotto2017, Saria2014}.
Personalized medicine is getting popular in healthcare. 
It  systematically optimizes patients' health care, in particular, for chronic conditions and cancers using individual patient information, potentially from electronic health/medical record (EHR/EMR).  
\citet{LiYuan2018NIPS} propose a hybrid retrieval-generation reinforced agent for medical image report generation.
\citet{Rajkomar2018} investigate applying deep learning to EHR data.
\citet{Rotmensch2017} learn a healthcare knowledge graph from EHR.
\citet{Fauw2018} apply deep learning for diagnosis and referral in retinal disease;
see a blog at \url{https://deepmind.com/blog/moorfields-major-milestone/}.
\citet{Gheiratmand2017} study network-based patterns of schizophrenia.
\citet{Rajpurkar2018} introduce a large dataset of musculoskeletal radiographs.
See a tutorial on deep reinforcement learning for medical imaging at \url{https://www.hvnguyen.com/deepreinforcementlearning}.
See \citet{Liu2017ICML} for a tutorial on deep learning for health care applications.
See a course on Machine Learning for Healthcare, at \url{https://mlhc17mit.github.io}.

Dynamic treatment regimes (DTRs) or adaptive treatment strategies are sequential decision making problems. Some issues in DTRs are not in standard RL. 
\citet{Shortreed2011} tackle the missing data problem,  and design methods to quantify the evidence of the learned optimal policy. 
\citet{Goldberg2012} propose methods for censored data (patients may drop out during the trial) and flexible number of stages. 
See \citet{Chakraborty2014} for a recent survey, and \citet{Kosorok2015} for an edited book about recent progress in DTRs. Currently Q-learning is the RL method in DTRs. 
\citet{Ling2017} apply deep RL to the problem of inferring patient phenotypes.
\citet{LiuYao2018NIPS} study off-policy policy evaluation and its application to sepsis treatment.
\citet{KallusZhou2018NIPS} study confounding-robust policy improvement and its application to acute ischaemic stroke treatment.
\citet{PengYuShao2018NIPS} study disease diagnosis.

Some recent conferences and workshops at the intersection of machine learning and healthcare are: 
Machine Learning for Healthcare, \url{https://www.mlforhc.org};
NIPS 2018 Workshop on Machine Learning for Health (ML4H): Moving beyond supervised learning in healthcare;
NIPS 2017 Workshop on Machine Learning for Health (ML4H), \url{https://ml4health.github.io/2017/};
NIPS 2016 Workshop on Machine Learning for Health (ML4H), \url{http://www.nipsml4hc.ws}; 
NIPS 2015 Workshop on Machine Learning in Healthcare, \url{https://sites.google.com/site/nipsmlhc15/}. 
See an issue of Nature Biomedical Engineering on machine learning at \url{https://www.nature.com/natbiomedeng/volumes/2/issues/10}.
See \citet{Hernandez2018}, a WSJ article about IBM Watson dilemma.


\subsection{Education}
\label{education}

\citet{Northcutt2017} presents tutorial-style slides for  AI in online education with an emphasis on personalization.
The author presents a framework for AI in online education, including the active/passive course, content, and student, and give examples as below.  Active AI refers to changes in course or experience; passive AI refers to unseen estimation or modeling.

\begin{itemize}
\item active student: cognitive tutor 
\item passive student: proficiency estimation (IRT)
\item active content: content recommendation engine
\item passive content: estimating points of confusion in instructional videos
\item active course: auto-generate new courses from pieces of other courses, students interact; measure outcomes, and iterate
\item passive course:  estimate optimal course prerequisite structure for a new field
\end{itemize}

\citet{Northcutt2017} present 18 problems, some with solutions, some with ideas. 

One problem is representation, which is about 
how to represent courses as vectors, 
how to measure similarity of two courses, videos, and students,
how to recommend content to students, and 
how to match students with other students, etc.

One approach to obtain representation is to use embeddings, on courses, content, or students.
High-dimensional feature matrices can be used to capture the interactions between courses, content, and students. 
Dimension reduction can be done using PCA, SVD, etc. or hidden layers of a neural network.
With such embedded dense low-dimensional representations, we can
generate new courses from existing courses, pair student, make inference, and structure content, etc.

We can employ a recommendation engine for MOOCs using embeddings,
with a Siamese neural network architecture,
one network to represent students, and another network for content, 
and produce representations so that  $cosine(student, content)$ measures the goodness of $content$ for $student$.


\citet{Northcutt2017} also discuss the following problems, 
detect struggling, rampant cheating, stats collaboration, how to personalize, independent treatment effect (ITE)/counterfactuals, trajectory prediction, how to order content, adaptive learning, Google Scholar, majority bias in forums, feature extraction, cognitive state, content likability, points of confusion, cognitive modeling, human intelligence vs. artificial intelligence, and the next edX.
See~\citep{Northcutt2017} for more details.

There are some recent work in education using RL. 

\citet{Mandel2014} propose an offline policy evaluation method, by combining importance sampling with cross-validation, to investigate generalization of representations. 
The authors propose a feature compaction algorithm for high-dimension problems, 
which benefit from both PCA and neural networks. 
Furthermore, the authors apply the method to an educational game, optimizing engagement by learning concept selection.

\citet{Liu2014Education} propose UCB-Explore, based on multi-armed bandit algorithm UCB1, to automaticaly allocate experimental samples, to balance between learning the effectiveness of each experimental condition and users' test performances, by explicitly specifying the tradeoff between these two objectives. The authors compare UCB-Explore with other multi-armed bandit algorithms like UCB1 and $\epsilon$-greedy with simulation on an educational game.

\citet{Upadhyay2018NIPS} apply reinforcement learning to marked temporal point processes with an application in personalized education.
See also \citet{LiShuang2018NIPS} for applying RL to temporal point processes.
\citet{Oudeyer2016} discuss theory and applications of intrinsic motivation, curiosity, and learning in educational technologies.

Watch a video titled Reinforcement Learning with People~\citep{Brunskill2017}.
See ACL 2018 Workshop on Natural Language Processing Techniques for Educational Applications (NLPTEA 2018) with a Shared Task for Chinese Grammatical Error Diagnosis (CGED), at \url{https://sites.google.com/view/nlptea2018/}. 

\subsection{Energy}
\label{energy}

A smart grid is a power grid utilizing modern information technologies to create an intelligent electricity delivery network for electricity generation, transmission, distribution, consumption, and control~\citep{Fang2012}. An important aspect is adaptive control~\citep{Anderson2011}. \citet{Glavic2017} review application of RL for electric power system decision and control. 
See \citet{Platt2017} for a talk about energy.
Here we briefly discuss demand response~\citep{Wen2015, Ruelens2016}.

Demand response systems motivate users to dynamically adapt electrical demands in response to changes in grid signals, like electricity price, temperature, and weather, etc. With suitable electricity prices, load of peak consumption may be rescheduled/lessened, to improve efficiency, reduce costs, and reduce risks.  
\citet{Wen2015} propose to design a fully automated energy management system with model-free reinforcement learning, so that it doesn't need to specify a disutility function to model users' dissatisfaction with job rescheduling. 
The authors decompose the RL formulation over devices, so that the computational complexity grows linearly with the number of devices, and conduct simulations using Q-learning.   
\citet{Ruelens2016} tackle the demand response problem with batch RL. 
\citet{Wen2015} take the exogenous prices as states, 
and \citet{Ruelens2016} utilize the average as feature extractor to construct states.


\subsection{Transportation}
\label{transportation}

Intelligent transportation systems~\citep{Bazzan2014} apply advanced information technologies for tackling issues in transport networks, like congestion, safety, efficiency, etc., to make transport networks, vehicles and users smart. 

Autonomous driving vehicles is an important topic of  intelligent transportation systems. We discuss it in Section~\ref{robotics:autodriving}. 

See NIPS Workshops on Machine Learning for Intelligent Transportation Systems,
in 2018 at \url{https://sites.google.com/site/nips2018mlits/},
in 2017 at \url{https://sites.google.com/site/nips2017mlits/}, 
and, in 2016 at \url{https://sites.google.com/site/nips2016intelligenttrans/}. 

\subsection*{Adaptive Control}
An important issue in intelligent transportation systems is adaptive control. 
\citet{El-Tantawy2013} propose to model the adaptive traffic signal control problem as a multiple player stochastic game, 
and  solve it with the approach of multi-agent RL~\citep{Shoham2007, Busoniu2008}. 
Multi-agent RL integrates single agent RL with game theory, facing challenges of stability, nonstationarity, and curse of dimensionality. 
\citet{El-Tantawy2013} approach the issue of coordination by considering agents at neighbouring intersections. 
The authors validate their proposed approach with simulations, and real traffic data from the City of Toronto. 
\citet{El-Tantawy2013} don't explore function approximation. 
See \citet{vanDerPol2017} for a recent work, 
and \citet{Mannion2016} for an experimental review, about applying RL to adaptive traffic signal control. 
\citet{Belletti2018} study expert level control of ramp metering based on multi-task deep reinforcement learning.

\subsection{Computer Systems}
\label{systems}

Computer systems are indispensable in our daily life and work, e.g., mobile phones, computers, and cloud computing. Control and optimization problems abound in computer systems, e,g., \citet{Mestres2017} propose knowledge-defined networks, \citet{Gavrilovska2013} review learning and reasoning techniques in cognitive radio networks, and \citet{Haykin2005} discuss issues in cognitive radio, like channel state prediction and resource allocation.  
We also note that Internet of Things (IoT)\citep{Xu2014} and wireless sensor networks~\citep{Alsheikh2014} play important roles in robotics and autonomous driving as discussed in Chapter~\ref{robotics}, in energy systems as discussed in Section~\ref{energy}, and in transportation as discussed in Section~\ref{transportation}. 
See \citet{Zhang2018survey} for a recent survey about applying deep learning and reinforcement learning to issues in mobile and wireless networking.
\citet{Mukwevho2018} discuss fault tolerance in cloud computing.

\citet{Kraska2018} propose learned indexes, by treating B-Tree, Hash, BitMap, etc. as models, 
and use neural networks to learn such models, 
by using the signal of learned model of structure or sort order of lookup keys to predict the existence or position of records. Experiments show promising results.
\citet{Wang2018Compiler} study compiler optimization.
\citet{Reichstaller2017} study software testing. 
\citet{Krishnan2018} study SQL join queries optimization.
\citet{Faust2018Sorting} study sorting.
See recent papers about neural approaches for program synthesis, e.g.,
\citet{Balog2017, Liang2017, LiangChen2018NIPS, Nachum2017, Parisotto2017, Reed2016, Vinyals2015, Zaremba2015, ZhangLisa2018NIPS}.

See SysML conference, at the intersection of system and machine learning, at \url{https://www.sysml.cc}.
See NIPS 2018 Workshop on Security in Machine Learning.

\subsubsection*{Performance Optimization}

\citet{Mirhoseini2017} propose to optimize device placement for Tensorflow computational graphs with RL. 
The authors deploy a seuqence-to-sequence model to predict how to place subsets of operations in a Tensorflow graph on available devices, using the execution time of the predicted placement as reward signal for REINFORCE algorithm.
The proposed method finds placements of Tensorflow operations on devices for Inception-V3, recurrent neural language model  and neural machine translation, yielding shorter execution time than those placements designed by human experts. 
Computation burden is one concern for a RL approach to search directly in the solution space of a combinatorial problem.
We discuss learning combinatorial optimization in Section~\ref{meta:learn-combinatorial}.
\citet{GaoYuanxiang2018NIPS} also study the problem of device placement.

\citet{Mao2016RM} study resource management in systems and networking with deep RL. 
The authors propose to tackle multi-resource cluster scheduling with policy gradient, in an online manner with dynamic job arrivals, optimizing various objectives like average job slowdown or completion time.
The authors validate their proposed approach with simulation.

\citet{Liu2017} propose a hierarchical framework to tackle resource allocation and power management in cloud computing with deep RL. The authors decompose the problem as a global tier for virtual machines resource allocation and a local tier for servers power management. The authors validate their proposed approach with actual Google cluster traces. Such hierarchical framework/decomposition approach was to reduce state/action space, and to enable distributed operation of power management. 

Google deploy machine learning for data centre power management, reducing energy consumption by 40\%. 
See blogs at
\url{http://goo.gl/4PHcos} and \url{http://goo.gl/N3Aoxm}.
\citet{Lazic2018NIPS} study data center cooling with model-predictive control (MPC).

Optimizing memory control is discussed in \citet{Sutton2018}.

\subsubsection*{Security}

There is a long history applying machine learning (ML) techniques to system security issues, e.g.,  
\citet{Chandola2009} survey ML techniques for anomaly detection,
and, \citet{Sommer2010} discuss issues in using ML techniques for network intrusion detection.

Adversarial machine learning is about learning in the presence of adversaries.
It is concerned with the training stage, when facing data poisoning issues, and learning wrong models hard to detect.
It is also concerned with the inference stage, when facing adversarial examples, and making wrong decisions.
Adversarial ML is a critical for some ML applications, like autonomous driving.

Adversarial ML is an emerging field, in this wave of deep learning, 
after researchers find adversarial examples to deep learning algorithms,
e.g., \citet{Szegedy2013} show that various images, like a truck, a building, or a dog,  after being added small noises, 
are all classified by AlexNet as "ostrich, Struthio camelus".
\citet{Goodfellow2015} also show a fast adversarial example generation method, so that an image of panda, after being added a small vector, is classified as a gibbon by GoogLeNet. 
\citet{Eykholt2018} show that physical images, like stop signs, yield signs, left turn signs etc., 
after being perturbed by adding black or white stickers, are misclassified by the state of art deep neural networks as speed limit 45 signs. 
\citet{Evtimov2017} discuss attacks to physical images, \url{http://bair.berkeley.edu/blog/2017/12/30/yolo-attack/}.
RL algorithms are also vulnerable to adversarial attacks, e.g., \citet{Huang2017} and \citet{Havens2018NIPS},
as well as multi-armed bandit algorithms, e.g., \citet{JunKwangSung2018NIPS}. See a blog at \url{http://rll.berkeley.edu/adversarial/}.

Adversarial ML is an active area, with fierce competition between the design of attack and defense algorithms.
\citet{Athalye2018} show that seven of nine defense techniques, shortly after their papers being accepted to ICLR 2018, cause the issue of obfuscated gradients and are vulnerable to their attacks.
See their open source at \url{https://github.com/anishathalye/obfuscated-gradients}
for implementations of their attack and the studied defense methods.


\citet{Anderson2018} propose a black-box attack approach against static portable executable (PE) anti-malware engines with reinforcement learning, which produces functional evasive malware samples to help improve anti-malware engines. The performance still needs improvements. 
See the open source at \url{https://github.com/endgameinc/gym-malware}.

See \citet{Song2018} for a tutorial on AI and security.
See \citet{Kantarcioglu2016} for a tutorial on adversarial data mining.
See \citet{Yuan2017} for a survey on attacks and defenses for deep learning.
\citet{Papernot2016cleverhans} present CleverHans, a software library for reference implementations adversarial ML algorithms.




\subsection{Science, Engineering and Art}
\label{science}

Reinforcement learning, deep learning, machine learning, and AI in general, have very wide interactions with science, engineering and art.
We see that RL and areas in science, engineering and art influence each other,
i.e., RL/AI has applications in these areas, with new observations or even new algorithms and new principles;
and intuitions and principles from these areas help further development of RL/AI.
For example, \citet{Sutton2018} discuss the interplay between RL and neuroscience and psychology; 
and in Chapter~\ref{meta}, we discuss learning to learn new algorithms. 
Here we focus on applications of RL/AI to these areas.   

\citet{Sutton2018} treat dynamic programming (DP) and Markov decision processes (MDPs) as foundations for RL, 
and also devote two chapters for neuroscience and psychology, respectively.
There are books discussing approximate dynamic programming, MDPs, operations research, optimal control, 
as well as the underlying optimization,  statistics, and probability, 
e.g.,\citet{Bertsekas96}, \citet{Bertsekas12}, \citet{Szepesvari2010}, and \citet{Powell11}.
There are strong relationships between these areas with RL.
\footnote{Check for a special issue of IEEE Transactions on Neural Networks and Learning Systems on Deep Reinforcement Learning and Adaptive Dynamic Programming, published in June 2018, \url{https://ieeexplore.ieee.org/document/8353782/}.}
\citet{Powell2010} discusses merging AI and operations research to solve high-dimensional stochastic optimization problems with approximate dynamic programming.
\citet{Lake2016} discuss incorporating human intelligence into the current DL/RL/AI systems.
\citet{Hassabis2017} discuss the connection between neuroscience and RL/AI.
\citet{Kriegeskorte2018} surveys cognitive computational neuroscience.

RL/AI is relevant to many areas,
e.g., mathematics, 
chemistry, 
physics~\citep{Cranmer2016},
biology~\citep{Mahmud2018}, 
music, drawing~\citep{Xie2012, Ha2018}, character animation~\citep{Peng2018, Peng2018SFV}, dancing~\citep{Chan2018dancing}, storytelling~\citep{Thue2007}, etc.
\citet{DeVries2018} study earthquakes with deep learning. 
Some topics, e.g., music,  drawing, storytelling, are at the intersection of science and art.

We discuss
games in Chapter~\ref{games}, 
robotics in Chapter~\ref{robotics},
computer vision in Chapter~\ref{CV},
natural language processing (NLP) in Chapter~\ref{NLP},
and, computer systems in Section~\ref{systems},
as areas in computer science.
We put many topics in computer science like 
indexing~\citep{Kraska2018},
compiler optimization~\citep{Wang2018Compiler},
software testing~\citep{Reichstaller2017}, 
SQL join queries optimization~\citep{Krishnan2018}, and
sorting~\citep{Faust2018Sorting} etc. in computer systems in Section~\ref{systems}.
See recent papers about neural approaches for program synthesis, e.g.,
\citet{Balog2017, Liang2017, LiangChen2018NIPS, Nachum2017, Parisotto2017, Reed2016, Vinyals2015, Zaremba2015, ZhangLisa2018NIPS}.

We discuss
finance and business management in Section~\ref{fin},
healthcare in Section~\ref{healthcare},
and, education in Section~\ref{education},
as areas in social science.
We discuss 
energy in Section~\ref{energy},
and transportation  in Section~\ref{transportation},
as areas in engineering.
\footnote{Computer vision and computer systems are also in engineering.}

Imagination is critical for creative activities, like science, engineering and art.
\citet{Mahadevan2018} discuss imagination machines as a new challenge for AI.
See \citet{Mahadevan2018tutorial} for a tutorial on this topic.

Quantum machine learning is about designing machine learning algorithms on quantum computing architectures,
at the interaction of theoretical computer science, machine learning, and physics.
\citet{Biamonte2017} survey quantum machine learning, including quantum reinforcement learning.

We list some workshops in the following.
\begin{itemize}
\item NIPS 2015 Workshop on Quantum Machine Learning at \url{https://www.microsoft.com/en-us/research/event/quantum-machine-learning/}
\item Machine Learning for Science  Workshop at LBNL at \url{https://sites.google.com/lbl.gov/ml4sci/}
\item Machine Learning for Physics and the Physics of Learning at 
\item[] \url{http://www.ipam.ucla.edu/programs/long-programs/machine-learning-for-physics-and-the-physics-of-learning/}
\item NIPS 2018 Workshop Machine Learning for Molecules and Materials at 
\url{http://www.quantum-machine.org/workshops/nips2018draft/}
\item NIPS Workshop on Machine Learning for Creativity and Design
\item[] $\circ$ in 2018 at \url{https://nips2018creativity.github.io/}
\item[] $\circ$ in 2017 at \url{https://nips2017creativity.github.io}
\end{itemize}


In this section, we attempt to put reinforcement learning in the wide context of science, engineering, and art.
We have already touched many aspects in previous chapters/sections.
Here we only discuss a small sample of the aspects we have not discussed before.

\subsubsection{ Chemistry}

Retrosynthesis is a chemistry technique to transform a target molecule into simpler precursors recursively.
\citet{Segler2018} propose to combine Monte Carlo tree search (MCTS) with symbolic rules for automatic retrosynthesis.
Three deep neural networks, namely, 
an expansion policy network to guide the search with transformations extracted automatically, 
a filter network to feasibility of the proposed reactions, 
and a rollout policy network to sample transformations to estimate the value of a position, 
are trained with almost all reactions published in organic chemistry.
The proposed approach improves previous computer-aided synthesis planning systems significantly.
\citet{Segler2018} follow the approach of AlphaGo in~\citet{Silver-AlphaGo-2016}.
It is interesting to study if the approach of AlphaGo Zero~\citep{Silver-AlphaGo-2017},
in particular, generalized policy iteration, self-play, and a single neural network, can be applied to retrosynthesis.

\citet{Popova2018} apply deep RL for computational de novo drug design, discovering molecules with desired properties. 
\citet{Jaques2017} as discussed below for music melody generation also study computational molecular generation.

\subsubsection{ Mathematics}

Deep learning has many applications in maths, e.g., 
neural ordinary differential equations (ODEs)~\citep{Chen2018ODE},
proofs~\citep{Irving2016, Loos2017, Rocktaschel2017, Urban2018NIPS}.
In the following, we discuss a case using deep RL for partial differential equations (PDEs). 

PDEs are mathematical tools for wide applications in science and engineering. 
It is desirable to design a PDE controller with minimal assumptions, without knowledge of the PDE, and being data-driven. 
\citet{Pan2018PDE} study how to control dynamical systems described by PDEs using RL methods, with high-dimensional continuous action spaces, having spatial relationship among action dimensions.
The authors propose action descriptors to encode such spatial regularities and to control such PDEs.
The authors show sample efficiency of action descriptors theoretically, 
comparing with conventional RL methods not considering such regularities.
The authors implement  action descriptors with Deep Deterministic Policy Gradient (DDPG)~\citep{Lillicrap2016},
and experiment with two high-dimensional PDE control problems.

\subsubsection{ Music}

\citet{Jaques2017} propose Sequence Tutor, combining maximum likelihood estimation (MLE) with RL, 
to consider both data and task-related goals,
to improve the structure and quality of generated sequences, 
and to maintain sample diversity and information learned from data,
by pre-training a Reward RNN using MLE, and training another RNN with off-policy RL, 
to generate sequences with high quality, 
considering domain-specific rewards, and penalizing divergence from the prior policy learned by Reward RNN. 
The authors investigate the connection between KL control and sequence generation,
and relationship among G-learning~\citep{Fox2016}, $\Psi$-learning~\citep{Rawlik2012}, and Q-learning.
The authors evaluate Sequence Tutor on musical melody generation.
It is nontrivial to design a reward function to capture the aesthetic beauty of generated melodies,
and a pure data-driven approach can not yield melodies with good structure.
Sequence Tutor incorporates rules from music theory into the model generating melodies,
to produce more pleasant-sounding and subjectively pleasing melodies than alternative methods.
The authors also conduct experiments with Sequence Tutor for computational molecular generation.

\citet{vandenOord2016WaveNet} propose WaveNet for raw audio waveforms generation.
See~\citet{Briot2018} about deep learning techniques for music generation.
See~\citet{Zhu2018XiaoIce} for pop music generation. 
See also \citet{Dieleman2018NIPS}.
See the Magenta project, \url{https://magenta.tensorflow.org}, 
for investigation of deep learning and reinforcement learning for art and music creation.
See the 2018 ICML, IJCAI/ECAI, and AAMAS Joint Workshop on Machine Learning for Music, \url{https://sites.google.com/site/faimmusic2018/}.

\clearpage


\section{Discussions}
\label{discussion}




We present deep reinforcement learning in this manuscript in an overview style.
We discuss the following questions: 
Why deep?
What is state of the art?
What are the issues, and potential solutions?
Briefly, the powerful and flexible representation by deep learning helps deep RL make many achievements.
We discuss state of the art, issues and potential solutions for deep RL in chapters on six core elements, six important mechanisms, and twelve applications.
In the following, we present a brief summary, discuss challenges and opportunities, and close with an epilogue.

\subsection{Brief Summary}

\begin{figure}
\centering
\includegraphics[width=1.0\linewidth]{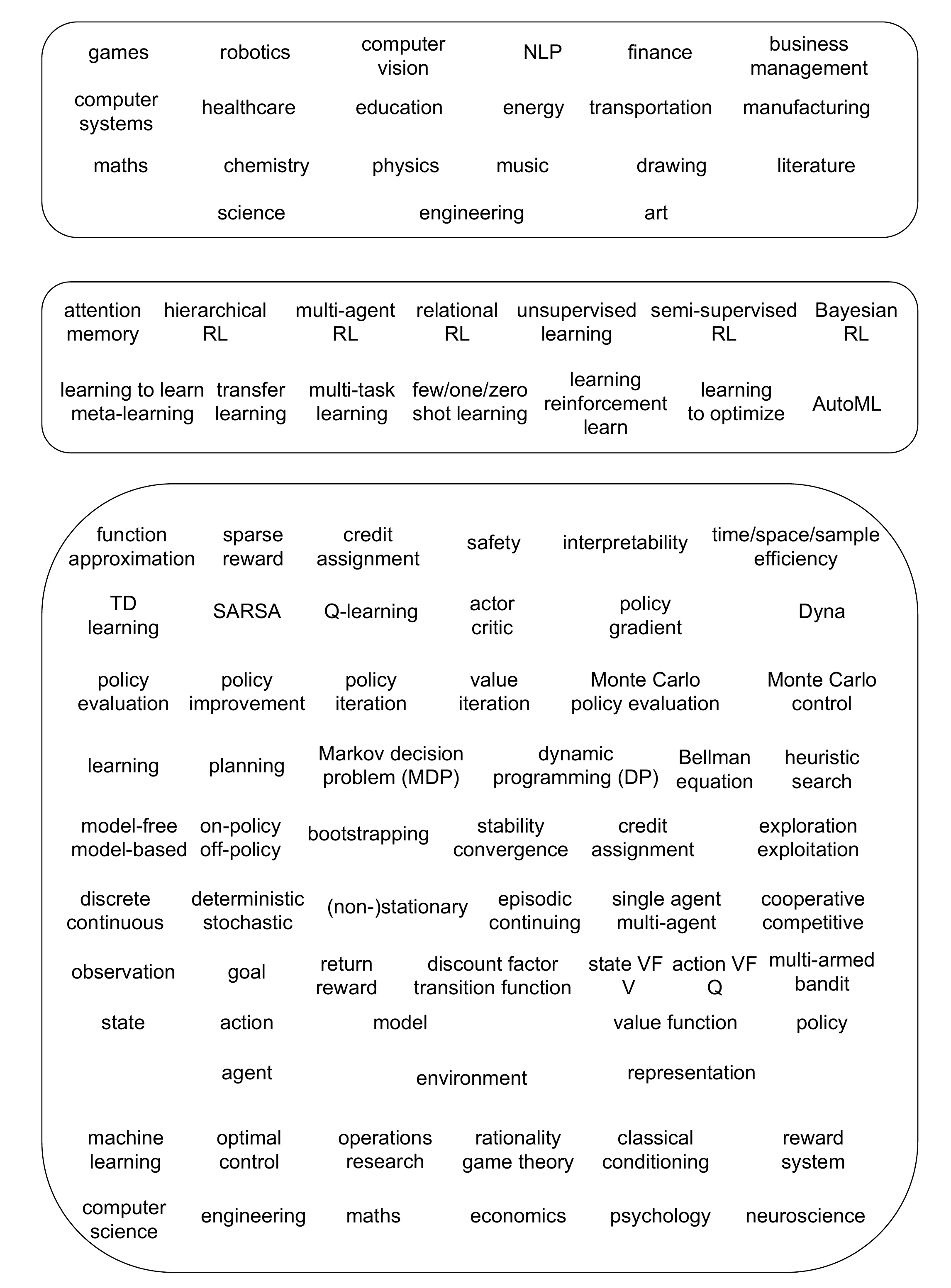}
\caption{Concepts, Algorithms, and Issues in Reinforcement Learning}
\label{issues}
\end{figure}

There are many concepts, algorithms, and issues in (deep) reinforcement learning (RL), as illustrated in Figure~\ref{issues}.
We touch many of them in this manuscript.

Credit assignment, sparse reward, and sample efficiency are common issues for RL problems.
We list several approaches proposed to address them in the following.
In off-policy learning, both on-policy and off-policy data can be used for learning.
Auxiliary reward and self-supervised learning are for learning from non-reward signals in the environment.
Reward shaping is for providing denser rewards.
Hierarchical RL is for temporal abstraction. 
General value function, in particular, Horde, universal function approximator, and hindsight experience replay, is for learning shared representation/knowledge among goals.
Exploration techniques are for learning more from valuable actions.
Model-based RL can generate more data to learn from.
Learning to learn, e.g., one/zero/few-shot learning, transfer learning,  and multi-task learning, 
learns from related tasks to achieve efficient learning.
Incorporating inductive bias, structure, and knowledge can help achieve more intelligent representation and problem formulation.
And so on and so forth.

\citet{Finn2017TalkModelRL} and \citet{Levine2018} discuss that sample efficiency improves roughly by ten times in each step from one RL method to another in the following,
from gradient-free methods, like CMA-ES, 
fully online methods like A3C, 
policy gradient methods, like TRPO, 
value estimation methods with reply buffer, like Q-learning, DQN, DDPG, and NAF,
model-based deep RL methods, like guided policy search (GPS), 
all the way to
model-based "shallow" RL methods, like PILCO.

\citet{Silver2018Principles} summarizes principles of deep RL:
evaluation drives progress,
scalability determines success,
generality future-proofs algorithms,
trust in the agent?s experience,
state is subjective,
control the stream,
value functions model the world,
planning: learn from imagined experience,
empower the function approximator, 
and,
learn to learn.

Some issues deserve more discussions, in particular, safety and interpretability.
We discuss several aspects of AI safety, e.g.,  
reward in Chapter~\ref{reward}, 
multi-agent RL in Chapter~\ref{MARL},
and, adversarial examples in Section~\ref{systems}.
There are recent work on RL safety, e.g.,
\citet{Chow2018NIPS},
\citet{HuangJiexi2018NIPS}, 
and \citet{WenMin2018NIPS}.
See surveys for AI safety~\citep{Amodei2016, Garcia2015}.
See a course on safety and control for artificial general intelligence, at \url{http://inst.eecs.berkeley.edu/~cs294-149/}.
See blogs about safety, e.g., at \url{https://medium.com/@deepmindsafetyresearch}.
\citet{Doshi-Velez2017}, 
\citet{Lage2018NIPS}, 
\citet{Lipton2018ACM},  and 
\citet{Melis2018NIPS} 
discuss issues of interpretability. 
\citet{Zhang2018Interpretability} survey visual interpretability for deep learning.

\subsection{Challenges and Opportunities}
\label{opportunities}


In the following, we discuss issues in deep RL,
and propose research directions as both challenges and opportunities,
which are challenging, or even widely open.


\citet{Rahimi2017} raise the concern that "machine learning has become alchemy". 
This alludes in particular to deep learning. 
See their blogs, \citet{Rahimi2017Alchemy}, \citet{Rahimi2017Addendum}.
In~\citet{Sculley2018}, Ali Rahimi and colleagues discuss productive changes for empirical rigor,
and recommend standards for empirical evaluation:
tuning methodology,
sliced analysis,
ablation studies,
sanity checks and counterfactuals,
and at least one negative result. 

\citet{Lipton2018} discuss troubling trends in machine learning, including 
failure to distinguish between explanation and speculation,
failure to identify the sources of empirical gains, 
mathiness, which obfuscates or impresses but not clarifies with mathematics, 
and, misuse of language, with suggestive definitions, overloading technical terminology, or suitcase words for a variety of meanings.
See NIPS 2018 Workshop on Critiquing and Correcting Trends in Machine Learning.

\citet{Henderson2018} investigate reproducibility, experimental techniques, and reporting procedures for deep RL.
The authors show that 
experimental results are influenced by
hyperparameters, including 
network architecture and reward scale, 
random seeds and trials,
environments (like Hopper or HalfCheetah etc. in OpenAI Baseline), 
and codebases.
This causes difficulties for reproducing deep RL research results.
The authors analyze the following reporting evaluation metrics:
online view vs. policy optimization (offline),
confidence bounds (sample bootstrap),
power analysis (about sample size),
significance (like $t$-test).
\citet{Henderson2018} recommend to 
report implementation details, all hyperparameter settings,  experimental setup, and evaluation methods for reproducibility.
The authors also recommend to
find the working set of hyperparameters,
match baseline algorithm implementation with the original codebase,
run many trails with different random seeds then average results, and
perform proper significance tests to validate better performance.

\citet{Khetarpal2018} discuss evaluation differences in RL and in supervised learning, and propose an evaluation pipeline.

\citet{Levine2018} discusses challenges with deep RL, 
including stability, efficiency, scalability, hyperparameters tuning, sample complexity, model-based learning, generalization, reward specification, prior knowledge, etc.

These papers
\footnote{
There is a blog titled Deep Reinforcement Learning Doesn't Work Yet at \url{https://www.alexirpan.com/2018/02/14/rl-hard.html}.
It summarizes issues with deep RL, including
sample inefficiency,
better results with non-RL methods,
issues with reward function,
local optimal hard to escape,
overfitting,
and, hard to reproduce due to instability.
The blog contains informative discussions; however, the title is wrong.
There is another blog titled Lessons Learned Reproducing a Deep Reinforcement Learning Paper at \url{http://amid.fish/reproducing-deep-rl}.
}
 discuss various issues with deep learning, machine learning, deep RL, and provide valuable insights. 
There are also benchmark papers like \citet{Duan2016}.
However, we still lack papers conducting systematic, comparative study of deep RL algorithms, 
so that we pick one or more benchmark problems, 
do a thorough study,
report both successes and failures,
summarize advices and lessons,
and, give guidelines about how to use deep RL algorithms.
Our deep RL community need such papers.
As well, most RL + NLP/computer vision papers use REINFORCE. A natural question is: how about other (deep) RL algorithms?
We can evaluate performance of many algorithms, like  DQN, A3C, DDPG, TRPO, PPO,  PCL, Trust-PCL, Retrace, Reactor, interpolated policy gradient, soft Q-learning, etc.
As such, we propose the following research direction.

\vspace{3mm}
\emph{Research Direction 1: systematic, comparative study of deep RL algorithms}
\vspace{3mm}

\citet{Bellemare2018Dopamine} open source Dopamine, 
aiming for a flexible, stable, and reproducible Tensorflow-based RL framework,
as an achievement in this direction.

We have seen exciting results in two-player and multi-agent games recently.
AlphaGo~\citep{Silver-AlphaGo-2016, Silver-AlphaGo-2017} has achieved super-human performance.
DeepStack~\citep{Moravcik2017} defeated professional poker players.
\citet{Jaderberg2018Quake} achieve human level performance in the game of Capture the Flag (Chapter~\ref{MARL}).
OpenAI Five has beaten  human players on 5v5 Dota 2, although with huge computation (\url{https://blog.openai.com/openai-five/}).  
\citet{Zambaldi2018} achieve decent results on StarCraft II mini-games (Chapter~\ref{relational}). 
\citet{SunPeng2018StarCraft} and \citet{Pang2018StarCraft} have beaten full-game built-in AI in StarCraft II.

However, multi-agent problems are still very challenging, with issues like non-stationarity
and even theoretical infeasibility, as we discuss in Chapter~\ref{MARL}.
Even so, we can endeavour to achieve decent results for multi-agent problems,
like approximation solutions with high quality, and/or super-human performance.
Multi-agent systems are a great tool to model interactions among agents,
with rich applications in human society;
and their advancements can significantly push the frontier of AI.   
We thus propose the second research direction as below. 

\vspace{3mm}
\emph{Research Direction 2: "solve" multi-agent problems}
\vspace{3mm}

StarCraft and Texas Hold'em Poker are great testbeds for studying multi-agent problems.
It is desirable to see extensions of DeepStack to multi-player settings, with many hands of playing, 
in tournament and cash game styles.

StarCraft features many possible actions, complex interactions between players, short term tactics and long term strategies, etc. 
Learning strategies for Starcraft following videos with commentary would be a feasible strategy.
There are many videos about StarCraft with excellent commentaries.
If we may be able to extract valuable information, like strategies, from the multi-modality signals, and apply these to the agent design, we may be able to achieve a human level AI StarCraft agent.
Such a system would be an integration of RL, computer vision, and NLP.
\citet{Aytar2018} achieve breakthrough results on three hard Atari games with self-supervision techniques by watching YouTube (Chapter~\ref{unsupervised}). This may give us more motivation and encouragements. 
As a related work, \citet{Branavan2012} propose to learn strategy games by reading manuals.
With achievements in \citet{SunPeng2018StarCraft} and \citet{Pang2018StarCraft},
it is interesting to watch if hierarchical RL approaches in these papers can achieve super-human performance.

We now discuss end-to-end learning with raw inputs, a trendy paradigm recently, e.g., 
AlexNet~\citep{Krizhevsky2012} with raw pixels for image classification, 
Seq2Seq~\citep{Sutskever2014} with raw sentences for machine translation, 
DQN~\citep{Mnih-DQN-2015} with raw pixels and score to play Atari games,
AlphaGo Zero~\citep{Silver-AlphaGo-2017} with piece information and score to play computer Go,
and, \citet{Jaderberg2018Quake} with raw pixels and score to play Quake III Arena Capture the Flag.

One question is, is such paradigm of end-to-end learning with raw input good? 
Sample efficiency is usually an issue. 
For example, as shown in~\citet{Hessel2018}, Rainbow needs  44 million frames to exceed the performance of distributional DQN~\citep{Bellemare2017Distributional}, which needs much less data than DQN~\citep{Mnih-DQN-2015}. 
Such huge amount of data require huge computation.

Another issue is adversarial examples, which may be more severe for critical applications.
\citet{Szegedy2013} show that various images, like a truck, a building, or a dog,  after being added imperceptible noises, 
are all classified by AlexNet as "ostrich, Struthio camelus". 
\citet{Eykholt2018} show that physical images, e.g., stop signs, left turn signs etc., after being perturbed by adding black or white stickers, are misclassified by state of the art deep neural networks as speed limit 45 signs. 

Some papers propose to learn fully autonomously.
AlphaGo Zero~\citep{Silver-AlphaGo-2017} and \citet{Jaderberg2018Quake} have achieved such a goal to some extend.
However, we have to admit that both computer Go and Quake III Arena Capture the Flag have perfect simulation models, and unlimited data can be generated relatively easily.
Many practical applications, like robotics, healthcare, and, education, do not have such luxury.
We may or may not ultimately achieve such a goal of fully autonomous learning in a general sense;
and it is not clear for problems with practical concerns.
Consider education.
Most of us follow some curricula designed by experts, and learn from experts,
rather than learning tabula rasa.
Even we will achieve such a goal, as in most scientific discoveries, 
we may encounter spiral development, rather than going straightforwardly to the goal. 
We probably need some hybrid solution in the interim.  

\begin{figure}[h]
\centering
\includegraphics[width=0.96\linewidth]{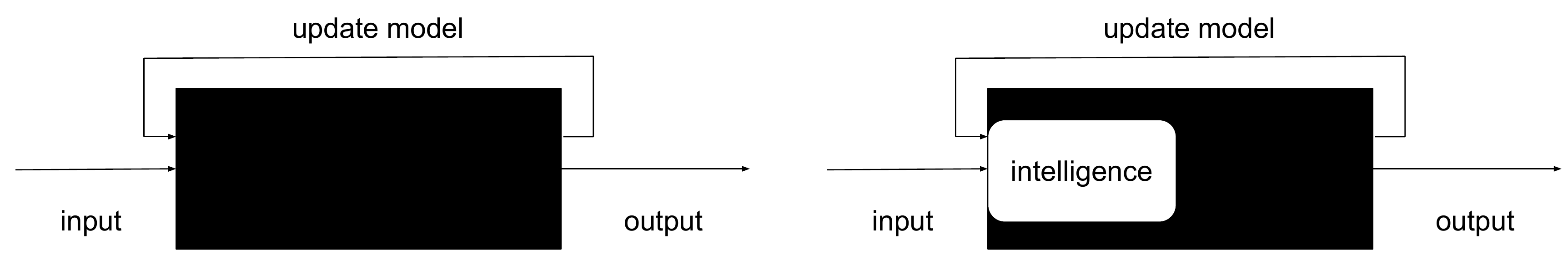}
\caption{Add An Intelligence Component to End-to-End Pipeline}
\label{intelligence}
\end{figure}

We expect that manual engineering reconciles with end-to-end learning, and symbolism reconciles with connectionism. 
We thus propose to add an "intelligence" component in the end-to-end processing pipeline, rather than treating the system as an entire blackbox, as most current deep neural networks do, as shown in Figure~\ref{intelligence}.

In the intelligence component, we may incorporate
common knowledge like common sense, inductive bias, knowledge base, etc.,
common principles like Newton's laws, Bellman equation, etc.,
and, 
common algorithms like gradient descent, TD learning, policy gradient, etc.

The idea of adding an intelligent component is aligned with incorporating human intelligence as discussed in \citet{Lake2016}.
For example, when we study how to play billiard with a computer program, we probably want to incorporate Newton's law, 
rather than using deep learning to rediscover such laws with many video clips.
\citet{Graves2013} follows end-to-end training for speech recognition, with a Fourier transformation of audio data.
Self-supervised learning would be a promising approach for adding this intelligence component,
e.g., \citet{Jaderberg2017} and \citet{Aytar2018}. 

Adding an intelligence component is abstract.
Now we discuss something more concrete, esp. for tasks with perception, like with visual inputs.
We then propose the following research direction.

\vspace{3mm}
\emph{Research Direction 3: learn from entities, but not just raw inputs} 
\vspace{3mm}

Our goal is to make the learning system more  efficient w.r.t. sample, time, and space, 
to achieve interpretability
and to avoid obvious mistakes like those in adversarial examples.
At the same time, we still strive for end-to-end processing, and being fully differentiable.
Our thesis is that, if we could process the raw input with some principle or knowledge, 
the resulting representation would be more convenient for the learning system to make further predictions or decisions.

Take the hard Atari game Montezuma's Revenge as an example.
Suppose there were an intelligent system, which could identify entities in video frames, like agent, road, ladder, enemy, key, door, etc., and their attributes and relationships.
Then a RL agent working on such representation would be much more efficient than working on pixels.
A question is, if RL, unsupervised learning, or some machine learning/AI techniques can help identify entities, attributes, and their relationships.
Successes in this direction would hinge on the maturity of computer vision, NLP, and AI. 

There are recent progress in this direction.
\citet{Goel2018NIPS} conduct unsupervised video object segmentation for deep RL.
\citet{Eslami2018GQN} present GQN for discovering representations with unsupervised learning.
\citet{Chen2018} propose a framework for iterative visual reasoning.   
For NLP, word2vec~\citep{Mikolov2013, Mikolov2017} is probably the most popular representation.
\citet{vandenOord2018} propose to learn representations for multi-modality, including speech, images, text, and reinforcement learning.
There are some recent papers about reasoning, e.g.~\citep{Santoro2017, Hudson2018, Battaglia2018GN},
as we discuss in Chapter~\ref{relational}.
We also discuss knowledge and reasoning in Section~\ref{representation:knowledge}.

\citet{Malik2018IJCAI} discusses that there are great achievements in the fields of vision, motor control, and language semantic reasoning,
and it is time to investigate them together.
This supports our proposal. 

Another fundamental, and related issue is about representation for RL problems.
In deep learning, as well as in deep RL, neural network architecture is critical for the performance.


There are classical ways for function approximation, 
several popular neural network architectures,
mechanisms for temporal abstraction,
neural network architectures designed for deep RL and/or for reasoning,
and discussions about causality and human intelligence.
We discuss such representation issues in Chapter~\ref{representation}.

When we talk about computer vision with deep learning, CNNs appear in many people's minds. 
When we talk about RL algorithms, many people think about TD learning, Q-learning, and policy gradient.
However, when we talk about  representation or neural network architecture for (deep) RL, 
different people may come up with different ideas.
It would be great to discover something for RL like CNNs for computer vision.
We thus propose the following research direction.

\vspace{3mm}
\emph{Research Direction 4: design an optimal representation for RL}
\vspace{3mm}

RL problems have their own structures and characteristics, e.g., value functions satisfy Bellman equation, 
so that they are probably different from those in deep learning, like image recognition and machine translation.
Consequently, RL problems probably have their own optimal representation and neural network architecture.
We conjecture that it is desirable to consider a holistic approach, 
i.e., considering perception and control together, rather than separately.
\citet{Srinivas2018} and \citet{Tamar2018NIPS} are efforts in this direction.

To go one step further, we propose the next research direction to automate reinforcement learning, namely, AutoRL.
A successful AutoRL tool would help us choose optimal 
state representation,
function approximator, 
learning algorithms, 
hyperparameters, etc.
There are efforts currently for machine learning tasks, namely, AutoML, as we discuss in Section~\ref{meta:AutoML}.

\vspace{3mm}
\emph{Research Direction 5: AutoRL}
\vspace{3mm}

We now talk about the last research direction. 
Reinforcement learning is very powerful and very important.
Quoted from the authoritative AI textbook~\citep{Russell2009}, 
"reinforcement Learning might be considered to encompass all of AI: an agent is placed in an environment and must learn to behave successfully therein", and,
"reinforcement learning can be viewed as a microcosm for the entire AI problem".
Moreover, David Silver
proposes a conjecture:  
artificial intelligence = reinforcement learning + deep learning~\citep{Silver2016Tutorial}.
We attempt to justify this conjecture as follows.
\citet{Hornik1989} establish that multilayer feedforward networks are universal approximators; 
that is, a feedback neural network with a single layer is sufficient to  approximate any continuous function to an arbitrary precision.
\citet{Hutter2005} proves that tasks with computable descriptions in computer science can be formulated as RL problems. 
With deep learning providing mechanisms, and reinforcement learning defining the objective and achieving it,
their integration can solve computable tasks, the aim of AI.
Note the number of units in the hidden layer may be infeasibly large though,
and, computability may be an issue, unless P = NP.

However, we see that deep learning is used much more widely, in supervised learning, unsupervised learning, and reinforcement learning.  
Furthermore, deep learning is also widely used in many applications, and is the core technique for many commercial products, like those with face recognition, speech recognition, etc.  
We enjoy the successes of deep RL, like those in Atari games and computer Go, which probably have limited commercial value. 
We see successes of a special RL family of techniques, namely, multi-armed bandits, for applications like news recommendation~\citep{Li2010}. 
We also see achievements like those for neural architecture design~\citep{Zoph2017}.  
However, reinforcement learning still needs more efforts to become more practical, 
and we are still lacking of wide and practical applications of reinforcement learning that generate considerable commercial value.
We thus propose the following research direction.

\vspace{3mm}
\emph{Research Direction 6: develop killer applications for (deep) RL}
\vspace{3mm}

Successes of this research direction require the maturity of RL algorithms, for efficiency, stability, and robustness, etc. 
We see a positive feedback loop between algorithms and applications; they will help each other to make further improvements.

We now discuss a concrete recommendation for this direction:
it is promising to invest more on applying AlphaGo techniques.
AlphaGo techniques, in particular, deep learning, reinforcement learning, MCTS, and self play, are successful techniques, and have many potential applications. In particular, the elegant algorithm of AlphaGo Zero~\citep{Silver-AlphaGo-2017} would be straightforwardly applicable to a big family of problems. 

We list potential applications of AlphaGo as suggested by the authors in their papers~\citep{Silver-AlphaGo-2016, Silver-AlphaGo-2017}, namely, 
general game-playing (in particular, video games)
classical planning,
partially observed planning,
scheduling,
constraint satisfaction,
robotics,
industrial control,
online recommendation systems,
protein folding,
reducing energy consumption, and
searching for revolutionary new materials. 
Although AlphaGo techniques have limitations, like requiring a good or even perfect simulator,
we expect to see more and more application of AlphaGo techniques.

We list six research directions, as both challenges and opportunities of deep RL.
Research direction 1, systematic, comparative study of deep RL algorithms,
is about reproducibility, and under the surface, 
about stability and convergence properties of deep RL algorithms.
Research direction 2, "solve" multi-agent problems,
is usually about sample efficiency, sparse reward, stability, non-stationarity, and convergence in a large-scale, complex setting, a frontier in AI research.
Research direction 3, learn from entities, but not just raw inputs,
is about sample efficiency, sparse reward, and interpretability, by incorporating more knowledge, structure, and inductive bias. 
Research direction 4, design an optimal representation for RL,
research direction 5, AutoRL, and,
research direction 6, develop killer applications for (deep) RL,
are about the whole RL problem, about all issues in RL, 
like credit assignment, sparse reward, time/space/sample efficiency, accuracy, stability, convergence, interpretability, safety, scalability, robustness, simplicity,  etc,
from different angles of representation, automation, and application, respectively.
We expect all these research directions to be open,
except the first one, which is also challenging,
and progress in these directions would deepen our understanding of (deep) RL,
and push further frontiers of AI.

\subsection{Epilogue}

It is both the best and the worst of times for the field of deep RL, for the same reason: it has been growing so fast and so enormously. We have been witnessing breakthroughs, exciting new algorithms, architectures, and applications, and we expect to see much more and much faster. As a consequence, this manuscript is incomplete, in the sense of both depth and width. However, we attempt to summarize important achievements and discuss potential directions and applications in this amazing field. 

Value function is central to reinforcement learning, e.g., in Deep Q-Network and its many extensions. 
Policy optimization approaches have been gaining traction, with many new algorithms, and in many, diverse applications, e.g., robotics, neural architecture design, spoken dialogue systems, machine translation, attention, and learning to learn, etc. This list is boundless. 
New learning mechanisms have emerged, e.g., using learning to learn, unsupervised learning, self-supervised learning, etc., to improve the quality and speed of learning, and more new mechanisms will be emerging. 
This is the renaissance of reinforcement learning~\citep{Krakovsky2016}. 
In fact, deep learning and reinforcement learning  have been making steady progress even during the last AI winter.  

We have seen breakthroughs about deep RL, including DQN~\citep{Mnih-DQN-2015}, AlphaGo~\citep{Silver-AlphaGo-2016, Silver-AlphaGo-2017}, and DeepStack~\citep{Moravcik2017}. 

Exciting achievements abound: 
differentiable neural computer~\citep{Grave-DNC-2016}, 
unsupervised reinforcement and auxiliary learning~\citep{Jaderberg2017}, 
asynchronous methods~\citep{Mnih-A3C-2016}, 
guided policy search~\citep{Levine2016}, 
generative adversarial imitation learning~\citep{Ho2016}, 
and neural architecture design~\citep{Zoph2017}, etc. 
There are also many recent, novel applications of (deep) RL in many, diverse areas as discussed in previous chapters.
Creativity would push the frontiers of deep RL further with respect to core elements, important mechanisms, and applications.
In general, RL is probably helpful, if a problem can be regarded as or transformed to a sequential decision making problem, 
and states, actions, maybe rewards, can be constructed.
Roughly speaking, if a task involves some manual designed "strategy", then there is a chance for reinforcement learning to help to automate and optimize the strategy.

Having a better understanding of how deep learning works is helpful for deep learning, machine learning, and AI. 
\citet{Poggio2017} review why and when deep- but not shallow-networks can avoid the curse of dimensionality.
See Stanford STATS 385 course on Theories of Deep Learning at \url{https://stats385.github.io}.
See \citet{Arora2018} about theoretical understanding of deep learning.
There are also papers for interpretability of deep learning, e.g.~\citet{Doshi-Velez2017}, \citet{Lipton2018ACM}, and \citet{Zhang2018Interpretability}.

It is important to investigate comments/criticisms for further progress.
A popular criticism about deep learning is that it is a blackbox, 
or even an "alchemy" during the NIPS 2017 Test of Time Award speech~\citep{Rahimi2007}. 
\citet{Lake2016} discuss incorporating machine intelligence with human intelligence for stronger AI;
one commentary, \citet{Botvinick2017}, discusses the importance of autonomy.  
\citet{Jordan2018AI} discusses issues with AI.
\citet{Darwiche2018} discusses deep learning in the context of AI.
See Peter Norvig's perspective~\citep{Press2016}. 
\citet{Marcus2018} criticizes deep learning,
and \citet{Dietterich2018} responds.
Watch two debates, \citet{LeCun2017Marcus}, \citet{LeCun2018Manning}.
See~\citet{Stoica2017} for systems challenges for AI.

It is worthwhile to envision deep RL considering  perspectives from the society, academia and industry on AI, e.g., Artificial Intelligence, Automation, and the Economy, Executive Office of the President, USA;  Artificial Intelligence and Life in 2030 - One Hundred Year Study on Artificial Intelligence: Report of the 2015-2016 Study Panel, Stanford University~\citep{Stone2016}; and AI, Machine Learning and Data Fuel the Future of Productivity by The Goldman Sachs Group, Inc., etc. 
\citet{Brynjolfsson2017} and \citet{Mitchell2017} discuss implications of AI and machine learning for workforce.
There are also many articles, e.g., in Harvard Business Review, like~\citet{Agrawal2017AI}, \citet{Ng2016ChiefAI}, and \citet{Ng2016AI}.
See recent books about the science and technology of AI and machine learning, and their implications for business and society, e.g., \citet{Agrawal2018}, \citet{Domingos2015}, and \citet{Lee2018}.

Nature in May 2015 and Science in July 2015 featured survey papers on machine learning and AI.  
Science Robotics was launched in 2016. 
Science has a special issue on July 7, 2017 about AI on The Cyberscientist. 
Nature Machine Intelligence was launched in January 2019.
These illustrate the apparent importance of AI. 
It is interesting to mention that NIPS 2018 main conference was sold out in less than 12 minutes after opening for registration; see~\citep{Li2018AI} .  

Deep learning was among MIT Technology Review 10 Breakthrough Technologies in 2013. We have been witnessing the dramatic development of deep learning in both academia and industry in the last few years. Reinforcement learning was among MIT Technology Review 10 Breakthrough Technologies in 2017. Deep learning has made many achievements, has "conquered" speech recognition,  computer vision, and now NLP, is more mature and well-accepted, and has been validated by products and market. In contrast, RL has lots of (potential yet promising) applications, yet not many wide-spread products so far.
RL may still need better algorithms, and may still need  products and market validation. 
Prediction is very difficult, especially about the future.
However, it is probably the right time to nurture, educate and lead the market for reinforcement learning. 
We will see both deep learning and reinforcement learning prospering in the coming years and beyond.  

Deep learning, in this third wave of AI, will have deeper influences, as we have already seen from its many achievements. Reinforcement learning, as a more general learning and decision making paradigm, will deeply influence deep learning, machine learning, and artificial intelligence in general.  It is interesting to mention that when Professor Rich Sutton started working in the University of Alberta in 2003, he named his lab RLAI: Reinforcement Learning and Artificial Intelligence.

\clearpage



These abbreviations are used for frequently cited conferences and journals.

\vspace{3mm}

\begin{tabular}  { l p{11cm} } 
AAAI &  the AAAI Conference on Artificial Intelligence\\
ACL &  the Association for Computational Linguistics Annual Meeting\\
AAMAS &  the International Conference on Autonomous Agents \& Multiagent Systems\\
ArXiv & ArXiv e-prints\\
CCS &   the ACM Conference on Computer and Communications Security\\
CVPR &  the IEEE Conference on Computer Vision and Pattern Recognition\\
EMNLP &  the Conference on Empirical Methods in Natural Language Processing\\
ICCV &  the IEEE International Conference on Computer Vision\\
ICLR &  the International Conference on Learning Representations\\
ICML &  the International Conference on Machine Learning\\
ICRA &  IEEE International Conference on Robotics and Automation\\
IJCAI &  the International Joint Conference on Artificial Intelligence\\
IROS &  International Conference on Intelligent Robots\\
JAIR & Journal of Artificial Intelligence Research\\
JMLR & the Journal of Machine Learning Research \\
KDD &   the ACM International Conference on Knowledge Discovery and Data Mining\\
NAACL & the Annual Conference of the North American Chapter of the Association for Computational Linguistics: Human Language Technologies\\
NIPS &   the Annual Conference on Neural Information Processing Systems\\
RSS & Robotics: Science and Systems\\
TNN & IEEE Transactions on Neural Networks\\
TPAMI & IEEE Transactions on Pattern Analysis and Machine Intelligence\\
UAI &  the Conference on Uncertainty in Artificial Intelligence\\
WWW & the International World Wide Web Conference




\end{tabular}

\clearpage



\clearpage

\clearpage
{\footnotesize
  \bibliography{DeepRL}
}
\bibliographystyle{apa}


\end{document}